\PassOptionsToPackage{capitalize}{cleveref} % capitalized cross-references ("Table 1"), as in the NeurIPS version
\documentclass{applemlr}
\usepackage{amsmath}
\usepackage{enumerate}
\usepackage{algorithm}
\usepackage{algpseudocode}
\usepackage{amsfonts}
\usepackage{amsthm}
\usepackage[noabbrev,nameinlink]{cleveref}
\usepackage{diagbox}
\usepackage{colortbl}
\usepackage{amssymb}
\usepackage{xspace}
\usepackage{wrapfig}
\usepackage{adjustbox}
\usepackage{tabularx}
\usepackage{booktabs}
\usepackage{mathtools}
\usepackage{tikz}
\usepackage{enumitem}
\usepackage{silence}
\usepackage{dsfont}
\usepackage[table]{xcolor}
\usepackage[dvipsnames]{xcolor}
\usepackage{multirow}
\usepackage{makecell}
\usepackage{xfakebold}
\usepackage{amsmath,amsfonts,bm}

\def\eqref#1{equation~\ref{#1}}
\def\1{\bm{1}}

\DeclareMathAlphabet{\mathsfit}{\encodingdefault}{\sfdefault}{m}{sl}
\SetMathAlphabet{\mathsfit}{bold}{\encodingdefault}{\sfdefault}{bx}{n}

\newcommand{\R}{\mathbb{R}}

\definecolor{textgray}{HTML}{6E6E73}
\usetikzlibrary{positioning, calc}
\usetikzlibrary{decorations.pathmorphing}
\makeatletter
\patchcmd{\wrong@fontshape}{\@gobbletwo}{}{}{}
\makeatother
\numberwithin{equation}{section}
\makeatletter
\AtBeginDocument{
  \urlstyle{sf}
  
}
\makeatother
\definecolor{light}{RGB}{125, 125, 125}
\crefname{tcb@cnt@pbox}{code}{code}
\Crefname{tcb@cnt@pbox}{Code}{Code}
\crefname{assumption}{assumption}{assumption}
\Crefname{assumption}{Assumption}{Assumptions}
\newtcolorbox[auto counter]{pbox}[2][]{
  colback=white,
  title=Code~\thetcbcounter: #2,
  #1,fonttitle=\sffamily,
  fontupper=\sffamily,
  arc=2pt,
  colframe=bgcolor,
  coltitle=fgcolor,
  colbacktitle=bgcolor,
  toptitle=0.25cm,
  bottomtitle=0.125cm
}

\makeatletter
\newcommand\applefootnote[1]{%
  \begingroup
  \renewcommand\thefootnote{}%
  \renewcommand\@makefntext[1]{\noindent##1}%
  \footnote{#1}%
  \addtocounter{footnote}{-1}%
  \endgroup
}
\makeatother
\definecolor{cverbbg}{gray}{0.90}

\theoremstyle{plain}

\usepackage{bbm}

\usepackage{subcaption}
\usepackage{etoc}

\usepackage{nicefrac}

\usepackage{pgfplots}
\pgfplotsset{compat=1.17}
\usetikzlibrary{patterns, arrows.meta, positioning, calc,
                decorations.pathreplacing, shapes.geometric,
                fit, backgrounds, decorations.markings}

\usepackage{fontawesome5}

\usepackage{tcolorbox}
\tcbuselibrary{breakable, skins}

\let\R\relax

\crefname{prop}{proposition}{propositions}
\Crefname{prop}{Proposition}{Propositions}

\usepackage{pifont}
\makeatletter
\let\REQUIRE\Require

\let\STATE\State
\let\ENDIF\EndIf
\let\ENDFOR\EndFor

\let\ELSE\Else

\newcommand{\IF}[1]{\If{#1}}

\newcommand{\FOR}[1]{\For{#1}}

\newcommand{\COMMENT}[1]{\Comment{#1}}

\newcommand{\RETURN}{\State \textbf{return}~}

\makeatother

\definecolor{energyblue}{HTML}{2563EB}
\definecolor{fsdfmgreen}{HTML}{16A34A}
\definecolor{baselinegray}{HTML}{6B7280}
\definecolor{v2orange}{HTML}{EA580C}
\definecolor{teacherred}{HTML}{DC2626}
\definecolor{noisezone}{HTML}{FEF3C7}
\definecolor{guidezone}{HTML}{DBEAFE}
\definecolor{signalgreen}{HTML}{DCFCE7}
\definecolor{trajgood}{HTML}{059669}
\definecolor{trajbad}{HTML}{B91C1C}
\definecolor{lightblue}{HTML}{EFF6FF}
\definecolor{ctlpurple}{HTML}{7C3AED}
\definecolor{ctllight}{HTML}{F3E8FF}
\definecolor{shapegold}{HTML}{D97706}
\definecolor{shapegoldlight}{HTML}{FEF3C7}

\definecolor{energyblue}{HTML}{3B82F6}
\definecolor{velgreen}{HTML}{22C55E}
\definecolor{safeguardgold}{HTML}{F59E0B}
\definecolor{rejectred}{HTML}{EF4444}
\definecolor{cDark}{HTML}{1F2937}

\definecolor{compassC}{HTML}{0E6655}
\definecolor{compassF}{HTML}{D1F2EB}
\definecolor{navC}{HTML}{7D3C98}
\definecolor{navF}{HTML}{F4ECF7}
\definecolor{threshC}{HTML}{B7950B}
\definecolor{threshF}{HTML}{FEF9E7}
\definecolor{teachC}{HTML}{2C3E50}
\definecolor{teachF}{HTML}{D5D8DC}
\definecolor{studC}{HTML}{C0392B}
\definecolor{studF}{HTML}{FADBD8}
\definecolor{emaC}{HTML}{1A5276}
\definecolor{emaF}{HTML}{D4E6F1}
\definecolor{rkC}{HTML}{6C3483}
\definecolor{rkF}{HTML}{E8DAEF}
\definecolor{lossC}{HTML}{D35400}
\definecolor{flowC}{HTML}{117A65}
\definecolor{flowF}{HTML}{D1F2EB}
\definecolor{noiseC}{HTML}{7F8C8D}
\definecolor{noiseF}{HTML}{F2F3F4}
\definecolor{dataC}{HTML}{1E8449}
\definecolor{dataF}{HTML}{D5F5E3}
\definecolor{goldC}{HTML}{B7950B}
\definecolor{goldF}{HTML}{FEF9E7}
\definecolor{navy}{HTML}{1B2631}
\definecolor{grayT}{HTML}{566573}
\definecolor{softG}{HTML}{D5D8DC}
\definecolor{initC}{HTML}{8E44AD}
\definecolor{selectedrow}{HTML}{E8F5E9}
\definecolor{beatteacher}{HTML}{E8F5E9}  % light green for cells beating teacher

\newcommand{\ours}{TS-DFM\xspace}

\newcommand{\bt}[1]{\cellcolor{beatteacher}{#1}}  % beat-teacher cell

\title{Trajectory as the Teacher: Few-Step Discrete Flow Matching via Energy-Navigated Distillation}

\author[1]{Amin Karimi Monsefi}
\author[1]{Dominic Culver}
\author[1]{Nikhil Bhendawade}
\author[1]{Manuel R. Ciosici}
\author[2,\ddagger]{Yizhe Zhang}
\author[1]{Irina Belousova}
\affiliation{$^{1}$Apple \quad $^{2}$Meta}

\abstract{
Discrete flow matching generates text by iteratively transforming noise tokens into coherent language, but may require hundreds of forward passes. Distillation uses the multi-step trajectory to train a student to reproduce the process in a few steps. When the student underperforms, the usual explanation is insufficient capacity. We argue the opposite: \emph{the trajectory is the bottleneck, not the student}. Each training trajectory is built through a chain of blind stochastic jumps with no evaluation of sequence quality. A single bad decision at an early midpoint propagates through subsequent steps, yet the student must imitate the result. Trajectory-Shaped Discrete Flow Matching (\ours) replaces these blind jumps with guided navigation: a lightweight energy compass evaluates candidate continuations at each midpoint, selecting the most coherent. All shaping is training-only; inference cost is unchanged. On 170M-parameter language modeling, the shaped student at 8 steps achieves 32\% lower perplexity than the unguided $1\,024$-step teacher while being $128\times$ faster, with gains consistent across source distributions and three evaluators of increasing scale. \ours achieves the best perplexity of any discrete-generation baseline we compare against, including methods trained on $6\times$ more data or using $5\times$ larger models.

}

\metadata[Correspondence]{\sffamily
  Amin Karimi Monsefi: \url{a_karimimonsefi@apple.com};
  Irina Belousova: \url{ibelousova@apple.com}}
\metadata[Venue]{\sffamily Accepted at NeurIPS 2026}

\begin{document}
\maketitle

% ------------------------------------------------------------
% First-page footnotes:
%   - work-done-at-Apple note for Yizhe
%   - Apple trademark line
% ------------------------------------------------------------
\applefootnote{\textcolor{textgray}{\sffamily%
  $^{\ddagger}$Work done while at Apple.\\
  Apple and the Apple logo are trademarks of Apple Inc., registered in the U.S. and other countries and regions.}}

% ============================================================
% Body sections
% ============================================================

% ============================================================
% INTRODUCTION — Revised for NeurIPS submission
% ============================================================

\section{Introduction}
\label{sec:intro}

Discrete flow matching (DFM) models generate text by iteratively transforming an uninformative source sequence, such as uniform random tokens or mask tokens, into coherent language through hundreds or thousands of refinement steps~\citep{gat2024discreteflow, campbell2024generative, nie2025llada, ye2025dream}. While quality is on par with that of other large language models (LLMs), it is orders of magnitude slower than autoregressive LLMs as a full forward pass is required to generate each token.

Distillation compresses this process: a student model learns to reproduce in a few steps what the iterative process achieves in many (e.g., FS-DFM~\citep{fsdfm2025}). The student never sees training data, but rather learns entirely from \emph{trajectories}, the specific paths through token space that the iterative process traces from a source state to finished text. When the distilled student underperforms, the standard explanation is insufficient model capacity. In this paper, we argue that this explanation ignores the trajectory and that \textit{the trajectory is the bottleneck, not the student.}

Trajectory construction requires converting continuous velocity logits into concrete token sequences at each intermediate step. Because tokens are discrete, this conversion demands a hard stochastic decision at every position. No evaluation of the resulting sequence ever occurs: the process commits to a single random realization and moves on. Each subsequent step then conditions on this potentially degraded state, so a single poor sample propagates forward and corrupts every downstream computation. In FS-DFM's RK-4 estimator~\citep{fsdfm2025}, three such blind jumps are chained; error introduced at the first midpoint shifts the velocity evaluation at the second, which degrades the third, and the final target absorbs the accumulated drift. The student faithfully imitates this drifted target because it is the only supervision it receives.

\begin{figure}[t]
\centering
\includegraphics[width=\textwidth]{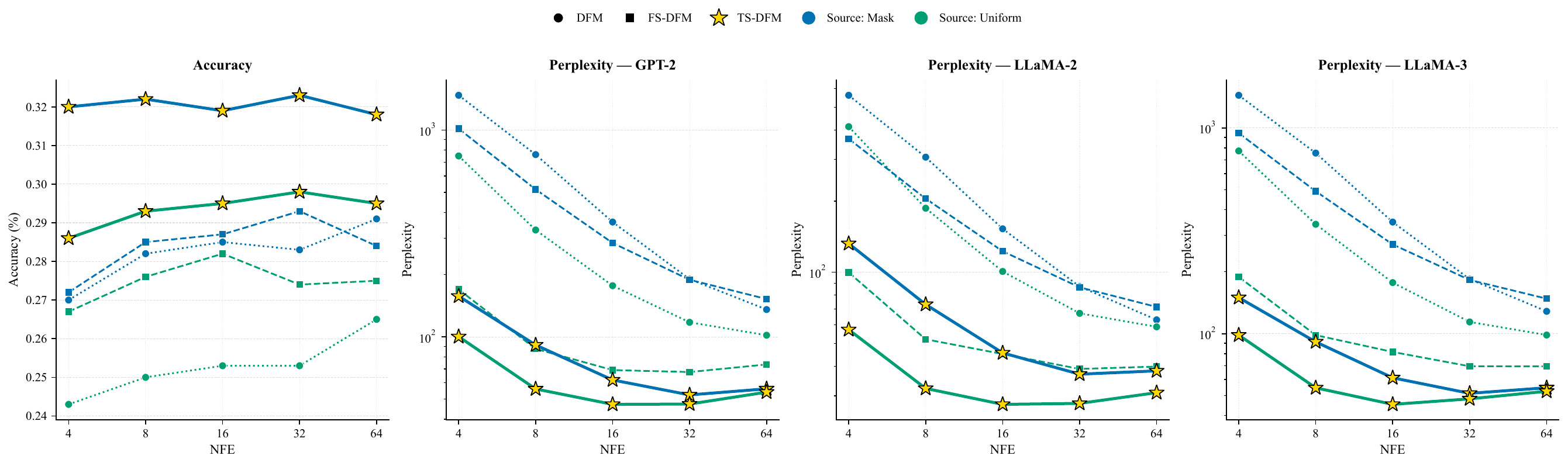}
\caption{\textbf{Generation quality across step budgets, evaluators, and source distributions.}
From left to right: next-token prediction accuracy and perplexity under three evaluators. \ours achieves the lowest perplexity, with the largest gains at low NFE (number of function evaluations) where few-step generation matters most.
}

\label{fig:teaser_result}
\end{figure}

\begin{figure}[t]
\centering
\begin{tikzpicture}[scale=0.87]

  % ---- Background regions (wider) ----
  \fill[noisezone, rounded corners=4pt] (-0.3, -0.8) rectangle (3.2, 5.8);
  \fill[guidezone, rounded corners=4pt] (3.2, -0.8) rectangle (13.0, 5.8);
  \node[font=\scriptsize\bfseries, text=orange!70!black, align=center] at (1.45, 6.25)
    {Compass Inactive\\[-2pt]\scriptsize(Source-dominated)};
  \node[font=\scriptsize\bfseries, text=blue!70!black, align=center] at (8.1, 6.25)
    {Navigation Active\\[-2pt]\scriptsize(Signal-rich)};
  \draw[densely dashed, thick, red!60!black] (3.2, -0.8) -- (3.2, 5.8);
  \node[font=\scriptsize\bfseries, text=red!60!black, rotate=90, anchor=south]
    at (3.15, 4.8) {$t = \tau$};

  % ---- Axes (wider x-axis, split label) ----
  \draw[-{Stealth[length=3mm]}, thick] (0, 0) -- (13.2, 0);
  \node[font=\small, align=center] at (6.6, -0.45) {Generation progress};
  \draw[-{Stealth[length=3mm]}, thick] (0, 0) -- (0, 5.4);
  \node[font=\small, rotate=90, anchor=south] at (-0.35, 2.7) {Sequence quality};
  \foreach \x/\l in {0/{source}, 6.3/{}, 12.6/{text}} {
    \node[font=\tiny, below] at (\x, -0.12) {\l};
  }

  % ---- Ideal trajectory (faint dotted, wider) ----
  \draw[thick, black!50, dotted, smooth]
    plot coordinates {(0,0.4) (1.26,0.7) (2.52,1.15) (3.78,1.8) (5.04,2.55)
      (6.3,3.25) (7.56,3.9) (8.82,4.4) (10.08,4.8) (11.34,5.05) (12.6,5.2)};
  \node[font=\tiny, black!55] at (12.2, 5.5) {Ideal};

  % ---- 1) Unshaped trajectory (gray dashed) ----
  \draw[thick, baselinegray, densely dashed, smooth]
    plot coordinates {(0,0.3) (1.26,0.5) (2.52,0.85) (3.78,1.3)
      (4.6,1.45) (5.04,1.2) (5.7,1.55) (6.3,1.35)
      (7.1,2.05) (7.56,1.75) (8.25,2.35) (8.82,2.1)
      (9.5,2.7) (10.08,2.5) (10.9,2.95) (12.6,3.1)};
  \draw[thick, trajbad, -{Stealth[length=2mm]}] (3.77, 1.5) -- (5.04, 1.2);
  \node[font=\tiny, trajbad, below] at (5.04, 0.95) {blind jump};
  \draw[{Stealth[length=1.3mm]}-{Stealth[length=1.3mm]}, trajbad, thick]
    (8.25, 2.35) -- (8.25, 3.3);
  \node[font=\tiny, trajbad, fill=white, inner sep=1pt, align=center]
    at (8.25, 2.84) {\scriptsize errors\\[-2pt]\scriptsize compound};
  \node[font=\scriptsize, baselinegray, fill=white, inner sep=2pt,
        rounded corners=2pt] at (11.5, 3.45) {Standard trajectory};

  % ---- 2) Navigation-shaped trajectory (blue) ----
  \draw[very thick, energyblue!60, smooth]
    plot coordinates {(0,0.3) (1.26,0.55) (2.52,0.95) (3.78,1.5)
      (5.04,2.1) (6.3,2.65) (7.56,3.2) (8.82,3.65) (10.08,4.0) (11.34,4.25) (12.6,4.45)};
  \foreach \x/\y in {3.78/1.5, 5.04/2.1, 6.3/2.65, 7.56/3.2, 8.82/3.65, 10.08/4.0} {
    \fill[energyblue] (\x, \y) circle (2pt);
  }
  % Candidate fan at one midpoint
  \draw[-{Stealth[length=1.6mm]}, energyblue!90, thick] (3.77, 1.5) -- (5.04, 2.1);
  \draw[-{Stealth[length=1.6mm]}, gray!90, thin] (3.77, 1.5) -- (4.98, 1.75);
  \draw[-{Stealth[length=1.6mm]}, gray!90, thin] (3.77, 1.5) -- (4.98, 1.5);
  % Energy compass label
  \node[font=\tiny, energyblue, fill=white, draw=energyblue, rounded corners=2pt,
        inner sep=2pt, text width=1.4cm, align=center] at (4.0, 3.2)
    {Energy compass\\[-1pt]picks best};
  \draw[-{Stealth[length=1.5mm]}, energyblue, thin] (4.3, 2.75) -- (4.7, 2.0);
  \node[font=\scriptsize\bfseries, energyblue, fill=white, inner sep=2pt,
        rounded corners=2pt] at (11.5, 4.75) {Shaped trajectory (ours)};

  % ---- Shaping gap annotation ----
  \draw[{Stealth[length=1.5mm]}-{Stealth[length=1.5mm]}, trajbad, very thick]
    (5.95, 1.45) -- (5.95, 2.55);
  \node[font=\tiny\bfseries, trajbad, fill=white, inner sep=1.5pt, right]
    at (6.0, 2.1) {shaping gap};

\end{tikzpicture}
\caption{During training, standard trajectory construction compounds errors through blind stochastic jumps. Navigation shaping selects among candidates via an energy compass, activating at $t \geq \tau$.}
\label{fig:teaser}
\end{figure}
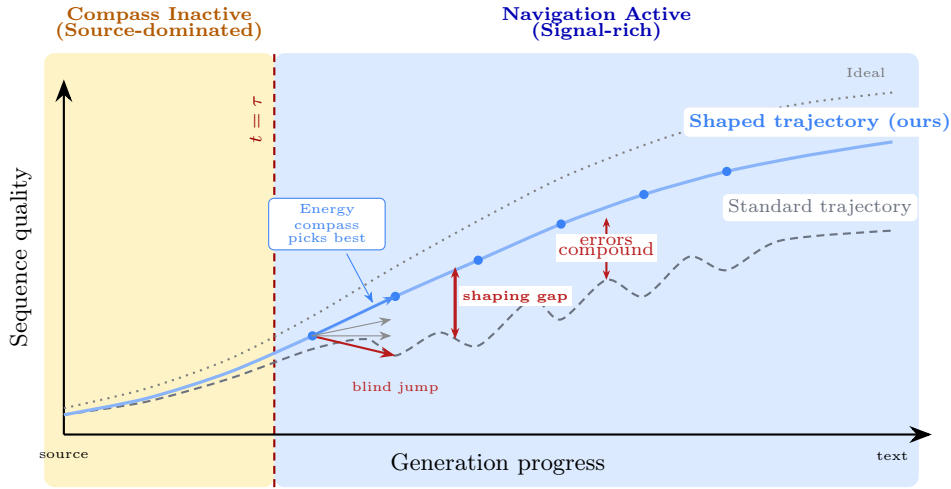

We introduce Trajectory-Shaped Discrete Flow Matching (\ours), a framework that shapes training trajectories to maximize their teaching value via a mechanism we call \emph{navigation shaping}. At each midpoint where the standard method makes a single blind stochastic jump, we generate multiple candidate continuations and use a lightweight \emph{energy compass}---a scalar energy model trained on generation-aware negatives---to select the highest-quality one. Empirically, the compass achieves 98.5\% accuracy at distinguishing real flow states from generation-relevant corruptions. A \emph{Sequence-to-Token} policy first picks the best candidate globally, then refines it at the token level using velocity confidence as a free proxy. A time threshold $\tau$ completes the design: at early flow times the sequence is too uninformative for the compass to discriminate among candidates, so navigation activates only once sufficient linguistic structure has emerged (\cref{fig:teaser}).

On 170M-parameter language modeling, \ours at 8 steps achieves 56.1 GPT-2 perplexity with a uniform source---a 36\% reduction over FS-DFM and 32\% below the full, unguided $1\,024$-step teacher, while being $128\times$ faster (\cref{fig:teaser_result}). For the mask source, improvements are even larger ($5.6\times$ over FS-DFM at 8 steps). \ours outperforms Duo~\citep{sahoo2025diffusion} and SDTT~\citep{deschenaux2024beyond}, achieving state-of-the-art perplexity among few-step discrete generation methods. Improvements hold across three evaluators of increasing scale (GPT-2, LLaMA-2, LLaMA-3). All shaping is training-only, adding $2.0$--$2.4\times$ one-time overhead; inference is unchanged.

To our knowledge, this is the first work to treat the intermediate states of a discrete flow trajectory---not just the endpoints---as a target for quality optimization during distillation. Navigation shaping is orthogonal to the distillation mechanism itself. We show it within the FS-DFM distiller, and in principle it can be applied on top of other trajectory-based distillation frameworks. Our contributions are:

\begin{itemize}
\item \textbf{Trajectory quality as the distillation bottleneck}: We show that the performance ceiling in few-step DFM is set by trajectory quality, not student capacity. Blind stochastic jumps at RK-4 midpoints introduce compounding errors that degrade supervision; optimizing these midpoint states breaks through the ceiling.

\item \textbf{Sequence-to-Token navigation shaping}: A two-phase policy that selects the best among $K$ candidate continuations at the sequence level, then refines it at the token level, gated by a time threshold $\tau$.

\item \textbf{Generation-aware energy compass}: A lightweight energy model trained with velocity near-miss negatives that achieves 98.5\% corruption-detection accuracy on partially-revealed flow states.

\item \textbf{No inference cost}: All shaping is training-only; inference remains unchanged.
\end{itemize}
\section{Related work}
\label{sec:related}
Discrete diffusion~\citep{austin2021structured, he2022diffusionbert, xie2025dream, song2025seed, arriola2025block, yang2025mmada, ye2024diffusion} has been explored for text generation in models like LLaDA~\citep{nie2025llada} and Dream~\citep{ye2025dream}. These generate text via iterative refinement from a source distribution, but quality degrades sharply when the number of sampling steps is reduced.

A natural remedy is distillation: in the continuous setting, progressive distillation~\citep{salimans2022progressive}, consistency models~\citep{song2023consistency}, and single-step methods~\citep{meng2023distillation, frans2025one} have all been used to compress many-step samplers into few-step students.

In the discrete setting, SDTT~\citep{deschenaux2024beyond} distills masked diffusion models (MDLM~\citep{sahoo2024simple}, SEDD~\citep{lou2023discrete}) through iterative self-distillation rounds, achieving strong few-step quality but requiring multiple rounds of training and a large base model. ReMDM~\citep{wang2025remasking} improves few-step masked diffusion at inference time through remasking strategies (cap, conf, rescale, loop), avoiding retraining but adding per-step overhead. Duo~\citep{sahoo2025diffusion} establishes a duality between Gaussian and uniform-state discrete diffusion, enabling consistency distillation in the discrete setting and achieving competitive few-step generation, though at the cost of substantially more training data. FS-DFM~\citep{fsdfm2025} brought distillation to discrete flow matching with step-size conditioning and RK-4 trajectory estimation. T3D~\citep{zhang2026t3dfewstepdiffusionlanguage} concurrently proposes trajectory self-distillation with a reverse-KL objective that encourages mode-seeking behavior, addressing the few-step gap through changes to the distillation loss rather than the trajectory construction.

This paper builds on FS-DFM's distillation mechanism but addresses an orthogonal question: not \emph{how} to distill, but \emph{what the student learns from}. That the quality of the training signal matters for distillation is well established---teacher capacity and temperature shape student performance in knowledge distillation~\citep{hinton2015distilling}, and trajectory sampling strategies influence outcomes in diffusion distillation~\citep{yin2024one}---yet prior work on discrete sequences has not attempted to optimize the trajectories themselves.
Unlike SDTT~\citep{deschenaux2024beyond}, which improves the student through multiple distillation rounds on the same trajectories, and ReMDM~\citep{wang2025remasking}, which corrects errors at inference time, \ours intervenes at the source: shaping the trajectories during training so that the student receives higher-quality supervision from the start.

To shape trajectories, we introduce an energy model that scores partially revealed flow states and selects the highest-quality candidate at each step. Energy-based models have a long history in generative modeling~\citep{lecun2006tutorial, song2021train}: in continuous diffusion they underlie classifier-free and classifier-based guidance~\citep{dhariwal2021diffusion, du2023reduce, du2024compositional, bansal2023universal}, while in the text domain they have been used for reranking and controlled generation of complete sequences~\citep{deng2020residual, khalifa2021distributional}. Our use differs in three respects: (i)~the energy model operates on \emph{partial} states---mixtures of meaningful and source tokens---rather than complete sequences or vectors; (ii)~it acts as a discrete selection criterion rather than providing gradients; and (iii)~it participates exclusively at training time, incurring no inference cost.
% ============================================================
% SYMBOL CONVENTION (same as intro — keep consistent)
% v_\theta          : pretrained DFM teacher (frozen)
% \tilde{v}_\omega  : semi-teacher (EMA of student)
% s_\psi            : student model
% E_\phi            : energy compass (frozen at training time)
% k_i               : RK-4 stage velocities (standard / blind)
% \hat{k}_i         : RK-4 stage velocities (navigated)
% v_{\text{RK-4}}   : standard RK-4 target (no hat = blind)
% \hat{v}_{\text{RK-4}} : navigated RK-4 target (hat = shaped)
% x_0               : source state (uniform, mask)
% x_1               : target text
% x_t               : interpolated state at flow time t
% \alpha(t)         : fraction of revealed positions at time t
% h                 : total step size for the student
% h'                : sub-step size used within RK-4 stages
% ============================================================

\section{Background and motivation}
\label{sec:background}

\label{sec:dfm}
Discrete flow matching (DFM)~\citep{gat2024discreteflow, campbell2024generative} generates text by traversing a trajectory from a source state to text. A velocity field $v_\theta$ iteratively transforms a source sequence $x_0$ into coherent text $x_1$ over $N$ steps, where $x_0$ is drawn from a source distribution (uniform over the vocabulary, or mask tokens). At intermediate time $t \in [0,1]$, the state $x_t$ is a position-wise mixture of source and target tokens: each position $j$ independently holds either the target or the source token,
\begin{equation}
  x_t[j] = \begin{cases}
    x_1[j] & \text{with probability } \alpha(t), \\
    x_0[j] & \text{with probability } 1 - \alpha(t),
  \end{cases}
  \label{eq:interpolation}
\end{equation}
where $\alpha(t)$ is a monotonically increasing schedule with $\alpha(0) = 0$ and $\alpha(1) = 1$. This interpolation defines both the forward sampling path during training and the partially formed sequences that the energy compass must evaluate.

\label{sec:fsdfm}

The \textbf{Few-Step Discrete Flow-Matching} framework~\citep{fsdfm2025} compresses long teacher trajectories into few-step student ones using three models.
The \textbf{teacher} $v_\theta(x_t, t)$ is a frozen pretrained DFM velocity model.
The \textbf{student} $s_\psi(x_t, t, h)$ shares the teacher's architecture with an additional step-size conditioning mechanism and is initialized from the teacher.
The \textbf{semi-teacher} $\tilde{v}_\omega(x_t, t, h)$ is an exponential moving average (EMA) of the student's weights. For small step sizes, the teacher's single-step velocity is used directly; for large step sizes, the semi-teacher estimates the velocity via fourth-order Runge--Kutta (RK-4) integration, constructing multi-step targets that the teacher alone cannot provide in a single evaluation:
\begin{equation}
  k_1 = \tilde{v}_\omega\!\big(x_t,\, t,\, \tfrac{h}{2}\big),\qquad
  k_{i+1} = \tilde{v}_\omega\!\big(x_{\text{mid},i},\, t_{\text{mid},i},\, \tfrac{h}{2}\big),\;\; i=1,2,3,
  \label{eq:rk4_stages}
\end{equation}
where $t_{\text{mid},1} = t_{\text{mid},2} = t + \tfrac{h}{2}$, $\;t_{\text{mid},3} = t + h$, and each midpoint $x_{\text{mid},i}$ is a concrete token sequence constructed from the previous stage's velocity via a stochastic continuous-time Markov chain (CTMC) jump. The final RK-4 target $v_{\text{RK-4}} = \tfrac{1}{6}(k_1 + 2k_2 + 2k_3 + k_4)$ is the standard weighted combination, and the student minimizes $\mathcal{L}_{\text{KL}} = \mathrm{KL}(s_\psi(x_t, t, h) \| v_{\text{RK-4}})$.
The student's \emph{entire} training signal comes from these RK-4 targets: the semi-teacher's constructed trajectory at each training step \emph{is} the lesson. The student has no independent access to data, no alternative examples, no error signal beyond ``match this target.''

\textbf{Why the trajectory is a poor teacher.}
\label{sec:poor_teacher}
To construct each midpoint $x_{\text{mid},i}$ in the RK-4 chain, velocity logits must be converted into a concrete token sequence. Since tokens are discrete, this requires a stochastic jump: a Poisson-rate process determines whether each position $j$ transitions, and a categorical sample chooses the new token. Concretely, given the current state $x$ and velocity logits $k$ at time $t$ with sub-step size $h'$:
\begin{equation}
  x_{\text{mid}}[j] = \begin{cases}
    \hat{x}_1[j] \sim \mathrm{Cat}\!\big(k[j]\big) & \text{with prob. } p_{\text{jump}}(j), \\
    x[j] & \text{otherwise},
  \end{cases}
  \label{eq:ctmc_jump}
\end{equation}
where $\hat{x}_1[j]$ is sampled from the velocity's categorical distribution at position $j$, and $p_{\text{jump}}(j)$ is the transition probability derived from the CTMC rate and sub-step size $h'$. This procedure is applied at each of the three internal RK-4 stages. Each application produces one specific realization \emph{without any quality evaluation}, and the problem compounds through the chain. A poor first sample affects the velocity evaluation for subsequent steps, each compounding the degradation:
\begin{equation}
  k_1 \;\xrightarrow{\text{sample}}\; \underbrace{x_{\text{mid},1}}_{\text{degraded}} \;\to\; k_2 \;\xrightarrow{\text{sample}}\; \underbrace{x_{\text{mid},2}}_{\text{worse}} \;\to\; k_3 \;\xrightarrow{\text{sample}}\; \underbrace{x_{\text{mid},3}}_{\text{worst}} \;\to\; k_4.
  \label{eq:error_compound}
\end{equation}
The final target $v_{\text{RK-4}} = \frac{1}{6}(k_1 + 2k_2 + 2k_3 + k_4)$ absorbs compounded errors from all midpoints, leading to trajectory drift (\cref{fig:teaser}). Because the student learns to imitate the drifted trajectory, it cannot recover the quality lost during trajectory construction, and learns the compounded errors.

\begin{figure}[t]
\centering
\begin{tikzpicture}[
  >=Stealth, font=\sffamily\scriptsize,
  scale=0.78, transform shape,
  mbox/.style={rectangle, rounded corners=5pt, minimum height=1.25cm,
    line width=1pt},
  arr/.style={-{Stealth[length=4pt,width=3.5pt]}, line width=0.9pt, color=#1},
  darr/.style={-{Stealth[length=4pt,width=3.5pt]}, line width=0.8pt, color=#1, dashed},
  tarr/.style={-{Stealth[length=5pt,width=4pt]}, line width=1.1pt, color=#1},
]

% ═══════════════════════════════════
%  ROW 1: Student + Semi-teacher
% ═══════════════════════════════════

\node[mbox, minimum width=2.7cm, draw=studC, fill=studF, text=studC,
  font=\sffamily\bfseries\small] (student) at (0, 3.6)
  {\begin{tabular}{c} Student $s_\psi$\\[-0pt]
   {\small $(x_t, t, h)$} \end{tabular}};
\node[font=\small, text=studC!70, anchor=north east]
  at ([xshift=-2pt, yshift=-2pt]student.north east) {\faFire};

\node[mbox, minimum width=3.0cm, draw=emaC, fill=emaF, text=emaC,
  font=\sffamily\bfseries\small] (ema) at (5.5, 3.6)
  {\begin{tabular}{c} Semi-teacher $\tilde{v}_\omega$\\[-0pt]
   {\tiny (EMA of student)} \end{tabular}};
\node[font=\small, text={rgb,255:red,100;green,180;blue,255}, anchor=north east]
  at ([xshift=1pt, yshift=-2pt]ema.north east) {\faSnowflake};

% EMA update arrow
\draw[tarr=studC] (student.east) --
  node[above, font=\sffamily\scriptsize\bfseries, text=grayT] {EMA update}
  node[below, font=\sffamily\scriptsize, text=emaC] {$\omega \!\leftarrow\! \mu\omega + (1{-}\mu)\psi$}
  (ema.west);

% ═══════════════════════════════════
%  ROW 2: Teacher + Loss + Nav RK-4 + Compass
% ═══════════════════════════════════

\node[mbox, minimum width=2.5cm, draw=teachC, fill=teachF!50, text=teachC,
  font=\sffamily\bfseries\small] (teacher) at (-2.8, 1.0)
  {\begin{tabular}{c} Teacher $v_\theta$\\[-0pt]
   {\small $(x_t, t)$} \end{tabular}};
\node[font=\small, text={rgb,255:red,100;green,180;blue,255}, anchor=north east]
  at ([xshift=-2pt, yshift=-2pt]teacher.north east) {\faSnowflake};

\node[circle, minimum size=1.05cm, draw=lossC, line width=1pt, fill=lossC!8,
  font=\sffamily\bfseries\small, text=lossC]
  (loss) at (0, 1.0) {$\mathcal{L}$};

\node[mbox, minimum width=2.9cm, draw=navC, fill=navF, text=navC,
  font=\sffamily\bfseries\small] (navrk) at (5.5, 1.0)
  {\begin{tabular}{c} Navigated RK-4\\[-0pt]
   {\small $\hat{v}_{\text{RK-4}}$}
   \end{tabular}};

\node[mbox, minimum width=2.7cm, draw=compassC, fill=compassF, text=compassC,
  font=\sffamily\bfseries\small] (compass) at (9.8, 1.0)
  {\begin{tabular}{c} Compass $E_\phi$\\[1pt]
   {\small $\mathcal{V}^L\!\to\!\mathbb{R}$}
   \end{tabular}};
\node[font=\small, text={rgb,255:red,100;green,180;blue,255}, anchor=north east]
  at ([xshift=-1pt, yshift=-1pt]compass.north east) {\faSnowflake};

% Threshold badge (below compass)
\node[rectangle, rounded corners=2pt, draw=threshC, fill=threshF,
  line width=0.6pt, font=\sffamily\tiny\bfseries, text=threshC,
  inner sep=2pt] (thresh) at (9.8, 0.1) {active when $t \!\geq\! \tau$};

% ═══════════════════════════════════
%  ARROWS (all straight)
% ═══════════════════════════════════

% Teacher → Loss
\draw[arr=teachC] (teacher.east) --
  node[above, font=\sffamily\tiny, text=teachC] {$h\!<\!\tau_h$}
  (loss.west);

% Nav RK-4 → Loss
\draw[arr=navC] (navrk.west) --
  node[above, font=\sffamily\tiny, text=navC] {$h\!\geq\!\tau_h$}
  (loss.east);

% Student → Loss
\draw[arr=studC] (student.south) -- (loss.north);

% Loss → Student (gradient)
\draw[darr=navy]
  ([xshift=-0.35cm]loss.north) --
  node[left, font=\sffamily\tiny\bfseries, text=navy] {$\nabla_\psi$}
  ([xshift=-0.35cm]student.south);

% EMA → Nav RK-4
\draw[arr=emaC] (ema.south) --
  node[right, font=\sffamily\tiny, text=emaC, xshift=1pt] {$\hat{k}_i$}
  (navrk.north);

% Compass ↔ Nav RK-4 (bidirectional)
\draw[{Stealth[length=4pt,width=3.5pt]}-{Stealth[length=4pt,width=3.5pt]},
  line width=1pt, color=compassC] (compass.west) --
  node[above, font=\sffamily\tiny, text=compassC] {\textsc{Navigate}}
  (navrk.east);

% ═══════════════════════════════════
%  "New in TSDFM" highlight
% ═══════════════════════════════════
\draw[rounded corners=3pt, densely dashed, compassC!40, line width=0.8pt]
  ([xshift=-0.3cm, yshift=0.2cm]navrk.north west)
  rectangle
  ([xshift=0.3cm, yshift=-0.5cm]compass.south east);
\node[font=\sffamily\small\bfseries, text=compassC!60, anchor=north east]
  at ([xshift=0.3cm, yshift=0.7cm]compass.north east)
  {New in \textsc{TS-DFM}};

\end{tikzpicture}
\caption{\textbf{\ours{} training overview.}
For small steps, the frozen teacher provides the target directly.
For large steps, the semi-teacher constructs targets via RK-4 with each midpoint guided by \textcolor{compassC}{\textbf{energy compass}} $E_\phi$.
\ours{}-specific components are in the dashed box; the rest inherits from FS-DFM.}
\label{fig:tsdfm-overview}
\end{figure}
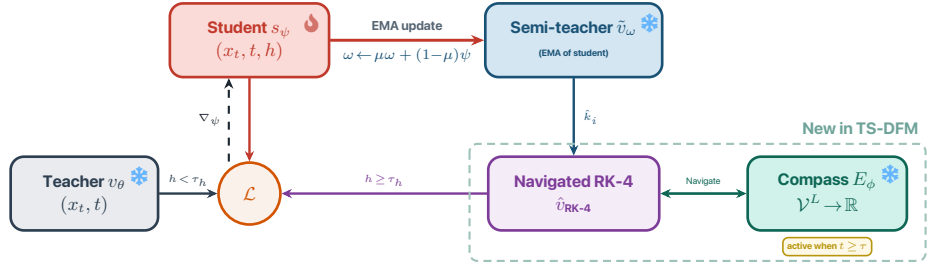

\section{Method: navigation shaping}
\label{sec:method}

We propose Trajectory-Shaped Discrete Flow Matching~(\ours{}), a framework that builds on FS-DFM but replaces the blind stochastic jumps with guided navigation. \cref{fig:tsdfm-overview} illustrates \ours{}'s training pipeline and the three new components: an energy compass, a Sequence-to-Token navigation procedure, and a time threshold that controls when guidance activates.

\subsection{The energy compass}
\label{sec:energy_model}

The core tool for navigation is a scalar energy model $E_\phi(x) : \mathcal{V}^L \to \mathbb{R}$ that evaluates the quality of a token sequence without time conditioning---it scores the raw token sequence only. Sequence quality is an intrinsic property of the content. A partially revealed sequence with coherent tokens should score well regardless of whether it appears at $t{=}0.3$ or $t{=}0.7$.

The compass must distinguish subtle quality differences between partially revealed flow states rather than scoring clean text. The standard diffusion corruption strategy of random token replacement produces trivial negatives that do not represent the decisions that appear during trajectory navigation. To obtain an effective energy compass $E_\phi$, we train with a noise-contrastive objective over \emph{generation-aware negatives} to assign lower energy to real flow states.

We use \emph{velocity near-miss} to construct negative examples. Given a positive $x_t$, we step backward along the flow to create $x_{t-h}$, run the frozen teacher velocity model forward to reconstruct $\hat{x}_t$, and use $\hat{x}_t$ as a negative. The result competes directly with the positive---both are plausible sequences at time $t$---but $\hat{x}_t$ contains the teacher's reconstruction errors at the unrevealed positions. These are the kind of subtle errors the compass must detect when comparing CTMC jump candidates during trajectory construction.
The total training loss is:
\begin{equation}
  \mathcal{L}_{\text{energy}} = \mathcal{L}_{\text{NCE}} + \lambda_{\text{reg}} \cdot \mathcal{L}_{\text{reg}} + \lambda_{\text{order}} \cdot \mathcal{L}_{\text{order}},
  \qquad \lambda_{\text{reg}} = \lambda_{\text{order}} = 1.0
  \label{eq:energy_total_loss}
\end{equation}
where $\mathcal{L}_{\text{NCE}}$ trains the compass to assign lower energy to real flow states than corruptions, $\mathcal{L}_{\text{reg}}$ penalizes energy-scale drift, and $\mathcal{L}_{\text{order}}$ enforces the time-ordering constraint. \cref{app:energy_training} contains the complete training details, additional negative strategies, and a time-ordering loss that enforces monotonically decreasing energy as $t$ increases for each sample $x_t$.
On held-out FineWeb-Edu, the trained compass distinguishes real flow states from generation-relevant corruptions with $98.5\%$--$99.8\%$ accuracy and exhibits perfectly monotone bin-level energy as $t$ increases, for both source distributions (\cref{app:energy_compass_validation}).

\subsection{Sequence-to-Token navigation}
\label{sec:guided_midpoint}

The compass provides \emph{sequence-level} scores, but trajectory quality depends on \emph{token-level} decisions. Scoring entire candidates is affordable but coarse---a sequence can score well while containing poor tokens. Factorizing guidance to individual positions would require $|\mathcal{V}| \times L$ energy evaluations per midpoint, which is intractable. We address this with \textsc{Navigate}, a guided two-phase approach:

\begin{enumerate}
  \item \textbf{Sequence phase} (sequence-level). Generate $K$ candidate next states via temperature-diverse CTMC transitions from the current state. Evaluate them in a single batched forward pass through $E_\phi$ and select the lowest-energy candidate. This ensures the trajectory enters a globally low-energy region of sequence space.

  \item \textbf{Token phase} (token-level). Refine the selected candidate at individual positions using the velocity network's own confidence as a \emph{free} proxy for per-token quality---the logits are already computed from the RK-4 velocity evaluation. Positions where the selected token disagrees with a high-confidence velocity prediction are \emph{candidates} for replacement, modulated by an adaptive guidance coefficient that decays with $t$. A single energy evaluation on the refined result serves as a safeguard: the refinement is accepted only if it does not increase energy beyond a tolerance $\epsilon_{\text{safe}}$.
\end{enumerate}

The sequence phase ensures global trajectory coherence. The token phase ensures local token-level accuracy. Together they cover both granularities without requiring per-token energy evaluation. One cannot do without the other: sequence selection cannot fix individual bad tokens hidden in a good sequence, and token refinement cannot recover from selecting the wrong candidate in the first place. Candidate diversity is sufficient for selection to matter: at $t{=}\tau{=}0.2$, $K$ independent CTMC jumps from a shared midpoint differ at ${\sim}147$ of $1\,024$ positions ($14\%$) on average (\cref{app:midpoint_diversity}). The safeguard accepts ${\geq}94.7\%$ of refinements, and rejections would on average have raised energy (\cref{app:refinement_behavior}). An inference-time ablation confirms complementarity: at 64 steps the full policy ($38.0$ PPL) beats sequence-only ($52.0$) and token-only ($47.6$) by ${\geq}20\%$ (\cref{app:policy_comparison}). \cref{app:c2f_navigation,app:energy_validation} contain complete details and ablations.

Replacing each blind jump with \textsc{Navigate} transforms the standard RK-4 estimation into a navigated one. The semi-teacher $\tilde{v}_\omega$ computes velocity logits as before; the only change is how midpoint states are constructed from those logits. We denote navigated velocities with hats ($\hat{k}_i$) and quality-checked midpoints with asterisks ($x_{\text{mid},i}^*$) to distinguish them from the blind equivalents:
\begin{align}
  \hat{k}_1 &= \tilde{v}_\omega(x_t,\, t,\, \tfrac{h}{2}), & x_{\text{mid},1}^* &= \textsc{Navigate}(x_t,\, \hat{k}_1,\, t,\, \tfrac{h}{2},\, E_\phi), \label{eq:nav_k1}\\[3pt]
  \hat{k}_2 &= \tilde{v}_\omega(x_{\text{mid},1}^*,\, t{+}\tfrac{h}{2},\, \tfrac{h}{2}), & x_{\text{mid},2}^* &= \textsc{Navigate}(x_{\text{mid},1}^*,\, \hat{k}_2,\, t{+}\tfrac{h}{2},\, \tfrac{h}{2},\, E_\phi), \label{eq:nav_k2}\\[3pt]
  \hat{k}_3 &= \tilde{v}_\omega(x_{\text{mid},2}^*,\, t{+}\tfrac{h}{2},\, \tfrac{h}{2}), & x_{\text{mid},3}^* &= \textsc{Navigate}(x_{\text{mid},2}^*,\, \hat{k}_3,\, t{+}\tfrac{h}{2},\, \tfrac{h}{2},\, E_\phi), \label{eq:nav_k3}\\[3pt]
  \hat{k}_4 &= \tilde{v}_\omega(x_{\text{mid},3}^*,\, t{+}h,\, \tfrac{h}{2}), & \hat{v}_{\text{RK-4}} &= \tfrac{1}{6}(\hat{k}_1 + 2\hat{k}_2 + 2\hat{k}_3 + \hat{k}_4). \label{eq:nav_rk4}
\end{align}
Compared with the standard chain where blind jumps produced midpoints
with compounding errors, navigated jumps produce midpoints
with quality-checked transitions. The advantage compounds multiplicatively: if each guided jump reduces error by a factor $\rho < 1$ relative to a blind jump, the overall target improves by ${\sim}\rho^3$ across the three midpoints.

\subsection{Time-threshold activation}
\label{sec:threshold}

The compass needs signal to navigate. At low flow time $t$, only a small fraction $\alpha(t)$ of positions carry meaningful tokens---the sequence lies far from the manifold of natural language, and all candidates appear similarly uninformative. Activating the compass here would add computational cost without benefit, and could introduce bias by selecting on noise. We therefore activate navigation only above a time threshold $\tau$:
\begin{equation}
  x_{\text{mid}}' = \begin{cases}
    \textsc{Jump}(x_t, k, t, h) & \text{if } t < \tau \quad \text{(standard blind jump)}, \\
    \textsc{Navigate}(x_t, k, t, h, E_\phi) & \text{if } t \geq \tau \quad \text{(compass-guided selection)}.
  \end{cases}
\label{eq:threshold}
\end{equation}
In the source-dominated regime ($t < \tau$), the trajectory explores freely. In the signal-rich regime ($t \geq \tau$), linguistic structure---local coherence, grammatical dependencies, topical consistency---has emerged sufficiently for the compass to make informed selections, and navigation takes over.

Within a single RK-4 estimation, different midpoints may cross the threshold naturally. For example, with $t{=}0.3$, $h{=}0.25$, $\tau{=}0.5$: the first midpoint at $t{=}0.3$ is unguided, while midpoints at $t_{\text{mid}}{=}0.55$ receive navigation. The trajectory automatically transitions from exploration to guided navigation as the sequence approaches the data manifold. As shown in \cref{fig:tsdfm-overview}, the student's objective is unchanged from FS-DFM distillation---the same KL divergence, but against the navigated target: $\mathcal{L} = \mathrm{KL}(s_\psi(x_t, t, h) \| \hat{v}_{\text{RK-4}})$,
where $\hat{v}_{\text{RK-4}}$ comes from \cref{eq:nav_rk4}. The student model, 
training loop, and inference procedure are identical to FS-DFM; the only 
modification is how midpoint states are constructed during RK-4 target 
estimation. At inference, the compass is not used---all quality improvements 
are encoded in the student's weights.

\section{Experiments}
\label{sec:experiments}

\begin{table}[ht]
	\centering
	\caption{\textbf{Main results across source distributions} for $1\,024$-token unconditional generation. \colorbox{beatteacher}{\textbf{Green}} cells surpass the  $1\,024$-step DFM teacher. Sampling variance at fixed checkpoints (5 decoding seeds) is in \cref{app:seed_variance}.}
	\label{tab:main}
	\vspace{2pt}
	\resizebox{\textwidth}{!}{%
		\setlength{\tabcolsep}{3.0pt}
		\renewcommand{\arraystretch}{1.10}
		\begin{tabular}{@{}r| rrrr| rrrr |rrrr| rrrr@{}}
			%\toprule
			& \multicolumn{4}{c}{DFM (Teacher)} 
			& \multicolumn{4}{c}{FS-DFM}
			& \multicolumn{4}{c}{\textbf{\ours\,(DFM init)}}
			& \multicolumn{4}{c}{\textbf{\ours\,(FS-DFM init)}} \\
			\cmidrule(lr){2-5}\cmidrule(lr){6-9}\cmidrule(lr){10-13}\cmidrule(lr){14-17}
			\textbf{Steps}
			& \footnotesize Ent & \footnotesize G\textsuperscript{2}$\downarrow$ & \footnotesize L\textsuperscript{2}$\downarrow$ & \footnotesize L\textsuperscript{3}$\downarrow$
			& \footnotesize Ent & \footnotesize G\textsuperscript{2}$\downarrow$ & \footnotesize L\textsuperscript{2}$\downarrow$ & \footnotesize L\textsuperscript{3}$\downarrow$
			& \footnotesize Ent & \footnotesize G\textsuperscript{2}$\downarrow$ & \footnotesize L\textsuperscript{2}$\downarrow$ & \footnotesize L\textsuperscript{3}$\downarrow$
			& \footnotesize Ent & \footnotesize G\textsuperscript{2}$\downarrow$ & \footnotesize L\textsuperscript{2}$\downarrow$ & \footnotesize L\textsuperscript{3}$\downarrow$ \\
			\midrule
			
			% ==== UNIFORM ====
			\multicolumn{17}{l}{\textit{Uniform source}} \\
			4    & 8.1 & 749.9 & 413.3 & 772.9
			& 7.0 & 169.7 & 99.8  & 188.9
			& 7.1 & 121.2 & 73.4  & 120.6
			& 7.0 & \textbf{100.3} & \textbf{57.1} & \textbf{98.5} \\
			8    & 8.1 & 328.9 & 186.7 & 340.4
			& 7.4 & 87.6  & 52.0  & 98.5
			& 7.3 & \bt{79.0}  & \bt{47.7}  & 79.9
			& 7.2 & \textbf{\bt{56.1}} & \textbf{\bt{32.2}} & \textbf{\bt{54.6}} \\
			16   & 8.0 & 176.9 & 100.8 & 177.2
			& 7.4 & \bt{73.2}  & \bt{45.1}  & 81.6
			& 7.6 & \bt{70.1}  & \bt{41.5}  & \bt{69.5}
			& 7.4 & \textbf{\bt{47.2}} & \textbf{\bt{27.1}} & \textbf{\bt{46.4}} \\
			32   & 8.0 & 117.7 & 67.0  & 114.1
			& 7.7 & \bt{67.8}  & \bt{39.0}  & \bt{69.5}
			& 7.8 & \bt{71.0}  & \bt{41.0}  & \bt{69.7}
			& 7.5 & \textbf{\bt{46.7}} & \textbf{\bt{28.5}} & \textbf{\bt{47.4}} \\
			\cellcolor{gray!15}1\,024 & \cellcolor{gray!15}7.8 & \cellcolor{gray!15}82.8 & \cellcolor{gray!15}47.8 & \cellcolor{gray!15}78.9
			& \multicolumn{4}{c}{\cellcolor{gray!15}---}
			& \multicolumn{4}{c}{\cellcolor{gray!15}---}
			& \multicolumn{4}{c}{\cellcolor{gray!15}---} \\
			
			\midrule
			
			% ==== MASK ====
			\multicolumn{17}{l}{\textit{Mask source}} \\
			4    & 8.5 & 1473.3 & 561.2 & 1441.0
			& 8.2 & 1015.7 & 367.1 & 943.5
			& 7.1 & \textbf{157.4} & \textbf{132.2} & \textbf{149.8}
			& 7.7 & 386.7  & 201.2 & 363.4 \\
			8    & 8.4 & 761.3  & 307.0 & 754.9
			& 8.2 & 516.6  & 204.9 & 492.4
			& 7.4 & \textbf{\bt{91.4}}  & \textbf{73.1}  & \textbf{91.3}
			& 7.9 & 188.7  & 94.8  & 177.4 \\
			16   & 8.4 & 359.3  & 152.7 & 348.9
			& 8.2 & 284.7  & 122.7 & 271.6
			& 7.6 & \textbf{\bt{61.9}}  & \textbf{\bt{45.6}}  & \textbf{\bt{61.1}}
			& 8.0 & 109.9  & 55.2  & 101.9 \\
			32   & 8.3 & 190.3  & 87.0  & 184.2
			& 8.2 & 189.0  & 86.3  & 182.5
			& 7.7 & \textbf{\bt{52.4}}  & \textbf{\bt{37.0}}  & \textbf{\bt{51.3}}
			& 8.0 & 91.4   & 45.4  & 84.5 \\
			\cellcolor{gray!15}1\,024 & \cellcolor{gray!15}8.1 & \cellcolor{gray!15}93.6 & \cellcolor{gray!15}45.7 & \cellcolor{gray!15}87.9
			& \multicolumn{4}{c}{\cellcolor{gray!15}---}
			& \multicolumn{4}{c}{\cellcolor{gray!15}---}
			& \multicolumn{4}{c}{\cellcolor{gray!15}---} \\
			
			%\bottomrule
		\end{tabular}%
	}
	
	\vspace{3pt}
	{\raggedright\scriptsize 
		\textbf{Metrics:} Ent\,=\,generation entropy; G\textsuperscript{2}\,=\,GPT-2 Large (124M), L\textsuperscript{2}\,=\,LLaMA-2 (7B), L\textsuperscript{3}\,=\,LLaMA-3 (8B). 
		All few-step methods use the same 170M-parameter student at inference. \colorbox{gray!15}{Shaded $1\,024$-step row} shows the DFM teacher reference.\par}
\end{table}

\subsection{Setup}

\begin{wraptable}{r}{0.58\textwidth}
	\vspace{-\intextsep}
	\centering
	\small
	\caption{\textbf{Effect of time threshold $\tau$.} Lower $\tau$ activates navigation earlier, improving perplexity but risking diversity loss ($^\ast$entropy below 6.5). \colorbox{selectedrow}{Green rows}: default $\tau{=}0.2$.}
	\label{tab:threshold_ablation}
	\setlength{\tabcolsep}{3pt}
	\footnotesize
	\begin{tabular}{@{}ll rrrr rrrr@{}}
		%\toprule
		& & \multicolumn{4}{c}{\textbf{DFM init}} & \multicolumn{4}{c}{\textbf{FS-DFM init}} \\
		\cmidrule(lr){3-6} \cmidrule(lr){7-10}
		& & \multicolumn{2}{c}{8 steps} & \multicolumn{2}{c}{16 steps} & \multicolumn{2}{c}{8 steps} & \multicolumn{2}{c}{16 steps} \\
		\cmidrule(lr){3-4} \cmidrule(lr){5-6} \cmidrule(lr){7-8} \cmidrule(lr){9-10}
		& & Ent & G$^2$$\downarrow$ & Ent & G$^2$$\downarrow$ & Ent & G$^2$$\downarrow$ & Ent & G$^2$$\downarrow$ \\
		\midrule
		\multicolumn{10}{l}{\textit{Uniform source}} \\
		& $\tau = 0.02$ & $6.4^\ast$ & 54.7 & 7.2 & 56.3 & $6.4^\ast$ & 43.5 & 6.8 & \textbf{35.8} \\
		\rowcolor{selectedrow}
		& $\tau = 0.2$  & 7.3 & 79.0 & 7.6 & 70.1 & 7.2 & \textbf{56.1} & 7.4 & 47.2 \\
		& $\tau = 0.4$  & 7.5 & 87.9 & 7.6 & 69.6 & 7.2 & 71.5 & 7.4 & 53.9 \\
		\midrule
		\multicolumn{10}{l}{\textit{Mask source}} \\
		& $\tau = 0.02$ & $6.2^\ast$ & 92.9 & 7.6 & 89.5 & 7.4 & 137.0 & 7.6 & 83.1 \\
		\rowcolor{selectedrow}
		& $\tau = 0.2$  & 7.4 & \textbf{91.4} & 7.6 & \textbf{61.9} & 7.9 & 188.7 & 8.0 & 109.9 \\
		& $\tau = 0.4$  & 7.8 & 139.4 & 7.8 & 78.8 & 8.0 & 254.2 & 8.1 & 118.9 \\
		%\bottomrule
	\end{tabular}
	\vspace{-\intextsep}
\end{wraptable} 

The DFM teacher and FS-DFM/\ours students are 170M-parameter DDiT transformers~\citep{gat2024discreteflow}.
The energy compass $E_\phi$ is a 6-block transformer pretrained on FineWeb-Edu~\citep{penedo2024finewebdatasetsdecantingweb} with noise-contrastive estimation and generation-aware negatives, then frozen throughout student training. Details of the energy compass training and evaluation are in \cref{app:energy_training,app:energy_compass_validation}. We evaluate across two source distributions:
(i)~\textbf{Uniform}: tokens drawn uniformly over the GPT-2 vocabulary ($|\mathcal{V}|{=}50\,257$);
(ii)~\textbf{Mask}: all positions initialized with a dedicated \texttt{[MASK]} token.

All models are trained on FineWeb-Edu~\citep{penedo2024finewebdatasetsdecantingweb}.
Each source distribution has its own DFM teacher (65.5B tokens) and FS-DFM student (13.2B additional fine-tuning tokens), both serving as potential initializations for \ours (6.4B additional shaping tokens).
Default configuration: $K{=}5$ candidates per midpoint, time threshold $\tau{=}0.2$. \cref{app:c2f_cost} discusses the training-time overhead of navigation shaping relative to standard FS-DFM.

We report generation entropy (Ent) and perplexity (PPL, lower is better) evaluated by three reference language models: GPT-2 Large (124M, denoted G\textsuperscript{2})~\citep{radford2019language}, LLaMA-2 7B (denoted L\textsuperscript{2})~\citep{touvron2023llama}, and LLaMA-3 8B (denoted L\textsuperscript{3})~\citep{grattafiori2024llama}.
All evaluations use $1\,024$-token unconditional generation on WikiText-103~\citep{merity2016pointer} at 4, 8, 16, and 32 sampling steps, with the $1\,024$-step DFM teacher as a quality reference.

\subsection{Main results}
\label{sec:main_results}

\cref{tab:main} presents the central result: across both source distributions, trajectory shaping produces a student that substantially outperforms the FS-DFM baseline and surpasses the unguided $1\,024$-step DFM teacher at just 8 inference steps.
 
\textbf{Uniform source.}
With FS-DFM initialization, \ours achieves 56.1 GPT-2 PPL at 8 steps---a 36\% reduction over the FS-DFM baseline (87.6) and 32\% below the $1\,024$-step teacher (82.8).
The improvement is consistent across all three evaluators and persists at 16 and 32 steps, where \ours reaches 47.2 and 46.7, respectively.
Even with DFM initialization (no prior distillation), \ours at 8 steps (79.0) already beats the $1\,024$-step teacher, confirming that the gains come from trajectory shaping rather than the initialization alone.

\textbf{Mask source.}
For the mask source, DFM initialization proves more effective than FS-DFM initialization.
\ours with DFM init achieves 91.4 GPT-2 PPL at 8 steps and 61.9 at 16 steps---both surpassing the $1\,024$-step mask teacher (93.6).
The mask source is notably more challenging for few-step generation: the FS-DFM baseline achieves only 516.6 PPL at 8 steps, making \ours's $5.6\times$ improvement especially notable.
This directly addresses a key limitation of FS-DFM, which reported strong results only for the uniform source and noted that mask-source distillation remained an open problem. \ours demonstrates that trajectory shaping makes few-step distillation effective for the mask source as well.

\begin{table}[t]
\centering
\small
\caption{\textbf{Compass at inference vs.\ at training} (uniform source, 170M, GPT-2 PPL $\downarrow$). Latency is per $1\,024$-token generation at the lowest step count shown. Teacher rows are a separate evaluation with 64 samples per cell. \ours{} rows are 5-seed means. Values therefore differ slightly from \cref{tab:main} (details in \cref{app:guided_teacher,app:inference_controls}).}
\label{tab:compass_deploy}
\setlength{\tabcolsep}{5pt}
\begin{tabular}{@{}llcccr@{}}
\textbf{Sampler} & \textbf{Compass} & \textbf{8 steps} & \textbf{16 steps} & \textbf{$1\,024$ steps} & \textbf{Lat.\ (s)} \\
\midrule
Teacher, blind                              & ---       & ---   & ---   & 83.98 & 18.9 \\
Teacher $+$ \textsc{Navigate} (every step)  & inference & ---   & ---   & 49.03 & 186 \\
\midrule
FS-DFM, blind                               & ---       & 89.30 & 63.24 & ---   & 0.18 \\
FS-DFM $+$ best-of-5 rerank                 & inference & 80.41 & 63.49 & ---   & 0.94 \\
FS-DFM $+$ final-step selection             & inference & 81.36 & 64.67 & ---   & 0.38 \\
FS-DFM $+$ full \textsc{Navigate}           & inference & 80.18 & 57.36 & ---   & 1.27 \\
\midrule
\rowcolor{selectedrow}
\ours{} (FS-DFM init), blind                & training  & $56.97 \pm 1.48$ & $45.73 \pm 1.27$ & --- & 0.18 \\
\end{tabular}
\end{table}

\paragraph{Training-time shaping vs.\ inference-time guidance.}
To test whether the gain comes from trajectory shaping or from the compass itself, \cref{tab:compass_deploy} gives the same frozen compass to fixed samplers at inference. Reranking, final-step selection and full \textsc{Navigate} all leave the unshaped FS-DFM student near 80 PPL at 8 steps, whereas blind \ours{} reaches 56.97 at the same inference cost. Guided at every step, the $1\,024$-step teacher improves from 83.98 to 49.03, yet the blind 16-step student is lower still (45.73), at ${\sim}0.2\%$ of its inference cost. What separates these rows is when the compass acts, not access to it (\cref{app:guided_teacher,app:inference_controls}).

\paragraph{Effect of time threshold $\tau$.}
\label{sec:threshold_ablation}

The time threshold $\tau$ controls when navigation shaping activates. We train \ours students with $\tau \in \{0.02, 0.2, 0.4\}$ for both source distributions, each with two initializations: from the DFM teacher (DFM init) and from the FS-DFM student (FS-DFM init). \cref{tab:threshold_ablation} reports GPT-2 perplexity and generative entropy at 8 and 16 steps. First, $\tau{=}0.2$ offers the best quality--diversity trade-off for both sources: it achieves strong perplexity while maintaining entropy above 7.0. The lowest threshold ($\tau{=}0.02$) yields better raw perplexity but drops entropy below 6.5, indicating diversity collapse. Second, the two sources prefer different initializations: uniform benefits from FS-DFM init (56.1 vs.\ 79.0 at 8 steps with $\tau{=}0.2$), while mask benefits from DFM init (91.4 vs.\ 188.7)---suggesting that the mask-source FS-DFM student has converged to velocity patterns that are harder to redirect with trajectory shaping. We adopt these defaults throughout. Third, the training overhead is modest: $2.0\times$ at $\tau{=}0.4$ to $2.4\times$ at $\tau{=}0.02$, with our default $\tau{=}0.2$ at $2.2\times$. Additional analysis is in \cref{app:c2f_cost,app:threshold_ablation}.

\begin{wraptable}{r}{0.38\textwidth}
	\vspace{-\intextsep}
	\centering
	\small
	\caption{\textbf{Training overhead}.}
	\label{tab:c2f_wallclock}
	\setlength{\tabcolsep}{3pt}
	\footnotesize
	\begin{tabular}{@{}lrr@{}}
		%\toprule
		\textbf{Method} & \textbf{s/step} & \textbf{vs.\ FS-DFM} \\
		\midrule
		FS-DFM              & 1.37 & $1.0\times$ \\
		\ours ($\tau{=}0.4$) & 2.79 & $2.0\times$ \\
		\rowcolor{selectedrow}
		\ours ($\tau{=}0.2$) & 3.07 & $2.2\times$ \\
		\ours ($\tau{=}0.02$)& 3.31 & $2.4\times$ \\
		%\bottomrule
	\end{tabular}
	\vspace{-\intextsep}
\end{wraptable}

\paragraph{Choice of $K$.} The number of candidates $K$ in the sequence phase controls the quality--cost trade-off. In inference-time ablations (\cref{app:k_ablation}), introducing any candidate selection ($K{=}2$) provides the largest single improvement (23\% PPL reduction). Returns diminish quickly: $K{=}5$ reaches 43\% reduction, while $K{=}8$ provides no further benefit. We use $K{=}5$ throughout. An inference-time sweep of $\beta_0$ and $\epsilon_{\text{safe}}$ on the fixed FS-DFM student shows a narrow spread, with every setting better than blind sampling (\cref{app:nav_sensitivity}). This measures the sensitivity of the operator, not of the shaped student.

\paragraph{Comparison with state-of-the-art.}
\label{sec:few_step_sota}
\begin{table}[t]
\centering
\caption{\textbf{Few-step generation comparison at 8 and 16 steps.} Methods span four families: masked diffusion (MDLM, SEDD, ReMDM variants), multi-round distillation (SDTT), consistency distillation (Duo), and flow-matching distillation (FS-DFM, \ours). For FS-DFM and \ours, the first 65.5B tokens correspond to DFM pretraining. Additional tokens (\ddag) denote fine-tuning. $^\dagger$Reproduced using official code. Sampling variance at fixed checkpoints (5 decoding seeds) is in \cref{app:seed_variance}.}
\label{tab:few_step_sota}
\small
\begin{tabular}{@{}lrr rr rr@{}}
%\toprule
& & & \multicolumn{2}{c}{\textbf{8 steps}} & \multicolumn{2}{c}{\textbf{16 steps}} \\
\cmidrule(lr){4-5} \cmidrule(lr){6-7}
\textbf{Method} & \textbf{Size (B)} & \textbf{Tokens (B)} & PPL$\downarrow$ & Ent & PPL$\downarrow$ & Ent \\
\midrule
ReMDM-loop$^\dagger$      & 0.17 & 262  & 1319.2 & 8.4 & 413.6 & 8.3 \\
ReMDM-cap$^\dagger$       & 0.17 & 262  & 546.8  & 8.3 & 181.8 & 8.2 \\
ReMDM-conf$^\dagger$      & 0.17 & 262  & 545.2  & 8.3 & 181.3 & 8.2 \\
ReMDM-rescale$^\dagger$   & 0.17 & 262  & 555.8  & 8.4 & 181.1 & 8.2 \\
MDLM-OWT$^\dagger$        & 0.17 & 262  & 815.6  & 8.5 & 304.3 & 8.3 \\
SEDD$^\dagger$            & 0.32 & --   & 683.7  & 8.4 & 305.6 & 8.3 \\
SDTT (1 round)$^\dagger$  & 0.86 & --   & 567.8  & 8.4 & 215.5 & 8.2 \\
SDTT (3 rounds)$^\dagger$ & 0.86 & --   & 305.4  & 8.2 & 128.2 & 8.1 \\
SDTT (5 rounds)$^\dagger$ & 0.86 & --   & 184.8  & 8.0 & 84.1  & 8.0 \\
SDTT (7 rounds)$^\dagger$ & 0.86 & --   & 105.8  & 7.8 & 53.6  & 7.7 \\
Duo$^\dagger$             & 0.17 & 524  & 70.6   & 7.6 & 53.6  & 7.7 \\
FS-DFM$^\dagger$          & 0.17 & $65.5{+}13.2^\ddag$ & 87.6  & 7.4 & 73.2  & 7.4 \\
\rowcolor{selectedrow}
\textbf{\ours} & 0.17 & $78.7{+}6.4^\ddag$  & \textbf{56.1} & 7.2 & \textbf{47.2} & 7.4 \\
%\bottomrule
\end{tabular}
\end{table}

\Cref{tab:few_step_sota} compares \ours against state-of-the-art few-step discrete generation methods at 8 and 16 steps. \ours achieves the best perplexity at both step counts. At 8 steps it reaches 56.1 GPT-2 PPL---19\% lower than Duo (70.6, trained on $6\times$ more tokens) and 46\% below SDTT after 7 rounds (105.8, using a $5\times$ larger model). At 16 steps the gap widens: 47.2 vs.\ 53.6 for both Duo and SDTT.

\paragraph{Training cost.}
\cref{tab:c2f_wallclock} reports wall-clock training overhead measured on a single GPU. At the default $\tau{=}0.2$, \ours adds $2.2\times$ overhead per step relative to FS-DFM, ranging from $2.0\times$ ($\tau{=}0.4$) to $2.4\times$ ($\tau{=}0.02$). This is modestly higher than the theoretical $1.7\times$ FLOP estimate (\cref{app:c2f_cost}), reflecting practical costs: generating $K$ candidate CTMC jumps, memory allocation for batched energy scoring, and Python-level control flow. The $K$ candidate energy evaluations are batched into a single GPU kernel, so overhead scales sublinearly with $K$. Higher thresholds reduce cost by activating guidance at fewer midpoints: $\tau{=}0.4$ guides only midpoints where over 40\% of tokens are revealed. The narrow spread ($2.0$--$2.4\times$) indicates that the fixed cost of maintaining the energy compass dominates the variable cost of per-midpoint scoring. This is a one-time training cost; inference is unchanged.

\paragraph{Scaling to 1.3B parameters.}
\label{sec:scaling_main}
We repeat the uniform-source comparison at 1.3B---$7.6\times$ our main model---guided by the same 90M energy compass (\cref{tab:scaling_1_3b}; \cref{app:scaling}).
(1) The shaped student beats its unguided $1\,024$-step teacher at just 8 steps:
\ours{} reaches 48.0 GPT-2 Large PPL versus the teacher's 57.5---a $128\times$
inference speedup with \emph{improved} quality.
(2) At 8 steps, the gap over FS-DFM remains at least as large at 1.3B: \ours{} cuts FS-DFM's PPL by $41\%$ at 1.3B (48.0 vs.\ 81.5) versus $36\%$ at 170M (56.1 vs.\ 87.6). At 16 and 32 steps the reductions are similar at both scales (36\% at 170M vs.\ 33\% at 1.3B at 16 steps, 31\% vs.\ 30\% at 32 steps; \cref{app:scaling}).
(3) Relative cost shrinks at scale. The compass-to-student ratio
drops from ${\sim}1{:}2$ at 170M to ${\sim}1{:}14$ at 1.3B, so navigation's (fixed-size) FLOP overhead becomes proportionally cheaper as students grow.
Beyond unconditional generation, a transfer study on GSM8K and SVAMP at 1.3B scale (\cref{sec:gsm8k}) applies the same mechanism to task-fine-tuned math reasoning. At the 4- and 8-step budgets we evaluate, \ours{} matches or exceeds FS-DFM, by small margins that we read as directionally consistent.

\begin{table}[h]
\centering
\small
\caption{\textbf{Scaling to 1.3B parameters (uniform source).} Perplexity (PPL $\downarrow$) and generation entropy (Ent) for $1\,024$-token unconditional generation at 4, 8, 16, and 32 sampling steps, with the $1\,024$-step DFM teacher as the quality reference. \colorbox{beatteacher}{\textbf{Green}} cells surpass the $1\,024$-step DFM teacher.}
\label{tab:scaling_1_3b}
\setlength{\tabcolsep}{3pt}
\begin{tabular}{@{}llcccccccccccc@{}}
%\toprule
& & \multicolumn{4}{c}{\textbf{DFM}} & \multicolumn{4}{c}{\textbf{FS-DFM}} & \multicolumn{4}{c}{\textbf{\ours{}}} \\
\cmidrule(lr){3-6} \cmidrule(lr){7-10} \cmidrule(lr){11-14}
\textbf{Steps} & & Ent & G$^2$ $\downarrow$ & L$^2$ $\downarrow$ & L$^3$ $\downarrow$
               & Ent & G$^2$ $\downarrow$ & L$^2$ $\downarrow$ & L$^3$ $\downarrow$
               & Ent & G$^2$ $\downarrow$ & L$^2$ $\downarrow$ & L$^3$ $\downarrow$ \\
\midrule
4    & & 8.1 & 913.6 & 458.6 & 915.2 & 7.2 & 188.6 & 80.8 & 170.3 & 7.0 & \textbf{98.9} & \textbf{51.0} & \textbf{92.4} \\
8    & & 8.1 & 320.7 & 169.6 & 320.6 & 7.4 & 81.5 & 38.2 & 73.5 & 7.3 & \cellcolor{beatteacher}\textbf{48.0} & \cellcolor{beatteacher}\textbf{26.6} & \cellcolor{beatteacher}\textbf{46.7} \\
16   & & 8.0 & 142.3 & 77.4 & 138.8 & 7.5 & 49.5 & 23.9 & 44.5 & 7.5 & \cellcolor{beatteacher}\textbf{33.1} & \cellcolor{beatteacher}\textbf{18.8} & \cellcolor{beatteacher}\textbf{32.0} \\
32   & & 7.9 & 91.9 & 51.3 & 88.5 & 7.7 & 44.9 & 22.8 & 41.0 & 7.6 & \cellcolor{beatteacher}\textbf{31.5} & \cellcolor{beatteacher}\textbf{18.1} & \cellcolor{beatteacher}\textbf{30.7} \\
\midrule
\cellcolor{gray!15}$1\,024$ & & \cellcolor{gray!15}7.9 & \cellcolor{gray!15}57.5 & \cellcolor{gray!15}32.1 & \cellcolor{gray!15}54.5
     & \multicolumn{4}{c}{\cellcolor{gray!15}---}
     & \multicolumn{4}{c}{\cellcolor{gray!15}---} \\
%\bottomrule
\end{tabular}
\end{table}

\section{Conclusion}
\label{sec:conclusion}
We introduced \ours{}, built on the observation that the performance ceiling in few-step discrete flow distillation is set by the training trajectories, not the student. Blind stochastic jumps at RK-4 midpoints compound errors that the student faithfully imitates. \ours{} replaces these jumps with guided navigation via a lightweight energy compass and a two-phase Sequence-to-Token policy, at training time only. At 170M parameters, \ours{} is state-of-the-art among few-step discrete generation methods on both uniform and mask sources. At 1.3B with the uniform source, guided by the same lightweight compass, the shaped student beats its unguided $1\,024$-step teacher at just 8 steps, a $128\times$ speedup with \emph{improved} quality. At this step budget, the gap over the FS-DFM baseline remains at least as large at 1.3B as at 170M ($36\%{\to}41\%$ PPL reduction).

\paragraph{Limitations.} The compass is trained once and frozen. As the student improves, its flow-state distribution shifts, potentially reducing compass effectiveness (iterative refinement could help but risks co-adaptation). In the worst-case run, the frozen compass shows no measurable degradation on our drift metrics over the full run (\cref{app:compass_drift}), but we have not run the compass-refresh or EMA-recompute control. The activation threshold $\tau$ is fixed throughout training; adaptive or annealed schedules remain unexplored. Our evaluation relies mainly on generative perplexity, which coherence--diversity trade-offs can game. We report entropy with every perplexity and score under three evaluators, but we include no human evaluation. The scaling evidence covers two model sizes, with only the uniform source at 1.3B, so a third scale and the mask source at 1.3B remain future work. Downstream evidence is limited to GSM8K and SVAMP. Instruction-tuned students at larger scale, where benchmarks such as MATH, HumanEval and IFEval become meaningful, are also left for future work.

\section*{Acknowledgments}
We thank Barry Theobald and Zhe Gan for feedback on early versions of this paper.

% ============================================================
% Bibliography
% ============================================================
\bibliographystyle{plainnat}
\bibliography{references}

% ============================================================
% Appendix
% ============================================================
\crefalias{section}{appendix}
\crefalias{subsection}{appendix}

\appendix

\clearpage
\appendix

\part*{Appendix}
\addcontentsline{toc}{part}{Appendix}

\etocsettocstyle{\section*{Appendix Contents}}{}
\etocsetnexttocdepth{subsection}
\localtableofcontents

\newpage

\section{Energy compass training details}
\label{app:energy_training}

\cref{sec:energy_model} describes the energy compass's role, training objective, and key design choices.
This section provides the full training specification, including the objective decomposition, negative construction details, source-distribution handling, and optimization configuration.

We train a separate energy compass for each source family: one for the \emph{uniform} source distribution ($x_0$ sampled uniformly over $\mathcal{V}$) and one for the \emph{mask} source distribution ($x_0 = [\textsc{Mask}]^L$). The two compasses share identical architecture and training procedure (\cref{tab:energy_arch,tab:energy_optim}); they differ only in the source tokens that populate unrevealed positions during positive and negative construction. All validation results in \cref{app:energy_compass_validation} are reported for both variants where applicable.

\subsection{Architecture}
\label{app:energy_arch}

\begin{wraptable}{r}{0.45\textwidth}
\vspace{-\intextsep}
\centering
\small
\caption{Energy compass architecture.}
\label{tab:energy_arch}
\begin{tabular}{@{}ll@{}}
%\toprule
\textbf{Component} & \textbf{Configuration} \\
\midrule
Transformer blocks & 6 \\
Hidden dimension & 768 \\
Attention heads & 12 \\
\shortstack[l]{Pooling\\~} & \shortstack[l]{Attention\\(learned query $\to$ scalar)} \\
Time conditioning & None (sequence only) \\
Parameters & ${\sim}$90M \\
%\bottomrule
\end{tabular}
\vspace{-\intextsep}
\end{wraptable}

The energy compass is a transformer encoder with attention pooling that maps a token sequence to a scalar energy.
It shares the same tokenizer (GPT-2, $|\mathcal{V}|{=}50\,257$) and block size ($L{=}1\,024$) as the velocity model but is substantially smaller (\cref{tab:energy_arch}). The lack of time conditioning is a deliberate design choice: the compass scores intrinsic sequence quality, not quality-relative-to-timestep.
In place of a time embedding, the model uses a learned constant buffer, ensuring the architecture is identical to a time-conditioned variant (for checkpoint compatibility) but receives no time information.

\subsection{Training objective}
\label{app:energy_objective}
The total loss is a weighted sum of three components, each addressing a distinct training challenge:
\begin{equation*}
  \mathcal{L}_{\text{energy}} = \mathcal{L}_{\text{NCE}} + \lambda_{\text{reg}} \cdot \mathcal{L}_{\text{reg}} + \lambda_{\text{order}} \cdot \mathcal{L}_{\text{order}},
  \qquad \lambda_{\text{reg}} = \lambda_{\text{order}} = 1.0
\end{equation*}

\textbf{The noise-contrastive loss ($\mathcal{L}_{\text{NCE}}$)}: for each positive flow state $x_t$, we generate $N_{\text{neg}}{=}12$ corrupted negatives $\{\tilde{x}_t^{(j)}\}$ via the strategies in \cref{app:energy_negatives} and minimize:
\begin{equation}
  \mathcal{L}_{\text{NCE}} = -\log \frac{\exp(-E_\phi(x_t))}{\exp(-E_\phi(x_t)) + \sum_{j=1}^{N_{\text{neg}}} \exp(-E_\phi(\tilde{x}_t^{(j)}))},
  \label{eq:nce}
\end{equation}
training the compass to assign lower energy to the real flow state than to any corruption.

\textbf{Regularization loss ($\mathcal{L}_{\text{reg}}$)}: prevents energy scale drifts during training---the mean energy can swing by several units between evaluation checkpoints, destabilizing the NCE gradients and collapsing the time-ordering signal.
We penalize deviation from zero, which anchors the energy scale without constraining the \emph{relative} ordering between samples:
\begin{equation}
	\mathcal{L}_{\text{reg}} = E_\phi(x_t)^2
\end{equation}

\textbf{Time-ordering loss ($\mathcal{L}_{\text{order}}$)}:
The NCE objective teaches within-timestep discrimination (corrupt vs.\ clean at fixed $t$), but does not coordinate energy levels \emph{across} timesteps.
To enforce monotonically decreasing energy as more tokens are revealed, we construct a less-revealed version $x_{t'}$ at $t' = \max(t - \delta, 0.01)$ where $\delta \sim \text{Uniform}(0.05, 0.5)$, using the same $(x_0, x_1)$ pair, and apply a margin-based hinge loss:
\begin{equation}
	\mathcal{L}_{\text{order}} = \max\!\big(0,\; E_\phi(x_t) - E_\phi(x_{t'}) + \gamma \cdot \delta\big),
\end{equation}
with a proportional margin $\gamma{=}0.3$.
The margin scales with $\delta$ so larger time gaps require larger energy differences, providing a smooth gradient signal toward monotonicity.

\subsection{Negative strategies}
\label{app:energy_negatives}

Negatives are constructed by corrupting \emph{revealed positions only}---positions where $x_t$ already carries a target token from $x_1$. Unrevealed positions retain their source tokens (uniform random tokens or \textsc{[Mask]}), since these are not informative to corrupt.

We use two general categories of negatives, each targeting a distinct failure mode the compass must detect:
\begin{itemize}
  \item \textbf{Generation-realistic negatives} simulate the actual errors that arise during discrete flow matching---teacher reconstruction mistakes (\texttt{velocity\_step}) and under-revealed flow states (\texttt{time\_downstep}). These are the hardest negatives and most directly represent the comparisons the compass makes during trajectory navigation.
  \item \textbf{Token-corruption negatives} degrade revealed tokens with wrong or repeated substitutions (\texttt{random\_replace}, \texttt{frequency\_replace}, \texttt{token\_repeat}). These are cheaper to construct and teach the compass to reject content-level corruptions that are unrelated to the flow dynamics.
\end{itemize}

Each negative is drawn from one of the five strategies in \cref{tab:energy_negatives}, sampled in proportion to the probabilities listed. Generation-realistic negatives together account for ${\sim}45\%$ of training negatives. Within this group, \texttt{velocity\_step} produces the most realistic negatives but is sampled less often (9.1\%) to limit the cost of the frozen teacher forward pass it requires; all other strategies incur no additional model inference.

\begin{table}[h]
\centering
\small
\caption{Negative strategies for energy model training, grouped by category. Rate ranges only apply to token-level strategies.}
\label{tab:energy_negatives}
\setlength{\tabcolsep}{4pt}
\begin{tabular}{@{}llcp{5.6cm}@{}}
%\toprule
\textbf{Strategy} & \textbf{Prob.} & \textbf{Rate range} & \textbf{Description} \\
\midrule[\lightrulewidth]

\multicolumn{4}{@{}l}{\emph{Generation-realistic negatives}} \\
\\
\texttt{time\_downstep}    & 36.4\% & $\delta \in [0.025, 0.5]$ & \emph{Temporal.} Construct $x_{t'}$ at lower $t' < t$ from same $(x_0, x_1)$. A real but less-revealed flow state. \\
\texttt{velocity\_step}    & 9.1\%  & --- & \emph{Near-miss.} Step backward then forward through the teacher to produce realistic reconstruction-error negatives (see below). \\
\\
\midrule[\lightrulewidth]

\multicolumn{4}{@{}l}{\emph{Token-corruption negatives}} \\
\texttt{random\_replace}   & 18.2\% & [0.05, 0.25] & \emph{Wrong tokens.} Replace revealed tokens with uniform random vocabulary items. \\
\texttt{frequency\_replace}& 18.2\% & [0.03, 0.20] & \emph{Wrong tokens.} Replace with tokens from the same unigram frequency bin---cannot be detected by frequency statistics alone. \\
\texttt{token\_repeat}     & 18.2\% & [0.05, 0.50] & \emph{Repetition.} Repeat a single token across revealed positions. Simulates degenerate repetition. \\
%\bottomrule
\end{tabular}
\end{table}

\subsubsection{Velocity near-miss construction}
The \texttt{velocity\_step} strategy produces the most realistic negatives and deserves additional detail. Given a positive $x_t$ at time $t$:
\begin{enumerate}
  \item Sample a step size $h$ from a set of discrete options spanning $[{\sim}0.001, 0.0625]$.
  \item Construct $x_{t-h}$ by probabilistically unrevealing positions: each revealed position in $x_t$ is kept with probability $\alpha(t{-}h)/\alpha(t)$ and replaced with a source token otherwise (uniform random for the uniform compass, \textsc{[Mask]} for the mask compass). This yields a sequence with ${\sim}\alpha(t{-}h)$ fraction revealed---consistent with the flow dynamics at time $t{-}h$.
  \item Run the frozen teacher velocity model on $(x_{t-h}, t{-}h)$ to obtain logits, then take a forward Euler step back to time $t$, producing $\hat{x}_t$.
  \item Return $\hat{x}_t$ as the negative, paired with the positive $x_t$ at the same time $t$.
\end{enumerate}
The resulting negative competes directly with the positive: both are plausible sequences at time $t$, but $\hat{x}_t$ contains the teacher's reconstruction errors at the positions that were unrevealed in step~2. These are precisely the kind of errors the compass must detect when comparing CTMC jump candidates during trajectory construction.

\paragraph{Teacher velocity model.}
The \texttt{velocity\_step} strategy requires a frozen teacher velocity model to generate realistic near-miss negatives. Each compass variant uses its matching DFM teacher (170M parameters)---the uniform-trained teacher for the uniform compass, the mask-trained teacher for the mask compass---identical to the teacher used in \ours{} and FS-DFM distillation. The teacher is loaded in eval mode with gradients disabled; its forward-pass cost is incurred only when \texttt{velocity\_step} is sampled for a given batch element.

\subsubsection{Corruption rate sampling}
For token-level strategies (\texttt{random\_replace}, \texttt{frequency\_replace}, \texttt{token\_repeat}), the corruption rate is sampled from a Beta(1,3) distribution over $[\text{rate}_{\min}, \text{rate}_{\max}]$. Beta(1,3) has mean 0.25 of its range, biasing toward low corruption rates and thus toward harder negatives where only a few tokens differ from the positive. A floor of 10\% of revealed positions ensures that even at high $t$---where few positions remain revealed to corrupt---negatives still differ from positives enough to provide a meaningful gradient signal.

\subsection{Positive sample construction}
\label{app:energy_positives}

Positive samples are flow states $x_t$ constructed via the standard interpolation path (\cref{eq:interpolation}) from $(x_0, x_1)$ pairs, where $x_1$ is a real text sequence from FineWeb-Edu~\citep{penedo2024finewebdatasetsdecantingweb}.
The source $x_0$ is drawn from the corresponding source distribution: uniform random tokens over $\mathcal{V}$ for the uniform compass, or $[\textsc{Mask}]^L$ for the mask compass.
The time $t$ is sampled uniformly from $(0, 1)$.

To anchor the energy scale, 10\% of training positives use the clean target $x_1$ directly (at $t{=}1$) rather than a path-interpolated $x_t$.
This ensures the compass learns that fully-revealed, real text should receive the lowest energy---providing a stable reference point for the energy landscape.

\subsection{Optimization}
\label{app:energy_optim}

\begin{wraptable}{r}{0.48\textwidth}
\vspace{-\intextsep}
\centering
\small
\caption{Energy compass optimization.
}
\label{tab:energy_optim}
\begin{tabular}{@{}ll@{}}
%\toprule
\textbf{Parameter} & \textbf{Value} \\
\midrule
\shortstack[l]{Optimizer\\~} & \shortstack[l]{AdamW\\($\beta_1{=}0.9$, $\beta_2{=}0.999$)} \\
Peak learning rate & $1 \times 10^{-4}$ \\
\shortstack[l]{LR schedule\\~} & \shortstack[l]{Cosine decay\\(min ratio 0.1)} \\
Warmup steps & $5\,000$ \\
Weight decay & 0.01 \\
Gradient clipping & 1.0 \\
Batch size & 64 \\
Precision & bf16 \\
GPUs & 8 \\
Training data & FineWeb-Edu \\
Validation data & WikiText-103 \\
%\bottomrule
\end{tabular}
\vspace{-\intextsep}
\end{wraptable}

The learning rate and warmup schedule were tuned to keep the energy scale stable during early training, which the time-ordering loss (\cref{app:energy_objective}) depends on. With a higher peak LR of $3 \times 10^{-4}$, the mean energy drifted by more than one unit across the warmup ramp---large enough to overwhelm the margin in $\mathcal{L}_{\text{order}}$ and collapse the time-ordering signal. Lowering the peak to $1 \times 10^{-4}$ with $5\,000$ warmup steps eliminated this instability.

\paragraph{Checkpoint selection.}
The energy compass must satisfy two properties: within-timestep discrimination (high NCE accuracy per strategy,~\cref{app:corruption_detection}) and across-timestep monotonicity (smooth energy decrease with $t$,~\cref{app:monotonicity}).
These can conflict: stronger NCE gradients widen per-timestep energy gaps without coordinating absolute energy levels across timesteps.
We select the best checkpoint via a composite score that sums the overall NCE accuracy across strategies and the fraction of consecutive time-bin pairs with correctly decreasing energy ($\text{smooth\_pct}$): 
\begin{equation}
  \text{score} = \text{strategy\_acc} + \text{smooth\_pct}
\end{equation}
Equal weighting reflects that both properties are necessary for reliable navigation: high discrimination without monotonicity produces a compass that ranks well within a timestep but assigns inconsistent absolute energies across timesteps, while monotonicity without discrimination fails to distinguish subtle quality differences among candidates.

%=============================================================================
% APPENDIX: Energy Compass Validation
%=============================================================================

\section{Energy compass validation}
\label{app:energy_compass_validation}

Before using the energy compass to guide trajectory construction, we must verify two properties:
(1)~that it assigns monotonically lower energy to sequences with more revealed tokens (reflecting increasing quality along the flow path), and
(2)~that it assigns lower energy to real flow states than to corrupted versions exhibiting the specific failure modes that arise during discrete flow matching generation---not just arbitrary noise.
We validate both properties on held-out sequences from FineWeb-Edu using the frozen, pretrained energy model.
Since we train a separate compass for each source family (\cref{app:energy_training}), below we report results for both the \emph{uniform} and \emph{mask} variants.

\subsection{Time-energy monotonicity}
\label{app:monotonicity}

In discrete flow matching, a sequence $x_t$ at time $t$ has approximately $\alpha(t)$ fraction of its positions revealed (matching the target $x_1$); the rest remain source noise.
As $t$ increases, $\alpha(t)$ grows and the sequence becomes intrinsically higher-quality text.
For the energy compass to be a reliable navigator, its scores must reflect this ordering: $E_\phi(x_t)$ should decrease monotonically with $t$.
Formally, for any two flow states constructed from the same $(x_0, x_1)$ pair at times $t > t'$:
\begin{equation}
  E_\phi(x_t) < E_\phi(x_{t'}) \qquad \text{(higher } t = \text{more revealed} = \text{lower energy)}.
  \label{eq:monotonicity_criterion}
\end{equation}

To measure how consistently the trained compass satisfies this criterion on held-out data, we take 200 held-out $(x_0, x_1)$ pairs and construct flow states at 20 randomly sampled timesteps and the clean target ($t{=}1.0$), producing $4\,200$ total evaluations per source distribution.
We compute $E_\phi(x_t)$ and report binned mean energy, monotonicity, rank correlation, and aggregate metrics in \cref{tab:monotonicity,tab:monotonicity_aggregate}.

\begin{table}[h]
\centering
\small
\caption{\textbf{Time-energy monotonicity.} Mean energy $E_\phi(x_t)$ per time bin; both source distributions. 
}
\label{tab:monotonicity}
\setlength{\tabcolsep}{3.5pt}
\footnotesize
\begin{tabular}{@{}cl*{11}{r}@{}}
%\toprule
& & \multicolumn{11}{c}{\textbf{Time bin}} \\
\cmidrule(lr){3-13}
\textbf{Src} & & {[0,.1)} & {[.1,.2)} & {[.2,.3)} & {[.3,.4)} & {[.4,.5)} & {[.5,.6)} & {[.6,.7)} & {[.7,.8)} & {[.8,.9)} & {[.9,1)} & {$t{=}1$} \\
\midrule
\multirow{2}{*}{Uni.}
  & $\bar{E}_\phi$ & $-$0.12 & $-$0.57 & $-$0.75 & $-$0.90 & $-$1.02 & $-$1.10 & $-$1.17 & $-$1.22 & $-$1.29 & $-$1.34 & $-$1.45 \\
  & \textit{std}    &    0.50 &    0.07 &    0.06 &    0.06 &    0.08 &    0.07 &    0.12 &    0.21 &    0.21 &    0.38 &    0.38 \\
\midrule
\multirow{2}{*}{Mask}
  & $\bar{E}_\phi$ & $-$1.94 & $-$2.24 & $-$2.35 & $-$2.43 & $-$2.47 & $-$2.55 & $-$2.59 & $-$2.66 & $-$2.68 & $-$2.75 & $-$3.20 \\
  & \textit{std}    &    0.34 &    0.13 &    0.19 &    0.21 &    0.34 &    0.31 &    0.42 &    0.38 &    0.52 &    0.55 &    0.61 \\
%\bottomrule
\end{tabular}
\end{table}

The binned means are \emph{perfectly} monotone for both source distributions: all 10 consecutive bin pairs show strictly decreasing energy. At the sample level, the uniform compass achieves 97.2\% smoothness with only 44 of 200 samples exhibiting any violation, while the mask compass achieves 94.0\% with 74 violating samples. The mask compass assigns lower (more negative) energies overall because \textsc{[Mask]} tokens provide a uniform, low-entropy background at unrevealed positions, making partially revealed sequences more structured. This compresses the total energy dynamic range (1.26 vs.\ 1.33 for uniform) and the gap between adjacent timesteps, which explains both the higher per-sample violation rate and the weaker Pearson correlation ($-0.54$ vs.\ $-0.80$)---the energy--time relationship is less linear for the mask source because the energy curve flattens more quickly in the mid-$t$ range. None of this affects navigation, however, which depends only on relative energies between candidates at the same midpoint; the perfect bin-level monotonicity of both variants confirms the compass's suitability for this purpose.

\begin{table}[h]
\centering
\small
\caption{\textbf{Aggregate monotonicity metrics.} Both compasses achieve perfect bin-level monotonicity. 
}
\label{tab:monotonicity_aggregate}
\begin{tabular}{@{}lcc@{}}
%\toprule
\textbf{Aggregate metric} & \textbf{Uniform} & \textbf{Mask} \\
\midrule
Adjacent bins strictly monotone & 10/10 (100\%) & 10/10 (100\%) \\
Per-sample smoothness (adj.\ pairs satisfying \cref{eq:monotonicity_criterion}) & 97.2\% ($3\,886$/$4\,000$) & 94.0\% ($3\,760$/$4\,000$) \\
Samples with any violation & 44/200 (22\%) & 74/200 (37\%) \\
Spearman $\rho$ ($t$ vs.\ $E_\phi$, all $4\,200$ points) & $-$0.94 & $-$0.85 \\
Pearson $r$ ($t$ vs.\ $E_\phi$, all $4\,200$ points) & $-$0.80 & $-$0.54 \\
Energy gap (mean $E$ at lowest $t$ $-$ mean $E$ at $t{=}1$) & 1.33 & 1.26 \\
%\bottomrule
\end{tabular}
\end{table}

\textbf{Violations by time region.}
\cref{tab:violations_by_time} breaks down per-sample violations by time region for both source distributions. The two distributions exhibit qualitatively different violation profiles.
For the uniform compass, violations concentrate at high $t$ where adjacent timesteps differ by only a few tokens and energy differences are small relative to estimation noise---as reflected in the growing standard deviation from 0.07 at $t{\approx}0.2$ to 0.38 at $t{=}1.0$ (\cref{tab:monotonicity}).
The mask compass instead distributes violations more broadly, peaking in the $[0.3, 0.7)$ range.
This pattern is consistent with the compressed energy curve: in the mid-$t$ regime, the mask compass's energy differences between adjacent bins are small (0.04--0.08 units), comparable to the per-sample standard deviation, making local ordering more susceptible to noise.

Importantly, violations in both cases are mild.
The largest energy gap in the wrong direction is 1.14 for the uniform compass and 1.40 for the mask compass, both concentrated in a small number of outlier samples.
These violations do not affect the bin-level monotonicity, which is perfect for both variants.
This result is notable because neither energy model receives \emph{any time input}---they score the raw token sequence only.
The monotonic decrease confirms that both compass variants learn to assess intrinsic text quality from content alone: sequences with more coherent, on-target tokens receive lower energy regardless of which timestep produced them.
This property is precisely what enables the compass to compare candidate midpoints during RK-4 trajectory construction,
 where the ``correct'' timestep is not well-defined for stochastically generated candidates.

\begin{table}[h]
\centering
\small
\caption{\textbf{Monotonicity violations by time region.} Number of adjacent time pairs where $E_\phi$ incorrectly increases with $t$, for both source distributions. 
}
\label{tab:violations_by_time}
\begin{tabular}{@{}lrrrrrrrrrr@{}}
%\toprule
\textbf{Time region} & [0, .1) & [.1, .2) & [.2, .3) & [.3, .4) & [.4, .5) & [.5, .6) & [.6, .7) & [.7, .8) & [.8, .9) & [.9, 1) \\
\midrule
Uniform & 3 & 4 & 4 & 4 & 11 & 12 & 18 & 19 & 17 & 22 \\
Mask    & 8 & 16 & 14 & 35 & 30 & 34 & 30 & 32 & 25 & 16 \\
%\bottomrule
\end{tabular}
\end{table}

\subsection{Generation-relevant corruption detection}
\label{app:corruption_detection}

A compass that assigns low energy to clean text is insufficient; it must assign \emph{higher} energy to sequences exhibiting the specific corruptions that arise during discrete flow matching generation.
We evaluate both energy model variants on their ability to correctly rank real flow states below corrupted versions across four types of corruption, organized by the failure mode they simulate. Formally, for a real flow state $x_t$ and a corrupted version $\tilde{x}_t$ constructed by applying any generation-relevant corruption strategy:
\begin{equation}
  E_\phi(x_t) < E_\phi(\tilde{x}_t) \qquad \text{(real flow state} = \text{lower energy than corrupted)}.
  \label{eq:corruption_criterion}
\end{equation}
We measure how consistently the trained compass satisfies this criterion across four corruption strategies on held-out data:
\begin{itemize}
\item \textbf{Random replacement} (\texttt{random\_replace}): Replace revealed tokens with uniform random vocabulary items (${\sim}$99 tokens changed on average). Tests basic quality detection.
  \item \textbf{Frequency-matched replacement} (\texttt{frequency\_replace}): Replace revealed tokens with tokens from the same unigram frequency bin (${\sim}$104 tokens changed). A harder variant that cannot be detected by frequency statistics alone.
  \item \textbf{Token repetition} (\texttt{token\_repeat}): Repeat a single token across revealed positions (${\sim}$93--103 tokens changed). Simulates the degenerate repetition failure mode common in discrete diffusion.
  \item \textbf{Time downstep} (\texttt{time\_downstep}): Construct $x_{t'}$ at $t' < t$ from the same $(x_0, x_1)$ pair (${\sim}$116 tokens changed). This is a less-revealed version of the same text---testing whether the compass detects the quality difference between two \emph{real} flow states at different times.
\end{itemize}

For each source distribution, we evaluate $1\,130$ held-out (positive, negative) pairs per strategy ($4\,520$ pairs total per variant), measuring the fraction satisfying $E_\phi(\text{real}) < E_\phi(\text{corrupted})$, i.e., \cref{eq:corruption_criterion}.

Both compass variants satisfy \cref{eq:corruption_criterion} near-perfectly across all four strategies.
The uniform compass achieves a mean accuracy of 98.5\%, with no strategy falling below 98\%.
The mask compass achieves a mean accuracy of 99.8\%, with perfect match on three of four strategies and 99.4\% on token repetition.

The mask compass's higher overall accuracy may appear to contradict the weaker monotonicity metrics shown in \cref{tab:monotonicity}; however, these measure different properties.
Corruption detection compares a real flow state against an explicitly degraded version at the \emph{same} timestep---a within-timestep discrimination task where the quality gap is large.
Monotonicity requires distinguishing real flow states at \emph{adjacent} timesteps---a much finer-grained comparison where even a few tokens of difference can be within the estimation noise.
The mask compass excels at the former while struggling more with the latter, consistent with its compressed energy range.

\begin{wraptable}{r}{0.55\textwidth}
\vspace{-\intextsep}
\centering
\small
\caption{\textbf{Per-strategy energy ranking accuracy.}
}
\label{tab:strategy_accuracy}
\begin{tabular}{@{}lcc@{}}
%\toprule
\textbf{Strategy} & %\textbf{Failure mode} &
\shortstack{\textbf{Uniform}\\(\%)} & \shortstack{\textbf{Mask}\\(\%)} \\
\midrule
\texttt{random\_replace}    & %Wrong tokens (easy)     &
98.6 & 100.0 \\
\texttt{frequency\_replace} & %Wrong tokens (hard)     &
98.4 & 100.0 \\
\texttt{token\_repeat}      & %Degenerate repetition   &
98.8 & 99.4  \\
\texttt{time\_downstep}     & %Less-revealed real text  &
98.2 & 100.0 \\
\midrule
\textbf{Overall}            & %                        &
\textbf{98.5} & \textbf{99.8} \\
%\bottomrule
\end{tabular}
\vspace{-\intextsep}
\end{wraptable}

Both variants' uniformly high performance across strategies is notable.
The strategies span fundamentally different corruption types---wrong tokens (random and frequency-matched replacement), degenerate repetition (token repeat), and reduced revelation (time downstep)---yet both compasses handle all of them equally well.
This indicates that $E_\phi$ has learned a general notion of sequence quality rather than a narrow definition that fits only a single failure mode.

Frequency-matched replacement achieves near-identical accuracy to random replacement for both variants (98.4\% vs.\ 98.6\% for uniform; both 100\% for mask).
Since frequency-matched negatives preserve the unigram token distribution of the original sequence, the energy model cannot rely on simple frequency cues---it must assess contextual coherence to distinguish real from corrupted flow states.

We deliberately exclude order-sensitive corruptions (token shuffling, segment reversal, neighbor swaps) from both training and evaluation.
During discrete flow matching generation, the velocity model reveals tokens \emph{at their correct positions}---it never permutes correct tokens into wrong positions.
Order-based corruptions test a capability irrelevant to the downstream use case and waste training signal on a task that a small transformer struggles with.
Focusing the training budget on generation-relevant corruptions produces a compass better calibrated for its actual role.
% ============================================================
% Sequence-to-Token Compass Navigation (Appendix)
% Figures integrated inline with corresponding text.
%
% Uses ONLY existing project colors — no new \definecolor needed:
%   energyblue, trajgood, shapegold, trajbad, baselinegray,
%   shapegoldlight, lightblue, signalgreen
%
% Required tikz libraries (should already be in preamble):
%   \usetikzlibrary{arrows.meta, decorations.pathreplacing,
%                    positioning, calc, fit, backgrounds,
%                    shapes.geometric}
% ============================================================

\section{Sequence-to-Token compass navigation}
\label{app:c2f_navigation}

We expand \cref{sec:guided_midpoint} with the complete specification of the $\textsc{Navigate}$ procedure used at each RK-4 midpoint during navigation shaping.

\subsection{Motivation: the granularity dilemma in discrete guidance}
\label{app:c2f_motivation}

Recall that the energy compass $E_\phi(x_t)$ provides a scalar assessment of global sequence quality.
Using it to guide CTMC midpoint jumps requires converting the global signal into per-position transition decisions. The mathematically ideal guided rate modifies each single-token transition by its energy differential:
\begin{equation}
  \tilde{u}_t(y, x) = u_t^\theta(y, x) \cdot \exp\!\bigl(-\beta_t\,[E_\phi(y) - E_\phi(x)]\bigr),
  \label{eq:ideal_rate_app}
\end{equation}
where $y = x[i \to v]$ differs from $x$ by one token at position $i$.
Evaluating this for all positions and all vocabulary entries requires $|\mathcal{V}| \times L$ energy forward passes per midpoint, which is intractable.

The two naive strategies for approximating \cref{eq:ideal_rate_app} occupy opposite extremes of a granularity spectrum, and neither is satisfactory for trajectory construction:

\begin{center}
\renewcommand{\arraystretch}{1.12}
\small
\begin{tabular}{@{}lcc@{}}
%\toprule
\textbf{Naive strategy} & \textbf{Cost per midpoint} & \textbf{Failure mode} \\
\midrule
Sequence-level only  & $K$ energy calls & Cannot fix individual bad tokens \\
Position-factorized  & $|\mathcal{V}| \cdot L$ calls & Assumes $E(x) \approx \textstyle\sum_i e_i(x_i)$; rarely holds \\
%\bottomrule
\end{tabular}
\end{center}

% ---- Figure: The Granularity Dilemma ----
\begin{figure}[t]
\centering
\begin{tikzpicture}[
    scale=0.95, transform shape,
    tok/.style={rectangle, minimum width=0.58cm, minimum height=0.50cm,
                inner sep=0pt, font=\scriptsize, rounded corners=1pt},
    good/.style={tok, fill=energyblue!15, draw=energyblue!40},
    bad/.style={tok, fill=trajbad!18, draw=trajbad!45, font=\scriptsize\bfseries},
    neutral/.style={tok, fill=gray!10, draw=gray!35},
    lbl/.style={font=\scriptsize, anchor=east},
]

% ---- Column A: Sequence-level only ----
\node[font=\small\bfseries, color=baselinegray!80!black, align=center] at (3.0, 5.6)
  {Sequence-level only};
\node[font=\tiny, color=baselinegray, align=center] at (3.0, 5.1)
  {$K$ energy calls --- picks globally best};

% Candidate 2 (loser, shown first on top)
\node[lbl, font=\tiny, baselinegray] at (-0.1, 4.2) {$c_2$:};
\foreach \i in {0,...,7} {
    \node[neutral] at ({0.5+\i*0.7}, 4.2) {};
}
\node[font=\tiny, baselinegray, anchor=west] at (5.6, 4.2) {$E{=}-1.8$};

% Candidate 1 (winner but has hidden bad tokens, below c2)
\node[lbl, font=\tiny\bfseries, energyblue] at (-0.1, 3.5) {$c^*$:};
\node[good]  at (0.5, 3.5) {$o_1$};
\node[good]  at (1.2, 3.5) {$o_2$};
\node[bad]   at (1.9, 3.5) {\textcolor{trajbad!80}{$o_3$}};
\node[good]  at (2.6, 3.5) {$o_4$};
\node[good]  at (3.3, 3.5) {$o_5$};
\node[bad]   at (4.0, 3.5) {\textcolor{trajbad!80}{$o_6$}};
\node[good]  at (4.7, 3.5) {$o_7$};
\node[good]  at (5.4, 3.5) {$o_8$};

% Energy label
\node[font=\tiny, energyblue, anchor=west] at (5.6, 3.5) {$E{=}\mathbf{-2.1}$ \checkmark};

% Problem annotation (arrows point straight down from c*, no crossing)
\draw[-{Stealth[length=1.5mm]}, trajbad, thick] (1.9, 3.2) -- (1.9, 2.8);
\draw[-{Stealth[length=1.5mm]}, trajbad, thick] (4.0, 3.2) -- (4.0, 2.8);
\node[font=\tiny, trajbad, align=center, fill=white, inner sep=1.5pt] at (3.0, 2.15)
  {bad tokens hidden in globally best sequence};
\node[font=\tiny, trajbad, align=center] at (3.0, 2.5)
  {$\rightarrow$ errors compound at next midpoint};

% ---- Divider ----
\draw[baselinegray!30, thick] (-0.5, 1.8) -- (6.5, 1.8);

% ---- Column B: Token-level only ----
\node[font=\small\bfseries, color=baselinegray!80!black, align=center] at (3.0, 1.4)
  {Position-factorized};
\node[font=\tiny, color=baselinegray, align=center] at (3.0, 0.8)
  {$|\mathcal{V}|{\times}L$ energy calls --- intractable};

% Token grid with individual scoring
\foreach \i in {0,...,7} {
    \node[good] at ({0.5+\i*0.7}, 0.15) {};
}

% Per-token energy arrows (showing intractable cost)
\foreach \i in {0,...,7} {
    \draw[-{Stealth[length=1mm]}, baselinegray!50, thin] ({0.5+\i*0.7}, -0.15) -- ({0.5+\i*0.7}, -0.7);
}
\node[font=\tiny, baselinegray, align=center] at (3.0, -1.0)
  {$50\,257 \times 1\,024 \approx 51$M energy calls per midpoint};

% ---- Right side: C2F resolution ----
\draw[baselinegray!30, thick, dashed] (7.0, 5.8) -- (7.0, -1.3);

\node[font=\small\bfseries, color=trajgood!80!black, align=center] at (10.5, 5.6)
  {Sequence-to-Token (ours)};

% Phase 1 box (y: 3.8 – 5.1)
\draw[energyblue!30, fill=lightblue, rounded corners=3pt]
  (7.5, 3.8) rectangle (13.5, 5.1);
\node[font=\scriptsize\bfseries, energyblue!80!black, anchor=west] at (7.7, 4.8)
  {Phase 1: Sequence};
\node[font=\tiny, color=baselinegray, anchor=west, text width=5.3cm] at (7.7, 4.35)
  {$K$ candidates scored in one batched call.\\
   Selects globally best continuation.};
\node[font=\tiny, energyblue, anchor=west] at (7.7, 3.95)
  {Cost: $K$ energy calls};

% Arrow between Phase 1 and Phase 2 (gap = 1.0)
\draw[-{Stealth[length=2mm]}, thick, baselinegray!80!black] (10.5, 3.6) -- (10.5, 3.0);

% Phase 2 box (y: 1.2 – 2.8)
\draw[trajgood!30, fill=signalgreen, rounded corners=3pt]
  (7.5, 1.2) rectangle (13.5, 2.8);
\node[font=\scriptsize\bfseries, trajgood!80!black, anchor=west] at (7.7, 2.5)
  {Phase 2: Token};
\node[font=\tiny, color=baselinegray, anchor=west, text width=5.3cm] at (7.7, 2.05)
  {Velocity alignment fixes individual tokens.\\
   Free proxy --- logits already computed.};
\node[font=\tiny, trajgood, anchor=west] at (7.7, 1.45)
  {Cost: 0 extra forward passes + 1 safeguard};

% Arrow between Phase 2 and Safeguard (gap = 1.0, matching above)
\draw[-{Stealth[length=2mm]}, thick, baselinegray!80!black] (10.5, 1.0) -- (10.5, 0.4);

% Safeguard box (y: -1.0 – 0.2, gap to Phase 2 = 1.0)
\draw[shapegold!40, fill=shapegoldlight, rounded corners=3pt]
  (7.5, -1.0) rectangle (13.5, 0.2);
\node[font=\scriptsize\bfseries, shapegold!80!black, anchor=west] at (7.7, -0.1)
  {Energy Safeguard};
\node[font=\tiny, color=baselinegray, anchor=west, text width=5.3cm] at (7.7, -0.55)
  {Accept refinement only if $E_\phi(x_{\text{refined}}) \leq E_{\text{best}} + \epsilon$.\\
   Prevents local fixes from degrading global quality.};

\end{tikzpicture}
\caption{\textbf{The granularity dilemma.} \emph{Left:} Sequence-level scoring selects the globally best candidate but cannot fix individually bad tokens (in red). Position-factorized guidance fixes each token at a cost of millions of energy evaluations per midpoint. \emph{Right:} Sequence-to-Token Navigation resolves the tension---Phase~1 finds the best continuation via $K$ energy calls; Phase~2 refines individual tokens using velocity alignment at zero model-inference cost; the energy safeguard ensures refinement never degrades global quality.}
\label{fig:granularity-dilemma}
\end{figure}
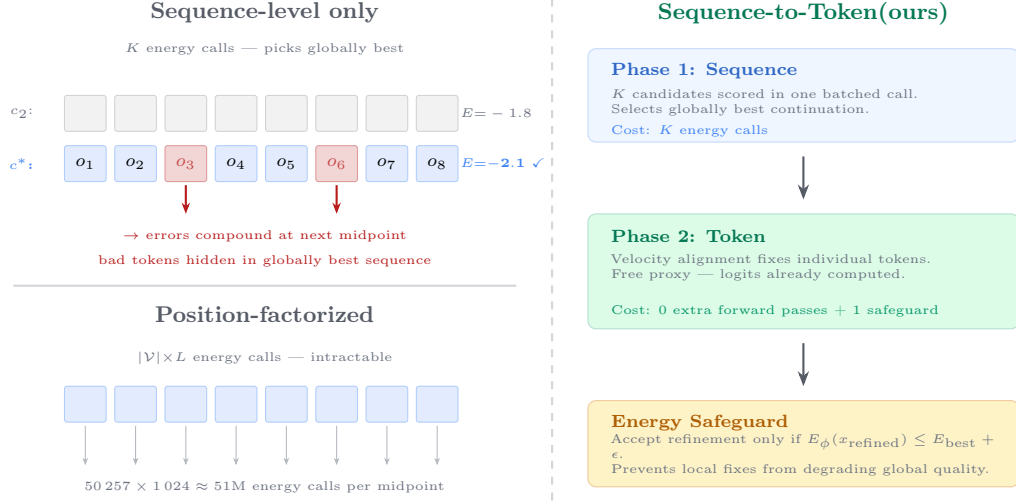

\noindent
One could score entire candidate sequences and select the best (sequence-level), which is affordable but coarse: the energy function evaluates holistically, so a candidate can win despite containing individually poor tokens simply because the rest of the sequence compensates.
Alternatively, one could assume the energy decomposes across positions (position-factorized), enabling per-token guidance, but assuming token independence is at odds with the energy model whose purpose is to capture global coherence.

\cref{fig:granularity-dilemma} illustrates the dilemma and how Sequence-to-Token Navigation sidesteps it by decomposing guidance into two complementary phases that operate at different granularities:
\begin{enumerate}
  \item \textbf{Sequence phase} (sequence-level): Generate $K$ diverse candidate jumps, score each with one energy call, and select the globally best candidate. This ensures the trajectory enters a low-energy region of sequence space.
  \item \textbf{Token phase} (fine-level): Refine the selected candidate using the velocity network's own confidence as a \emph{free proxy} for per-token quality (no additional energy calls). A single energy evaluation on the refined result serves as a safeguard against degradation.
\end{enumerate}

The key insight of the token phase is that the velocity model $v_\theta$ already encodes a distribution over plausible tokens at every position.
Positions where the selected candidate disagrees with a high-confidence velocity prediction are likely suboptimal; those already aligned with the velocity's top prediction are likely correct.
The velocity-alignment signal is available at zero cost since the logits are already computed and provides a principled, position-level refinement criterion that complements the sequence-level energy signal.

\subsection{Phase 1: sequence exploration via diverse candidate scoring}
\label{app:c2f_coarse}

\textbf{Temperature-diverse candidate generation.}
To ensure the $K$ candidates span a meaningful range of the trajectory space, we sample each at a different temperature~\citep{zhang2024edt,du2025optimizing}.
The base temperature adapts to the velocity's current uncertainty via its per-position entropy:
\begin{equation}
  T_{\text{base}} = T_{\min} + \Delta_T \cdot (1 - \tilde{H}),
  \label{eq:entropy_temp}
\end{equation}
where $\tilde{H} \in [0,1]$ is the batch-normalized mean entropy of the velocity distribution, $T_{\min}$ ($=0.8$) sets the minimum base temperature, and $\Delta_T$ ($=0.4$) controls the entropy-dependent range, so that $T_{\text{base}} \in [T_{\min},\; T_{\min} + \Delta_T]$.
The values $T_{\min}{=}0.8$ and $\Delta_T{=}0.4$ are chosen so that the base temperature stays in the near-unity regime: temperatures far below 1.0 would collapse diversity among candidates, while temperatures far above 1.0 would introduce too much noise into the CTMC jumps, producing candidates that are diverse but uniformly poor.
The range $[0.8, 1.2]$ keeps sampling close to the velocity's learned distribution while allowing meaningful variation.

This schedule is counterintuitive: when the velocity is \emph{confident} ($\tilde{H}$ low), we raise the temperature slightly to verify that no better alternatives exist; when the velocity is \emph{uncertain} ($\tilde{H}$ high), we lower it to concentrate on high-probability tokens rather than amplifying noise. Each of the $K$ candidates is sampled at temperature
\begin{equation}
  T_c = T_{\text{base}} \cdot \bigl(\lambda_{\min} + (\lambda_{\max} - \lambda_{\min}) \cdot \tfrac{c-1}{K-1}\bigr), \qquad c = 1, \ldots, K,
  \label{eq:temp_diversity}
\end{equation}
where the per-candidate multiplier spans $[\lambda_{\min},\, \lambda_{\max}]$ ($=[0.7,\, 1.3]$), producing effective temperatures in $[\lambda_{\min} T_{\text{base}},\; \lambda_{\max} T_{\text{base}}]$.
Low-temperature candidates ($\lambda_{\min}$) exploit the velocity's modal predictions; high-temperature candidates ($\lambda_{\max}$) explore the tails.
The symmetric spread of $\pm 30\%$ around $T_{\text{base}}$ reflects a balance: narrower spreads
produce near-identical candidates that waste the multi-candidate budget, while wider spreads
push high-temperature candidates into a regime where the sampled tokens are effectively random, contributing little useful signal to the energy-based selection. Each sampled prediction $\hat{x}_1^{(c)}$ is evolved under the standard CTMC jump dynamics to produce the candidate midpoint state $x^{(c)} = \text{CTMCJump}(x_t, \hat{x}_1^{(c)}, h)$.

\textbf{Energy-based selection.}
The energy compass scores all candidates in a single batch to identify the best candidate $x_{\text{best}} = x^{(c^*)}$ with energy $E_{\text{best}}$:
\begin{equation}
  c^* = \arg\min_{c \in \{1,\ldots,K\}} E_\phi(x^{(c)}).
  \label{eq:coarse_select}
\end{equation}

The sequence phase guarantees that the selected candidate has the lower \emph{global} energy.
Since the energy function evaluates sequences holistically, a sequence can win despite containing some weak tokens. The weak tokens, which may lead to compounding errors in subsequent RK-4 midpoints, go uncorrected (the red tokens in the left side of \cref{fig:granularity-dilemma}).

\subsection{Phase 2: token refinement via velocity-aligned token correction}
\label{app:c2f_fine}

The token phase corrects individual token-level weaknesses in $x_{\text{best}}$ using the velocity network's own predictions as a refinement signal---\emph{without any per-token energy calls}.

\textbf{Velocity alignment as a free proxy for token quality.}
The velocity network's logits at $x_t$ have already been computed as part of the RK-4 velocity evaluation---no additional forward pass is needed.
We reuse these logits to sample a fresh prediction $\hat{x}_1^{\text{ref}} \sim p_{1|t}^\theta(\cdot \mid x_t)$ and to extract the velocity's argmax prediction $x_{\text{top}}^{(i)} = \arg\max_v p_{1|t}^{(i)}(v \mid x_t)$ and its confidence $\text{conf}_i = \max_v p_{1|t}^{(i)}(v \mid x_t)$.
All three quantities---the refinement sample, the top prediction, and the confidence---are derived from the same cached logits via cheap categorical operations (sampling, argmax, and max), with zero model-inference cost.
The alignment signal compares the refinement proposal against the selected candidate at each position:
\begin{equation}
  \Delta_{\text{align}}^{(i)}
  = \mathbf{1}\!\bigl[\hat{x}_{1,i}^{\text{ref}} = x_{\text{top}}^{(i)}\bigr]
  - \mathbf{1}\!\bigl[x_{\text{best}}^{(i)} = x_{\text{top}}^{(i)}\bigr]
  \;\in\; \{-1,\, 0,\, +1\}.
  \label{eq:align_change}
\end{equation}
The interpretation is direct:
$\Delta_{\text{align}}^{(i)} = +1$ means the refinement proposal matches the velocity's most confident prediction while the current token does not---a strong signal to replace.
$\Delta_{\text{align}}^{(i)} = -1$ means the current token already matches but the proposal does not---leave it.
$\Delta_{\text{align}}^{(i)} = 0$ means no alignment difference---default to the base jump rate.

% ---- Figure: Fine-Phase Token-Level Detail ----
\begin{figure}[t]
\centering
\begin{tikzpicture}[
    scale=0.95, transform shape,
    tok/.style={rectangle, minimum width=0.9cm, minimum height=0.55cm,
                inner sep=0pt, font=\scriptsize, rounded corners=1.5pt},
    good/.style={tok, fill=energyblue!12, draw=energyblue!35},
    mismatch/.style={tok, fill=trajbad!15, draw=trajbad!50, font=\scriptsize\bfseries},
    fixed/.style={tok, fill=trajgood!18, draw=trajgood!50, font=\scriptsize\bfseries},
    keep/.style={tok, fill=gray!8, draw=gray!30},
    confbar/.style={thick, rounded corners=0.5pt},
]

% Title
\node[font=\small\bfseries, baselinegray!80!black] at (5.5, 6.0) {Token-Phase: Velocity-Aligned Token Correction};

% Position labels
\foreach \i/\lab in {0/pos 1, 1/pos 2, 2/pos 3, 3/pos 4, 4/pos 5, 5/pos 6} {
    \node[font=\tiny\bfseries, baselinegray] at ({\i*2.0 + 0.5}, 5.3) {\lab};
}

% --- Row 1: x_best (selected candidate) ---
\node[font=\scriptsize, baselinegray!80!black, anchor=east] at (-0.8, 4.6) {$x_{\text{best}}$};
\node[good]     at (0.5, 4.6) {the};
\node[good]     at (2.5, 4.6) {cat};
\node[mismatch] at (4.5, 4.6) {sit};
\node[good]     at (6.5, 4.6) {on};
\node[mismatch] at (8.5, 4.6) {a};
\node[good]     at (10.5, 4.6) {mat};

% --- Row 2: x_top (velocity argmax) ---
\node[font=\scriptsize, baselinegray!80!black, anchor=east] at (-0.8, 3.6) {$x_{\text{top}}$};
\node[tok, fill=trajgood!8, draw=trajgood!25]  at (0.5, 3.6) {the};
\node[tok, fill=trajgood!8, draw=trajgood!25]  at (2.5, 3.6) {cat};
\node[tok, fill=trajgood!15, draw=trajgood!45, font=\scriptsize\bfseries]
  at (4.5, 3.6) {sat};
\node[tok, fill=trajgood!8, draw=trajgood!25]  at (6.5, 3.6) {on};
\node[tok, fill=trajgood!15, draw=trajgood!45, font=\scriptsize\bfseries]
  at (8.5, 3.6) {the};
\node[tok, fill=trajgood!8, draw=trajgood!25]  at (10.5, 3.6) {mat};

% --- Confidence bars ---
\node[font=\scriptsize, baselinegray!80!black, anchor=east] at (-0.8, 2.7) {conf$_i$};
\foreach \i/\c/\w in {0/0.92/1.6, 1/0.85/1.5, 2/0.91/1.6, 3/0.78/1.35, 4/0.43/0.75, 5/0.88/1.55} {
    % Bar background
    \draw[gray!15, confbar, fill=gray!8]
      ({\i*2.0 - 0.3}, 2.45) rectangle ({\i*2.0 + 1.3}, 2.7);
    % Filled bar (color by confidence level)
    \pgfmathsetmacro{\fillcolor}{(\c > 0.8) ? "trajgood!50" : ((\c > 0.6) ? "shapegold!50" : "baselinegray!30")}
    \draw[\fillcolor, confbar, fill=\fillcolor]
      ({\i*2.0 - 0.3}, 2.45) rectangle ({\i*2.0 - 0.3 + \w}, 2.7);
    % Value
    \node[font=\tiny, baselinegray] at ({\i*2.0 + 0.5}, 2.25) {\c};
}

% --- Alignment signal ---
\node[font=\scriptsize, baselinegray!80!black, anchor=east] at (-0.8, 1.7) {$\Delta_{\text{align}}$};
\node[font=\scriptsize, baselinegray]                at (0.5, 1.7) {$0$};
\node[font=\scriptsize, baselinegray]                at (2.5, 1.7) {$0$};
\node[font=\scriptsize\bfseries, trajgood!80!black] at (4.5, 1.7) {$+1$};
\node[font=\scriptsize, baselinegray]                at (6.5, 1.7) {$0$};
\node[font=\scriptsize\bfseries, trajgood!80!black] at (8.5, 1.7) {$+1$};
\node[font=\scriptsize, baselinegray]                at (10.5, 1.7) {$0$};

% --- Decision ---
\node[font=\scriptsize, baselinegray!80!black, anchor=east] at (-0.8, 0.9) {Action};
\node[font=\tiny, baselinegray]                     at (0.5, 0.9) {keep};
\node[font=\tiny, baselinegray]                     at (2.5, 0.9) {keep};
\node[font=\tiny\bfseries, trajgood!80!black] at (4.5, 0.9) {replace};
\node[font=\tiny, baselinegray]                     at (6.5, 0.9) {keep};
\node[font=\tiny, baselinegray!60]                  at (8.5, 0.9) {keep};
\node[font=\tiny, baselinegray]                     at (10.5, 0.9) {keep};

% Annotation for pos 5: low confidence suppresses replacement
\draw[shapegold, thick, dashed, rounded corners=2pt]
  (7.8, 0.55) rectangle (9.2, 2.95);
\node[font=\tiny, shapegold!80!black, align=center, text width=2.2cm]
  at (8.5, 0.25) {low conf $\rightarrow$\\suppressed};

% Annotation for pos 3: high confidence triggers replacement
\draw[trajgood, thick, rounded corners=2pt]
  (3.8, 0.55) rectangle (5.2, 2.95);
\node[font=\tiny, trajgood!80!black, align=center, text width=2.2cm]
  at (4.5, 0.25) {high conf $\rightarrow$\\replaced};

% --- Result row ---
\draw[baselinegray!30, thick] (-1.0, -0.2) -- (11.5, -0.2);

\node[font=\scriptsize\bfseries, baselinegray!80!black, anchor=east] at (-0.8, -0.8) {$x_{\text{refined}}$};
\node[good]  at (0.5, -0.8) {the};
\node[good]  at (2.5, -0.8) {cat};
\node[fixed] at (4.5, -0.8) {sat};
\node[good]  at (6.5, -0.8) {on};
\node[mismatch] at (8.5, -0.8) {a};
\node[good]  at (10.5, -0.8) {mat};

\end{tikzpicture}
\caption{\textbf{Token-phase token-level detail.} Phase~1's $x_{\text{best}}$ is compared against the velocity's argmax prediction $x_{\text{top}}$ at each position. Where the two disagree ($\Delta_{\text{align}}{=}{+}1$), the replacement decision is modulated by velocity confidence: position~3 (``sit''$\to$``sat'') has high confidence (0.91) and is replaced; position~5 (``a''$\to$``the'') has low confidence (0.43) and is kept. \emph{Confidence gating} prevents overriding well-calibrated local predictions when velocity is uncertain.}
\label{fig:fine-phase-detail}
\end{figure}

\textbf{Why velocity alignment works as a proxy.}
The velocity network was trained on the data distribution and captures the conditional probability of each token given the current flow state.
At positions where the velocity is highly confident (large $\text{conf}_i$), its top prediction is a strong indicator of the ``correct'' token under the data distribution.
A candidate token that disagrees with a high-confidence velocity prediction is statistically more likely to be an artifact of the stochastic CTMC jump than a genuine feature of the target sequence.
By using this signal, which needs no additional forward pass, we obtain position-level guidance that is complementary to the energy model's sequence-level assessment.
\cref{fig:fine-phase-detail} illustrates this mechanism on a concrete example, showing how confidence gating prevents overriding uncertain positions.

\textbf{Adaptive guidance strength.}
The token refinement modulates jump probabilities using a guidance coefficient $\beta_{\text{adp}}$ that controls how strongly the velocity-alignment signal influences token replacement decisions.
A fixed coefficient throughout the flow trajectory is suboptimal: at low $t$ (mostly noise), the velocity network's predictions are uncertain and the alignment signal should carry more weight; at high $t$ (mostly revealed text), the velocity is increasingly reliable on its own and heavy-handed modulation risks overriding well-calibrated local predictions.
We therefore decay the guidance intensity:
\begin{equation}
  \beta_{\text{adp}} = \beta_0 \cdot (1 - \alpha_\beta \, t),
  \label{eq:adaptive_beta}
\end{equation}
where $\alpha_\beta$ ($=0.5$) controls the decay rate, yielding full strength $\beta_0$ at $t{=}0$ and $\beta_0/2$ at $t{=}1$.
The linear form is the simplest monotone schedule, and the choice $\alpha_\beta{=}0.5$ ensures guidance is never fully switched off---even at late timesteps where the sequence is mostly revealed, a moderate energy signal helps avoid last-mile errors at the remaining unrevealed positions.

\paragraph{Confidence-modulated jump probability.}
The base CTMC jump probability $p_{\text{base}}^{(i)} = 1 - \exp(-h \cdot \lambda_{\text{base}})$ (where $\lambda_{\text{base}} = \dot{\kappa}_t / (1 - \kappa_t)$, applied only at positions where the refinement proposes a different token) is modulated by the alignment signal:

\begin{equation}
  p_{\text{jump}}^{(i)} = \text{clamp}\!\Bigl(
    p_{\text{base}}^{(i)} \cdot \exp\!\bigl(\Delta_{\text{align}}^{(i)} \cdot \text{conf}_i \cdot \beta_{\text{adp}} \cdot \alpha_{\text{soft}}\bigr),\;
    0,\; p_{\max}
  \Bigr),
  \label{eq:adjusted_jump}
\end{equation}
where $\alpha_{\text{soft}}$ ($=0.5$) is a softening factor and $p_{\max}$ ($=0.95$) caps the jump probability.
Three design choices in \cref{eq:adjusted_jump} merit explanation:
\begin{itemize}
  \item \textbf{Confidence gating} ($\text{conf}_i$): The modulation magnitude scales with the velocity's certainty at that position. At uncertain positions, the alignment signal is unreliable and the adjustment is automatically suppressed, preventing overconfident refinement of ambiguous tokens (see position~5 in \cref{fig:fine-phase-detail} for a concrete example).
  \item \textbf{Softening factor} $\alpha_{\text{soft}}$ ($=0.5$): Without softening ($\alpha_{\text{soft}}{=}1$), the exponential modulation can produce extreme adjustments---a position with $\text{conf}_i{=}0.9$ and $\beta_{\text{adp}}{=}1$ would scale the jump rate by $\exp(\pm 0.9) \approx 2.5\times$ or $0.4\times$. The factor $\alpha_{\text{soft}}{=}0.5$ halves the exponent, keeping rate adjustments within a moderate range ($\exp(\pm 0.45) \approx 1.6\times$ or $0.6\times$) that steers without overwhelming the base dynamics.
  \item \textbf{Upper bound} $p_{\max}$ ($=0.95$): Capping below 1.0 preserves stochasticity. Deterministic jumps ($p{=}1$) risk oscillations between competing token choices at successive midpoints and remove the randomness needed for the CTMC dynamics. The value $0.95$ is the standard near-deterministic cap used in discrete sampling; any value in $[0.9, 0.99]$ would serve the same purpose.
\end{itemize}

Tokens are then stochastically replaced according to the adjusted probabilities:
\begin{equation}
  x_{\text{refined}}^{(i)} = \begin{cases}
    \hat{x}_{1,i}^{\text{ref}} & \text{with probability } p_{\text{jump}}^{(i)}, \\
    x_{\text{best}}^{(i)} & \text{otherwise.}
  \end{cases}
  \label{eq:refine_apply}
\end{equation}

\subsection{Energy safeguard}
\label{app:c2f_safeguard}

The token-level refinement in Phase~2 is guided by velocity alignment, which is a proxy for---but not identical to---sequence-level energy.
It is possible (though uncommon in practice) that a set of individually well-aligned token replacements produces a sequence with higher energy when considered jointly.
A single energy evaluation on the refined sequence provides a safety net:
\begin{equation}
  x'_{\text{mid}} = \begin{cases}
    x_{\text{refined}} & \text{if } E_\phi(x_{\text{refined}}) \leq E_{\text{best}} + \epsilon_{\text{safe}}, \\
    x_{\text{best}} & \text{otherwise,}
  \end{cases}
  \label{eq:safeguard}
\end{equation}
with tolerance $\epsilon_{\text{safe}}$ ($=0.1$).
The tolerance accommodates energy-estimation noise: a refinement that slightly increases estimated energy may still be genuinely beneficial (e.g., improving velocity alignment at the cost of a negligible energy increase within the estimator's error margin).
The value $\epsilon_{\text{safe}}{=}0.1$ is small relative to the typical energy range across candidates (which spans several units in our experiments), so it permits only marginal degradation while avoiding the brittleness of an exact $\leq$ check on a noisy scalar.
In our experiments, the safeguard triggers on fewer than 8\% of midpoints, confirming that velocity-aligned refinement is well-correlated with energy improvement.

\subsection{Full algorithm}
\label{app:c2f_algorithm}

\cref{alg:c2f_navigate} consolidates the complete $\textsc{Navigate}$ procedure described in the preceding subsections.
Phase~1 (lines 4--12) generates $K$ temperature-diverse candidates and selects the lowest-energy one via a single batched energy call.
Phase~2 (lines 13--19) refines the winner through velocity-aligned token replacement, modulated by the adaptive guidance coefficient $\beta_{\text{adp}}$ (\cref{eq:adaptive_beta}) and the velocity's per-position confidence.
The energy safeguard (lines 20--24) accepts the refinement only if it does not significantly increase energy, falling back to the Phase~1 winner otherwise.
All symbols and hyperparameters are summarized in \cref{tab:c2f_hyperparams}.

\begin{algorithm}[t]
\caption{Sequence-to-Token compass navigation ($\textsc{Navigate}$)}
\label{alg:c2f_navigate}
\begin{algorithmic}[1]
\REQUIRE State $x_t$, velocity logits, energy compass $E_\phi$, time $t$, step $h$, threshold $\tau$, candidates $K$
\REQUIRE Hyperparameters: $T_{\min}$, $\Delta_T$, $\lambda_{\min}$, $\lambda_{\max}$, $\alpha_\beta$, $\alpha_{\text{soft}}$, $p_{\max}$, $\epsilon_{\text{safe}}$
\IF{$t < \tau$}
  \RETURN StandardCTMCJump($x_t$, logits, $h$) \COMMENT{Below threshold: no guidance}
\ENDIF

\STATE \textit{// --- Phase 1: Sequence exploration ---}
\STATE Compute per-position entropy $H$ from logits
\STATE $T_{\text{base}} \leftarrow T_{\min} + \Delta_T \cdot (1 - \tilde{H})$
\FOR{$c = 1, \ldots, K$}
  \STATE $T_c \leftarrow T_{\text{base}} \cdot (\lambda_{\min} + (\lambda_{\max} - \lambda_{\min}) \cdot (c-1)/(K-1))$
  \STATE $\hat{x}_1^{(c)} \sim \text{softmax}(\text{logits} / T_c)$;\quad $x^{(c)} \leftarrow \text{CTMCJump}(x_t, \hat{x}_1^{(c)}, h)$
\ENDFOR
\STATE $\mathbf{E} \leftarrow E_\phi(x^{(1:K)})$;\quad $c^* \leftarrow \arg\min_c \mathbf{E}^{(c)}$
\STATE $x_{\text{best}} \leftarrow x^{(c^*)}$;\quad $E_{\text{best}} \leftarrow \mathbf{E}^{(c^*)}$

\STATE \textit{// --- Phase 2: Token refinement ---}
\STATE $\hat{x}_1^{\text{ref}} \sim p_{1|t}^\theta(\cdot \mid x_t)$;\quad $x_{\text{top}} \leftarrow \arg\max_v\, p_{1|t}^{(i)}(v \mid x_t)$;\quad $\text{conf} \leftarrow \max_v\, p_{1|t}^{(i)}(v \mid x_t)$
\STATE $\beta_{\text{adp}} \leftarrow \beta_0 \cdot (1 - \alpha_\beta\,t)$
\STATE $\Delta_{\text{align}}^{(i)} \leftarrow \mathbf{1}[\hat{x}_{1,i}^{\text{ref}} = x_{\text{top}}^{(i)}] - \mathbf{1}[x_{\text{best}}^{(i)} = x_{\text{top}}^{(i)}]$
\STATE $p_{\text{base}}^{(i)} \leftarrow \mathbf{1}[\hat{x}_{1,i}^{\text{ref}} \neq x_{\text{best}}^{(i)}] \cdot (1 - \exp(-h \cdot \lambda_{\text{base}}))$
\STATE $p_{\text{jump}}^{(i)} \leftarrow \text{clamp}\!\bigl(p_{\text{base}}^{(i)} \cdot \exp(\Delta_{\text{align}}^{(i)} \cdot \text{conf}_i \cdot \beta_{\text{adp}} \cdot \alpha_{\text{soft}}),\; 0,\; p_{\max}\bigr)$
\STATE $x_{\text{refined}}^{(i)} \leftarrow \begin{cases} \hat{x}_{1,i}^{\text{ref}} & \text{w.p. } p_{\text{jump}}^{(i)}, \quad x_{\text{best}}^{(i)} \text{ otherwise} \end{cases}$

\STATE \textit{// --- Energy safeguard ---}
\STATE $E_{\text{ref}} \leftarrow E_\phi(x_{\text{refined}})$
\IF{$E_{\text{ref}} \leq E_{\text{best}} + \epsilon_{\text{safe}}$}
  \RETURN $x_{\text{refined}}$
\ELSE
  \RETURN $x_{\text{best}}$
\ENDIF
\end{algorithmic}
\end{algorithm}

\begin{table}[h]
\centering
\small
\caption{Sequence-to-Token compass navigation hyperparameters with default values and rationale. Only $\beta_0$ and $\epsilon_{\text{safe}}$ are swept (\cref{app:nav_sensitivity}), and $K$ is ablated separately (\cref{app:k_ablation}). The remaining constants, including the temperature range ($T_{\min}$, $\Delta_T$, $\lambda_{\min}$, $\lambda_{\max}$), are set structurally and not swept.}
\label{tab:c2f_hyperparams}
\begin{tabular}{@{}clll@{}}
%\toprule
\\
\textbf{Symbol} & \textbf{Value} & \textbf{Role} & \textbf{Rationale} \\
\midrule
$K$ & 5 & Candidates per midpoint & Balances diversity vs.\ cost (\cref{tab:c2f_cost}) \\
$T_{\min}$ & 0.8 & Min base temperature & Keeps sampling near the velocity's mode \\
$\Delta_T$ & 0.4 & Entropy-dependent range & Bounds $T_{\text{base}} \in [0.8, 1.2]$, near-unity regime \\
$\lambda_{\min}$ & 0.7 & Per-candidate temp lower & $\pm 30\%$ spread: diverse yet informative \\
$\lambda_{\max}$ & 1.3 & Per-candidate temp upper & (see text) \\
$\beta_0$ & 0.3 & Base guidance strength & Scales token-refinement modulation (\cref{eq:adaptive_beta}) \\
$\alpha_\beta$ & 0.5 & Guidance decay rate & Halves guidance at $t{=}1$; never fully off \\
$\alpha_{\text{soft}}$ & 0.5 & Jump-rate softening & Keeps rate adjustments in moderate range \\
$p_{\max}$ & 0.95 & Max jump probability & Preserves CTMC stochasticity \\
$\epsilon_{\text{safe}}$ & 0.1 & Safeguard tolerance & Accommodates energy-estimation noise \\
%\bottomrule
\end{tabular}
\end{table}

\paragraph{Hyperparameter choices.}
We did not tune the navigation constants by search. A one-factor-at-a-time sweep of $\beta_0$ and $\epsilon_{\text{safe}}$ at inference shows a narrow spread (\cref{app:nav_sensitivity}). The two values that materially affect quality, $K$ and $\tau$, are ablated in the paper. The remainder are chosen from structural requirements of the method---keeping sampling near the velocity's distribution, ensuring guidance never fully switches off, capping jumps below determinism to preserve stochasticity---and are held fixed across all experimental settings: 170M and 1.3B, uniform and mask sources, both initializations. The same constants work unchanged at two model scales and two source distributions, which, together with that sweep, is our evidence that they are not tuned artifacts.

\subsection{Why both phases are necessary}
\label{app:c2f_complementarity}

The two phases address fundamentally different failure modes of trajectory construction, and neither alone suffices. \cref{fig:granularity-dilemma}, left panel, illustrates the failure mode of each naive strategy in isolation.

\paragraph{Phase~1 (sequence) alone is insufficient.}
Best-of-$K$ selection by energy is powerful but blunt.
The energy function scores sequences holistically: a candidate can win despite containing individually weak tokens, provided the remainder of the sequence compensates.
These uncorrected weak tokens become compounding errors through subsequent RK-4 midpoints---precisely the ``blind navigation'' problem that navigation shaping aims to solve.
Sequence selection reduces the \emph{frequency} of such errors but cannot eliminate them.

\paragraph{Phase~2 (token) alone is insufficient.}
Token-level refinement via velocity alignment is cheap and fine-grained, but it has no access to the energy landscape.
Velocity confidence is a useful proxy for per-token quality, but it is not a substitute for energy: the velocity was trained to predict the data distribution, not to minimize the compass energy.
Without a good starting point from Phase~1, local refinement would be polishing the wrong candidate---a trajectory that is locally coherent at each position but globally incoherent.

\paragraph{Together: global selection + local correction.}
Phase~1 ensures the trajectory enters a globally low-energy region of sequence space.
Phase~2 then makes targeted, confidence-weighted corrections within that region---fixing the individual token-level weaknesses that the holistic energy score cannot isolate.
The energy safeguard (\cref{eq:safeguard}) ensures that local refinement never degrades the globally low-energy state identified by Phase~1.
This sequence-to-token decomposition applies a natural two-stage principle: first select the best global continuation, then refine it locally.

\subsection{Computational cost and practical considerations}
\label{app:c2f_cost}

\subsubsection{Forward-pass budget}
\label{app:c2f_forward_pass}

Each call to $\textsc{Navigate}$ requires $K$ energy evaluations for sequence scoring (parallelized in a single batched call) plus one energy evaluation for the safeguard, totaling $K + 1$ energy forward passes per navigated midpoint. With our default $K{=}5$, this is 6 energy calls per midpoint, or up to $3 \times 6 = 18$ calls per RK-4 step when all three midpoints are navigated. The energy model (${\sim}$90M parameters) is $1.9\times$ smaller than the 170M velocity model, so each energy call costs proportionally fewer FLOPs.

Two factors temper this overhead when amortized across training. First, not every training step invokes RK-4: following the FS-DFM training schedule~\citep{fsdfm2025}, roughly 10 out of every 30 training steps use the RK-4 estimator, while the remaining 20 steps operate in the shortcut regime where the target is the ground-truth $x_1$ directly---requiring neither trajectory construction nor energy guidance. Second, within each RK-4 step, the time-threshold mechanism (\cref{sec:threshold}) skips navigation at midpoints where $t < \tau$. \cref{tab:c2f_cost} decomposes the resulting per-step cost and amortization. In the worst case (all midpoints navigated), the theoretical overhead is ${\sim}2.4\times$ FS-DFM. In practice this is an upper bound---the time threshold $\tau$ skips some midpoints entirely, and for $K{<}5$ the energy cost shrinks proportionally. Measured wall-clock overhead and analysis are in the main paper.

\begin{table}[h]
\centering
\small
\caption{Theoretical cost decomposition for navigation shaping at the 170M scale with default $K{=}5$. FLOPs are expressed relative to a single velocity forward pass on the 170M student. The amortization accounts for the RK-4/shortcut split in the training schedule.}
\label{tab:c2f_cost}
\begin{tabular}{@{}lccrr@{}}
%\toprule
\\
& \multicolumn{2}{c}{\textbf{Model calls}} & \multicolumn{2}{c}{\textbf{Relative FLOPs}} \\
\cmidrule(lr){2-3} \cmidrule(lr){4-5}
\textbf{Component} & \textbf{Velocity} & \textbf{Energy} & \textbf{FS-DFM} & \textbf{\ours} \\
\midrule
\multicolumn{5}{@{}l}{\textit{Per RK-4 step ($\sim$10 of 30 training steps):}} \\[2pt]
Velocity (4 semi-teacher + 1 student)                  & 5 & 0        & $5.0\times$ & $5.0\times$ \\
Energy: sequence scoring ($K{=}5 \times \leq$3 midpts) & 0 & $\leq 15$ & ---         & $\leq 7.9\times$ \\
Energy: safeguard ($\leq$3 midpoints)                  & 0 & $\leq 3$  & ---         & $\leq 1.6\times$ \\
\cmidrule(lr){1-5}
RK-4 step subtotal                                     & 5 & $\leq 18$ & $5.0\times$ & ${\sim}14.5\times$ \\
\midrule
\multicolumn{5}{@{}l}{\textit{Per shortcut step ($\sim$20 of 30 training steps):}} \\[2pt]
Velocity (student only)                                & 1 & 0 & $1.0\times$ & $1.0\times$ \\
\midrule
\multicolumn{5}{@{}l}{\textit{Amortization} $(10 \cdot \text{RK-4} + 20 \cdot \text{shortcut})/30$:} \\[2pt]
Amortized per step                                     & & & $2.33\times$ & ${\sim}5.5\times$ \\
\cmidrule(lr){1-5}
\textbf{Theoretical TS-DFM overhead vs.\ FS-DFM}       & & & \multicolumn{2}{c}{$\mathbf{\sim\!2.4\times}$} \\
%\bottomrule
\end{tabular}
\end{table}

%=============================================================================
% APPENDIX: Empirical Validation of the Navigation Policy
%=============================================================================

\section{Empirical validation of the navigation policy}
\label{app:energy_validation}

The energy compass and Sequence-to-Token (S2T) navigation policy are used exclusively at training time in our main experiments.
To validate these components independently of the training dynamics, we apply them \emph{at inference time}---directly guiding the baseline FS-DFM student's generation process.
Inference-time guidance is too slow for practical deployment (\cref{app:c2f_cost}), but it provides a controlled testbed: the student's weights are held fixed, so any improvement is attributable solely to the compass and navigation policy.
All experiments in this section use the same 170M-parameter baseline FS-DFM student trained with the \emph{uniform} source distribution and \emph{without} trajectory shaping.

The energy compass itself is validated separately in \cref{app:energy_compass_validation}, where it achieves 98.5--99.8\% corruption-detection accuracy and perfect bin-level monotonicity across both source distributions.
The experiments below focus on whether the navigation policies built on top of the compass---sequence selection, token refinement, and their combination---translate these properties into actual generation improvements.

\subsection{Navigation policy comparison across step counts}
\label{app:policy_comparison}

We compare four navigation policies applied to the same baseline FS-DFM student: \emph{no guidance} (standard blind CTMC jumps), \emph{sequence only} ($K{=}5$ best-of-$K$ selection), \emph{token only} (velocity-aligned refinement without candidate selection), and \emph{Sequence-to-Token (S2T)} (the full two-phase policy). All four use identical student weights and the same energy compass; they differ only in how midpoint states are constructed. We apply the policies at inference time so that any perplexity difference is attributable solely to the policy itself, with student weights held fixed. We evaluate at 8, 16, 32, and 64 inference steps: the 8-step setting is the student's final inference target, and the larger step counts probe how each policy scales when steps are smaller and more numerous.

\begin{table}[ht]
\centering
\small
\caption{\textbf{Navigation policy comparison across step counts (inference time).} All policies use the same baseline student and energy compass; they differ only in how midpoint states are constructed during generation.}
\label{tab:policy_comparison}
\setlength{\tabcolsep}{4pt}
\begin{tabular}{@{}llcccc@{}}
%\toprule
\\
\textbf{Steps} & \textbf{Policy} & \textbf{Ent} & \textbf{PPL (G$^2$-L)}$\downarrow$ & \textbf{PPL (L$^2$-7B)}$\downarrow$ & \textbf{PPL (L$^3$-8B)}$\downarrow$ \\
\midrule
\multirow{4}{*}{8}
  & No guidance           & 7.32 & 89.6 & 58.2 & 109.7 \\
  & Sequence only ($K{=}5$) & 6.99 & \textbf{81.2} & \textbf{52.2} & \textbf{99.6} \\
  & Token only             & 7.04 & 90.7 & 62.7 & 106.8 \\
  & Sequence-to-Token        & 6.93 & 86.2 & 58.3 & 104.3 \\
\midrule
\multirow{4}{*}{16}
  & No guidance           & 7.51 & 70.4 & 47.3 & 88.7 \\
  & Sequence only ($K{=}5$) & 7.21 & 51.9 & \textbf{33.3} & 61.3 \\
  & Token only             & 7.29 & 58.9 & 41.7 & 69.3 \\
  & Sequence-to-Token        & 7.10 & \textbf{50.8} & 36.3 & \textbf{59.5} \\
\midrule
\multirow{4}{*}{32}
  & No guidance           & 7.70 & 70.6 & 39.7 & 68.1 \\
  & Sequence only ($K{=}5$) & 7.40 & 53.3 & 36.5 & 58.4 \\
  & Token only             & 7.39 & 46.8 & 35.7 & 55.4 \\
  & Sequence-to-Token        & 7.26 & \textbf{43.1} & \textbf{31.2} & \textbf{49.1} \\
\midrule
\multirow{4}{*}{64}
  & No guidance           & 7.82 & 75.0 & 41.1 & 72.0 \\
  & Sequence only ($K{=}5$) & 7.50 & 52.0 & 31.5 & 51.3 \\
  & Token only             & 7.54 & 47.6 & 31.0 & 47.1 \\
  & Sequence-to-Token        & 7.36 & \textbf{38.0} & \textbf{24.4} & \textbf{37.0} \\
%\bottomrule
\end{tabular}
\end{table}

\paragraph{Key observations.}

(1) At 8 steps, differences are modest.
At the student's target inference budget (8 steps), all navigation policies produce similar perplexity: the gap between no guidance and the best policy is ${\sim}$9\% on GPT-2 Large (89.6 $\to$ 81.2) and ${\sim}$10\% on LLaMA-2 7B (58.2 $\to$ 52.2).
With so few steps, each step covers a large portion of the flow trajectory, leaving little room for per-step guidance to accumulate improvements.
Sequence selection alone provides the largest benefit in this regime by avoiding the worst stochastic realizations at each of the few available decision points.

(2) At higher step counts, S2T dominates.
Starting at 16 steps, S2T consistently achieves the best or near-best perplexity across all three evaluators.
The advantage grows with step count: at 32 steps, S2T reduces GPT-2 Large perplexity by 39\% relative to no guidance (70.6 $\to$ 43.1) and outperforms both sequence-only (53.3) and fine-only (46.8).
At 64 steps---the regime most relevant to RK-4 midpoint construction---S2T achieves 38.0 GPT-2 Large perplexity, a 49\% reduction from the unguided baseline and 20\% better than fine-only (47.6).
This confirms the complementarity argument from \cref{app:c2f_complementarity}: neither phase alone matches their combination.

(3) Unguided generation degrades at high step counts.
Counterintuitively, the unguided baseline's perplexity \emph{worsens} from 32 to 64 steps (GPT-2 Large: 70.6 $\to$ 75.0).
More steps mean more stochastic jumps, each introducing error; without guidance, errors compound and eventually outweigh the benefit of finer-grained integration.
All guided policies avoid this degradation entirely, with S2T continuing to improve from 32 to 64 steps (43.1 $\to$ 38.0 on GPT-2 Large).
This divergence underscores exactly why navigation shaping is needed during RK-4 target construction: the blind-jump regime where training operates is precisely where compounding errors are most damaging.

(4) The two phases have complementary strengths across step regimes.
At 8 steps, sequence selection dominates: with few decision points, picking the globally best candidate among $K$ options is more impactful than token-level corrections diluted across large steps.
At 16 steps, S2T begins to overtake sequence-only as token refinement accumulates enough small corrections to matter.
By 32--64 steps, the combination is decisive: sequence selection ensures each midpoint starts in a low-energy region, and token refinement corrects the individual token-level errors that sequence scoring cannot isolate---yielding improvements that neither phase achieves alone.
This is precisely the operating regime of RK-4 midpoint construction during training, where the S2T design provides its largest advantage.

\paragraph{Connection to training-time dynamics.}
The behavior at 16--64 inference steps is more than a stress test---it is a proxy for how S2T behaves when it runs inside RK-4 target construction during training. Training a student for 8-step generation ($h{=}0.125$) requires RK-4 midpoints at sub-step sizes of $h/2 = 0.0625$, corresponding to the 16--32-step regime, and S2T is called at each midpoint. The inference-time results at 16--64 steps therefore reflect the regime in which S2T actually operates when it shapes the student's training signal. Two of the patterns above are directly informative in this light: S2T's growing advantage with step count (observation~2) predicts that its benefit compounds inside RK-4 chains, and the unguided baseline's high-step degradation (observation~3) shows exactly the failure mode navigation shaping prevents.

\subsection{Token refinement behavior across step counts}
\label{app:refinement_behavior}

To understand what the token refinement phase actually does, we collect refinement statistics over 504 sequences at each step count.
Recall that at each guided midpoint, the token phase proposes token-level replacements based on velocity alignment (\cref{app:c2f_fine}), then the energy safeguard decides whether to accept or reject the refined sequence (\cref{app:c2f_safeguard}).
We measure four quantities:
\begin{itemize}
  \item \textbf{Accept \%}: The fraction of midpoints where the energy safeguard accepts the refined sequence, i.e., the refinement does not increase energy beyond the tolerance $\epsilon_{\text{safe}}$. A high rate indicates that velocity-aligned corrections are well-correlated with energy improvement.
  \item \textbf{Tokens / step}: The mean number of tokens replaced per midpoint by the token phase (out of $1\,024$ total positions). This measures how aggressive the refinement is---lower values indicate more targeted correction.
  \item \textbf{$\Delta E_{\text{acc}}$}: The mean energy change for accepted refinements. Negative values confirm that accepted refinements genuinely improve sequence quality.
  \item \textbf{$\Delta E_{\text{rej}}$}: The mean energy change for rejected refinements. Positive values confirm that the safeguard correctly blocks modifications that would have degraded quality.
\end{itemize}

\cref{tab:refinement_stats} reports these metrics across step counts.
The results reveal a consistent pattern: as step count increases and each step covers a smaller portion of the trajectory, refinement becomes more selective, more reliable, and smaller in magnitude.

\begin{table}[ht]
\centering
\small
\caption{\textbf{Token refinement statistics} (504 sequences per step count).}
\label{tab:refinement_stats}
\setlength{\tabcolsep}{5pt}
\begin{tabular}{@{}rcccc@{}}
%\toprule
\\
\textbf{Steps} & \textbf{Accept \%} & \textbf{Tokens / step (\%)} & \textbf{$\Delta E_{\text{acc}}$} & \textbf{$\Delta E_{\text{rej}}$} \\
\midrule
8  & 94.7\% & 79.4\;\;(7.8\%) & $-$0.081 & $+$0.629 \\
16 & 97.8\% & 36.1\;\;(3.5\%) & $-$0.041 & $+$0.757 \\
64 & 99.9\% & \phantom{0}8.2\;\;(0.8\%) & $-$0.010 & $+$0.348 \\
%\bottomrule
\end{tabular}
\end{table}

The results reveal several consistent patterns.
First, refinement is selective at every step count: even at 8 steps, only 7.8\% of tokens are replaced per step, dropping to 0.8\% at 64 steps.
This confirms the design intent of targeted correction rather than wholesale rewriting.
Second, the safeguard acceptance rate increases monotonically from 94.7\% at 8 steps to 99.9\% at 64 steps, indicating that velocity-aligned token replacements are increasingly well-correlated with energy improvement as step size shrinks---the velocity network's confidence is a more reliable proxy for token quality when each step is small.
Third, rejected refinements would have \emph{increased} energy at every step count ($\Delta E_{\text{rej}} > 0$), confirming that the safeguard correctly identifies and blocks the small fraction of refinements that would degrade sequence quality.

\subsection{Effect of candidate count K}
\label{app:k_ablation}
 
The number of candidates $K$ in the sequence phase controls the trade-off between navigation quality and computational cost: more candidates increase the chance of finding a low-energy midpoint, but each requires an additional energy forward pass.
We ablate $K$ at inference time by applying S2T guidance with varying candidate counts to the baseline student (\cref{tab:k_ablation}).
All experiments generate 504 sequences at 8 steps on a single GPU.
Wall-clock time measures the total generation time for all 504 sequences (excluding perplexity evaluation), synchronized across GPU operations.
 
\begin{table}[ht]
\centering
\small
\caption{\textbf{Effect of candidate count $K$ (S2T, 8 steps).}}
\label{tab:k_ablation}
\setlength{\tabcolsep}{4pt}
\begin{tabular}{@{}cccccc@{}}
%\toprule
\\
$K$ & \textbf{Ent} & \textbf{PPL (G$^2$-L)}$\downarrow$ & \textbf{PPL (L$^2$-7B)}$\downarrow$ & \textbf{PPL (L$^3$-8B)}$\downarrow$ & \textbf{Wall time (s)} \\
\midrule
\multicolumn{1}{@{}l}{No guidance} & 7.26 & 90.8 & 59.4 & 115.9 & 31.0 \\
\midrule
2 & 6.56 & 61.2 & 49.9 & 88.7 & 147.8 \\
3 & 6.50 & 59.3 & 41.8 & 75.7 & 185.7 \\
\rowcolor{selectedrow}
5 & 6.68 & \textbf{56.0} & \textbf{37.8} & \textbf{66.2} & 261.9 \\
8 & 6.57 & 57.6 & 40.3 & 70.6 & 376.5 \\
10 & 6.56 & 57.4 & 40.4 & 69.9 & 449.1 \\
%\bottomrule
\end{tabular}
\end{table}
 
The results show clear diminishing returns.
The largest improvement comes from introducing candidate selection at all: $K{=}2$ reduces LLaMA-3 8B PPL by 23\% relative to no guidance (115.9 $\to$ 88.7).
Each subsequent increase in $K$ yields a progressively smaller gain---$K{=}3$ reaches 75.7 (35\% reduction) and $K{=}5$ reaches 66.2 (43\% reduction)---while $K{=}8$ provides no additional benefit (70.6), confirming that $K{=}5$ is an effective default.
 
\paragraph{Inference-time cost vs.\ training-time cost.}
The wall-clock times in \cref{tab:k_ablation} reflect inference-time guidance, where the energy compass runs at \emph{every} step of generation---producing the $8.5\times$ slowdown observed at $K{=}5$.
This cost structure is fundamentally different from our actual approach, where all navigation shaping occurs during training and inference is completely unchanged: the shaped student generates with the same architecture, forward pass, and speed as a standard distilled model.
 
At training time, the cost of navigation shaping is further reduced by three factors.
First, the FS-DFM training schedule allocates only approximately one-third of training steps to the RK-4 regime (large step sizes where trajectory construction is needed); the remaining two-thirds use direct teacher targets with no trajectory construction and therefore no energy guidance (\cref{sec:fsdfm}).
Second, within each RK-4 step, the time-threshold mechanism (\cref{sec:threshold}) activates navigation only at midpoints where $t \geq \tau$; midpoints in the source-dominated regime ($t < \tau$) use standard blind jumps at zero additional cost.
Third, the energy compass (${\sim}$90M parameters) is substantially smaller than the velocity model, and the $K$ candidate evaluations are batched into a single GPU call.
Together, these factors reduce the amortized training overhead to ${\sim}2\times$ at the 170M scale (see the full cost breakdown in \cref{app:c2f_cost}).

%=============================================================================
% APPENDIX: Additional Experiments
%
% Requires in preamble:
%   \definecolor{selectedrow}{RGB}{220, 245, 220}
%=============================================================================

\section{Additional experiments}
\label{app:additional_experiments}

This section presents experiments omitted from the main text due to space constraints.
All experiments use the 170M-parameter model with the uniform source distribution unless otherwise noted.

\subsection{Effect of time threshold $\tau$: additional details}
\label{app:threshold_ablation}

The main results and table for the threshold ablation are in \cref{sec:threshold_ablation}. This section provides additional analysis.

\paragraph{Why $\tau{=}0.02$ rather than $\tau{=}0$.}
We use $\tau{=}0.02$ as the lowest threshold instead of $\tau{=}0$ for two reasons.
First, at $t{=}0$ the sequence is pure source noise---$\alpha(0){=}0$, so no positions are revealed---and the energy compass receives a sequence with no meaningful content to score; all candidates are indistinguishable noise, making energy-based selection equivalent to random selection.
Second, at very early flow times the CTMC transition rates involve the ratio $\dot{\kappa}_t / (1 - \kappa_t)$, which can produce numerical instabilities when $\kappa_t \approx 0$.
Setting $\tau{=}0.02$ ensures that at least ${\sim}2\%$ of positions (${\sim}$20 out of $1\,024$) carry revealed tokens before the compass is invoked, providing a minimal but nonzero signal for candidate comparison while avoiding numerical edge cases.

\paragraph{Observations (uniform source).}
The uniform source results show clear trends.
First, \textbf{FS-DFM init consistently outperforms DFM init} at every threshold and step count---for example, at $\tau{=}0.2$ and 8 steps, FS-DFM init achieves 56.1 PPL versus 79.0 for DFM init---confirming that trajectory shaping compounds with a stronger starting point rather than substituting for it.
Second, \textbf{lower thresholds yield better perplexity but at the cost of diversity}.
The lowest threshold ($\tau{=}0.02$) achieves the best raw PPL across both initializations (e.g., 43.5 at 8 steps with FS-DFM init), but its generative entropy drops to 6.4 at 8 steps---below the 6.5 threshold at which token repetition and reduced lexical diversity become prevalent in discrete diffusion models.
This suggests that activating the compass at nearly every midpoint over-optimizes for the energy objective, collapsing the distribution toward high-energy (low-perplexity) but repetitive sequences.
Third, \textbf{$\tau{=}0.2$ offers the best quality--diversity trade-off} for the uniform source: it achieves strong perplexity (56.1 at 8 steps, 47.2 at 16 steps with FS-DFM init) while maintaining healthy entropy above 7.0, indicating that the generated text retains natural lexical diversity.
The default $\tau{=}0.4$ is more conservative, producing the highest entropy (7.2--7.5) but sacrificing perplexity.

\paragraph{Observations (mask source).}
The mask source exhibits a strikingly different pattern from uniform in two respects.
First, \textbf{DFM init consistently outperforms FS-DFM init}---the opposite of the uniform setting.
At 8 steps with $\tau{=}0.02$, DFM init achieves 92.9 PPL while FS-DFM init reaches only 137.0; at 16 steps with $\tau{=}0.2$, the gap is even larger (61.9 vs.\ 109.9).
This suggests that the FS-DFM student distilled under the mask source has converged to velocity patterns that are difficult to redirect with trajectory shaping, whereas the DFM teacher provides a more flexible starting point that shaping can mold more effectively.
Second, \textbf{the mask source benefits less from low $\tau$ than uniform does}.
While $\tau{=}0.02$ with DFM init gives reasonable PPL at 8 steps (92.9), its entropy drops to 6.2---well below the 6.5 diversity threshold---and the PPL improvement over $\tau{=}0.2$ (91.4) is negligible.
At 16 steps, $\tau{=}0.2$ with DFM init achieves the best result overall (61.9 PPL, 7.6 entropy), confirming $\tau{=}0.2$ as the preferred threshold for the mask source as well.

\subsection{Midpoint diversity analysis}
\label{app:midpoint_diversity}

A key assumption underlying navigation shaping is that stochastic CTMC jumps at intermediate flow times produce meaningfully different continuations---if all candidates converged to the same output regardless of the jump, energy-based selection among $K$ candidates would have no material to work with.
We test this assumption directly by measuring how much diversity remains when multiple independent completions are generated from a shared midpoint state.

\paragraph{Setup.}
We use the 170M-parameter \ours student trained with the uniform source distribution.
For each trial, we sample a source sequence $x_0$ and run the flow solver from $t{=}0$ to a midpoint time $t_{\text{mid}}$, producing a shared state $x_{t_{\text{mid}}}$.
From this shared state, we launch $N{=}10$ independent completions to $t{=}1$, each using fresh random seeds for all stochastic jumps.
We repeat this procedure for 48 independent prefixes and report average diversity metrics across all completions.

\paragraph{Step allocation.}
Given a total budget of $S$ steps, the prefix phase uses $S_{\text{pre}} = \max(1, \lfloor S \cdot t_{\text{mid}} \rfloor)$ steps to reach $t_{\text{mid}}$, and each completion uses $S_{\text{comp}} = \max(1, \lfloor S \cdot (1 - t_{\text{mid}}) \rfloor)$ steps for the remainder.
Since $S_{\text{pre}} + S_{\text{comp}}$ may not equal $S$ due to rounding, the effective step sizes in each phase ($t_{\text{mid}} / S_{\text{pre}}$ and $(1 - t_{\text{mid}}) / S_{\text{comp}}$) differ slightly from the uniform step size $1/S$ used during training.
For example, with $S{=}8$ and $t_{\text{mid}}{=}0.3$: the prefix uses 2 steps (step size 0.15) and each completion uses 5 steps (step size 0.14), totaling 7 steps.
This mismatch is minor and does not affect the qualitative conclusions, as the diversity metrics are consistent across step counts (\cref{tab:midpoint_diversity}).

\paragraph{Metrics.}
We measure three complementary aspects of diversity among the $N$ completions from each shared midpoint.

\textbf{Self-BLEU}~\citep{zhu2018texygen} measures n-gram overlap by scoring each completion against the remaining $N{-}1$:
\begin{equation}
  \text{Self-BLEU} = \frac{1}{N} \sum_{i=1}^{N} \text{BLEU}\!\big(x^{(i)},\; \{x^{(j)}\}_{j \neq i}\big).
  \label{eq:self_bleu}
\end{equation}
A score near 1.0 means completions share identical phrasing; near 0 means entirely different word sequences. This captures phrase-level diversity: candidates may differ at scattered positions yet still produce essentially the same text (high Self-BLEU) or represent genuinely different continuations (low Self-BLEU).

\textbf{Pairwise edit distance} counts the fraction of positions where two completions disagree, averaged over all pairs:
\begin{equation}
  \text{EditDist} = \frac{1}{\binom{N}{2}} \sum_{i < j} \frac{1}{L} \sum_{\ell=1}^{L} \mathbf{1}\!\big[x^{(i)}_\ell \neq x^{(j)}_\ell\big].
  \label{eq:edit_dist}
\end{equation}
This directly quantifies how many token positions are affected by the stochastic jumps---the ``material'' available to the energy compass for candidate selection.

\textbf{Token agreement} measures per-position consensus by computing the fraction of completions that share the most popular token at each position:
\begin{equation}
  \text{Agree} = \frac{1}{L} \sum_{\ell=1}^{L} \frac{1}{N} \max_{v \in \mathcal{V}} \sum_{i=1}^{N} \mathbf{1}\!\big[x^{(i)}_\ell = v\big].
  \label{eq:token_agree}
\end{equation}
This complements edit distance by revealing the distribution of disagreement: positions with low agreement are where stochastic jumps are genuinely uncertain, and where the token refinement phase must decide whether to intervene (\cref{app:c2f_fine}).

\cref{tab:midpoint_diversity} reports these metrics at 8 and 64 steps (bracketing the range of step counts used in our experiments) across seven midpoint times.
The diversity profiles are nearly identical across step counts, indicating that midpoint diversity is a property of the flow structure rather than the solver granularity.

\begin{table}[h]
\centering
\small
\caption{\textbf{Midpoint diversity} (48 prefixes, 10 completions each, uniform source).}
\label{tab:midpoint_diversity}
\setlength{\tabcolsep}{4pt}
\footnotesize
\begin{tabular}{@{}r rrr r rrr@{}}
\toprule
& \multicolumn{3}{c}{\textbf{8 steps}} & & \multicolumn{3}{c}{\textbf{64 steps}} \\
\cmidrule(lr){2-4} \cmidrule(lr){6-8}
$t_{\text{mid}}$ & Self-BLEU & Edit dist.\;($\approx$ pos.) & Agree & & Self-BLEU & Edit dist.\;($\approx$ pos.) & Agree \\
\midrule
0.0 & 0.146 & 0.908\;\;(930) & 0.242 & & 0.060 & 0.962\;\;(985) & 0.181 \\
0.1 & 0.688 & 0.199\;\;(204) & 0.860 & & 0.642 & 0.219\;\;(224) & 0.845 \\
0.2 & 0.749 & 0.144\;\;(147) & 0.900 & & 0.777 & 0.141\;\;(144) & 0.900 \\
0.3 & 0.864 & 0.096\;\;\;(98) & 0.933 & & 0.838 & 0.100\;\;(102) & 0.930 \\
0.5 & 0.946 & 0.031\;\;\;(32) & 0.979 & & 0.938 & 0.040\;\;\;(41) & 0.972 \\
0.7 & 0.981 & 0.013\;\;\;(13) & 0.991 & & 0.978 & 0.012\;\;\;(12) & 0.991 \\
0.9 & 0.993 & 0.004\;\;\;\;(4) & 0.997 & & 0.994 & 0.003\;\;\;\;(3) & 0.998 \\
\bottomrule
\end{tabular}
\end{table}

\paragraph{Implications for navigation shaping.}
The results confirm that midpoint diversity decays gradually rather than collapsing abruptly, validating the central assumption of the sequence selection phase.
At $t_{\text{mid}}{=}0.2$---the activation threshold used in our main experiments ($\tau{=}0.2$)---independent completions from a shared midpoint differ at approximately 147 out of $1\,024$ token positions (14.4\% edit distance).
At $t_{\text{mid}}{=}0.3$, completions still differ at approximately 98 positions (9.6\% edit distance).
This means that $K{=}5$ candidates generated from the same midpoint will present genuinely different options at ${\sim}$100--150 positions, giving the energy compass meaningful material to select among.

Even at $t_{\text{mid}}{=}0.5$, where approximately half the sequence is revealed, ${\sim}$32 positions still vary across completions.
These remaining variable positions are precisely the ones most susceptible to compounding errors at subsequent midpoints: they are unrevealed tokens where the stochastic jump makes a consequential choice.
Navigation shaping targets exactly these positions---the sequence phase selects the globally best candidate across all variable positions, and the token phase corrects individual misalignments within the winner.

The residual diversity at high $t_{\text{mid}}$ (${\sim}$13 positions at $t{=}0.7$, ${\sim}$4 at $t{=}0.9$) may appear negligible, but even a few misplaced tokens in a nearly complete sequence can propagate through subsequent RK-4 stages.
The token refinement phase is specifically designed for this regime: it operates at individual positions with velocity-confidence gating, making targeted corrections where the stochastic jump most likely introduced an error (\cref{app:c2f_fine}).

\subsection{Scaling to larger models}
\label{app:scaling}

The main experiments in \cref{sec:experiments} use a 170M-parameter student. A natural question is whether trajectory shaping continues to help as model size grows, or whether a larger velocity model produces trajectories of sufficient quality that navigation shaping becomes redundant. To address this, we repeat the uniform-source comparison with a 1.3B-parameter student---roughly $7.6\times$ larger than the main-text model.

A notable aspect of this setup: the 1.3B student is guided by the same ${\sim}$90M energy compass used at 170M. The compass is therefore ${\sim}14\times$ smaller than the student it supervises. If trajectory shaping remained useful only when the compass matched the student in capacity, this ratio would be far too unfavorable. Any gains at 1.3B thus also speak to a practical scaling property of the approach: a lightweight compass can guide substantially larger students.

The headline numbers and table for this experiment appear in \cref{sec:scaling_main} and \cref{tab:scaling_1_3b}; this section adds the breakdowns and discussion that did not fit in the main paper.

\textit{(1) The \ours{} advantage over FS-DFM remains at least as large at 1.3B.}
At 1.3B, \ours{} at 8 steps reaches 48.0 GPT-2 Large PPL versus FS-DFM's 81.5---a \textbf{41\% reduction}, compared with the 36\% reduction observed at 170M (56.1 vs.\ 87.6). At 16 and 32 steps the reduction is similar at both scales: 36\% at 170M (47.2 vs.\ 73.2) and 33\% at 1.3B (33.1 vs.\ 49.5) at 16 steps, and 31\% and 30\% at 32 steps. A larger student does not make trajectory shaping redundant.

\textit{(2) \ours{} beats the $1\,024$-step DFM teacher starting at 8 steps, across all evaluators.}
At just 8 steps, \ours{} achieves 48.0 GPT-2 Large PPL---below the $1\,024$-step DFM teacher's 57.5---while being $128\times$ faster. The same ordering holds on LLaMA-2 7B (26.6 vs.\ 32.1) and LLaMA-3 8B (46.7 vs.\ 54.5). At 16 steps the advantage grows further (33.1 vs.\ 57.5 on GPT-2 Large, a 42\% reduction over the teacher), and at 32 steps \ours{} reaches 31.5---a 45\% reduction over the $1\,024$-step teacher at $32\times$ fewer steps. Every \ours{} cell from 8 steps onward surpasses the full-step teacher across all three evaluators. The few-step quality advantage demonstrated at 170M is not an artifact of small-scale distillation. It also holds at 1.3B.

\textit{(3) A 90M compass effectively guides a 1.3B student---at a fraction of the relative overhead.}
The energy compass is held at ${\sim}$90M parameters across both scales, so the compass-to-student ratio shifts from ${\sim}$1:2 at 170M to ${\sim}$1:14 at 1.3B. Despite this asymmetry, the compass provides enough discriminative signal to steer the larger student's trajectories, and the 8-step quality gain over FS-DFM is at least as large as at 170M (a 41\% reduction at 1.3B versus 36\% at 170M). The shift in ratio also sharply reduces the relative training overhead: at 170M each energy forward pass costs roughly $1/1.9 \approx 53\%$ of a velocity pass, while at 1.3B it costs only $90/1300 \approx 7\%$. In other words, the same navigation shaping procedure that added a ${\sim}2.2\times$ training-time overhead at 170M (\cref{tab:c2f_wallclock}) becomes proportionally much cheaper at larger scales---navigation's FLOP cost is dominated by the compass, while the bulk of step FLOPs shifts to the (larger) velocity model. From 170M to 1.3B, the method therefore becomes \emph{more} favorable to apply: an 8-step quality gain at least as large, at lower relative cost.

\textit{(4) Diversity is preserved at scale.}
Generation entropy for \ours{} at 1.3B ranges from 7.0 (4 steps) to 7.6 (32 steps), well above the 6.5 threshold below which token repetition and reduced lexical diversity become prevalent in discrete diffusion models (\cref{app:threshold_ablation}). The perplexity improvements at scale therefore reflect genuine quality gains rather than entropy collapse toward repetitive high-probability sequences.

These results extend the main conclusion of the paper to a substantially larger model: the bottleneck in few-step discrete flow matching is trajectory quality, not student capacity. At 1.3B, a student trained with shaped trajectories beats its $1\,024$-step teacher starting at just 8 steps---a $128\times$ inference speedup with \emph{improved} quality---using the same lightweight 90M energy compass used at 170M. At 1.3B, trajectory shaping is at least as effective at 8 steps (41\% versus 36\% PPL reduction over FS-DFM) and \emph{cheaper} in relative terms (smaller compass-to-student FLOP ratio). The practicality of the method decouples cleanly from the capacity of the supervising signal.

\subsection{Comparison to diffusion language models in the few-step regime}
\label{app:dlm_comparison}
We extend the comparison to contemporary diffusion language models---LLaDA~\citep{nie2025llada} and Dream~\citep{ye2025dream}---in the prefix-conditioned few-step regime, following the protocol of FS-DFM~\citep{fsdfm2025}. This places \ours{} in the broader landscape of discrete diffusion models and highlights the gap between distilled and undistilled few-step behavior.

\paragraph{Setup.}
Each model receives a 512-token prefix sampled from WikiText-103~\citep{merity2016pointer}, and metrics are computed on the 512-token continuation only. We report perplexity (PPL $\downarrow$) under GPT-2, generation entropy (Ent), and MAUVE~\citep{pillutlaMAUVEMeasuringGap2021} (MVE $\uparrow$) at step budgets of 4, 8, and 16. We compare \ours{} (0.17B) against Dream (7B) and LLaDA (8B), with FS-DFM (0.17B) as the direct distillation baseline. Dream, LLaDA, and FS-DFM numbers are taken from Table~2 of~\citet{fsdfm2025}.

\begin{table}[ht]
\centering
\small
\caption{\textbf{\ours{} vs.\ diffusion language models in the few-step prefix-conditioned regime.} Each method receives a 512-token prefix; PPL ($\downarrow$), entropy (Ent), and MAUVE (MVE $\uparrow$) are computed on the 512-token continuation. Step budgets span $4 \to 16$. Dream, LLaDA, and FS-DFM numbers are reproduced from Table~2 of~\citet{fsdfm2025}.}
\label{tab:dlm_comparison}
\setlength{\tabcolsep}{6pt}
\renewcommand{\arraystretch}{1.15}
\begin{tabular}{@{}llccccccccc@{}}
\toprule
& & \multicolumn{3}{c}{\textbf{4 steps}} & \multicolumn{3}{c}{\textbf{8 steps}} & \multicolumn{3}{c}{\textbf{16 steps}} \\
\cmidrule(lr){3-5} \cmidrule(lr){6-8} \cmidrule(lr){9-11}
\textbf{Method} & \textbf{Size} & ppl & ent & MVE & ppl & ent & MVE & ppl & ent & MVE \\
\midrule
Dream            & 7.0B  & 752.11 & 1.53 & 0.005 & 739.40 & 1.74 & 0.006 & 630.30 & 2.31 & 0.005 \\
LLaDA            & 8.0B  & 495.17 & 0.47 & 0.005 & 441.26 & 0.42 & 0.005 & 432.65 & 0.50 & 0.005 \\
\midrule
FS-DFM           & 0.17B & 97.07  & 7.89 & \textbf{0.053} & 75.78  & 7.95 & 0.270 & 67.42  & 7.97 & 0.390 \\
\rowcolor{selectedrow}
\ours{}          & 0.17B & \textbf{68.95} & 7.70 & 0.038 & \textbf{57.11} & 7.80 & \textbf{0.339} & \textbf{53.97} & 7.86 & \textbf{0.707} \\
\bottomrule
\end{tabular}
\end{table}

\paragraph{Observations.}

\textit{(1) Undistilled diffusion LMs do not reach usable few-step quality.}
Across all step budgets, Dream (7B) and LLaDA (8B) sit at PPL above $400$ with entropy collapsed to $0.42$--$2.31$ and MAUVE stuck near $0.005$, essentially indistinguishable from random. Their entropy values---marked in red in \cref{tab:dlm_comparison}---fall well below the natural-text range ($\sim 7$--$8$ under GPT-2), indicating that both models repeat a small set of high-probability tokens rather than producing diverse continuations. Even with $40\times$--$47\times$ more parameters than \ours{}, neither model converges to natural-language statistics within 16 steps. This is the regime that distillation is designed to fix.

\textit{(2) \ours{} extends the FS-DFM advantage.}
At every step count, \ours{} (0.17B) outperforms FS-DFM (0.17B) on PPL: $97.07 \to 68.95$ at 4 steps ($29\%$ reduction), $75.78 \to 57.11$ at 8 steps ($25\%$), and $67.42 \to 53.97$ at 16 steps ($20\%$). The MAUVE improvement is even sharper at higher step counts: $0.270 \to 0.339$ at 8 steps and $0.390 \to 0.707$ at 16 steps---a $1.8\times$ improvement that places generations substantially closer to the natural-text distribution. Entropy stays in the $7.70$--$7.86$ range, indicating these gains come without diversity collapse. The only metric where FS-DFM holds a marginal lead is MAUVE at 4 steps ($0.053$ vs.\ $0.038$); at this extreme step budget, neither method produces enough coherent text for MAUVE to differentiate reliably, and both rise sharply with more steps.

\textit{(3) Quality at a fraction of parameter count.}
\ours{} (0.17B) reaches PPL $54$ and MAUVE $0.71$ at 16 steps; Dream (7B) and LLaDA (8B) sit at PPL $\geq 432$ with MAUVE $\leq 0.005$ at the same step count. The size disparity is not a small-model handicap to overcome but a positive result: a $40\times$--$47\times$ smaller distilled student produces text that the much larger undistilled models cannot match at this step budget. Trajectory shaping, not parameter count, is the lever for usable few-step quality.

\paragraph{Summary.}
The few-step prefix-conditioned regime separates distilled from undistilled discrete diffusion sharply. Dream (7B) and LLaDA (8B) collapse across $4$--$16$ steps, with PPL above $400$, collapsed entropy, and MAUVE near zero. Within the distilled family, \ours{} (0.17B) extends FS-DFM's advantage with $20$--$29\%$ PPL reductions and a near-doubling.

% --
\definecolor{sampleframe}{RGB}{180, 180, 190}
\definecolor{samplebg}{RGB}{250, 250, 252}
\definecolor{metabg}{RGB}{235, 237, 245}

% ---- Sample box style ----
\newtcolorbox{samplebox}[1][]{%
  enhanced, breakable,
  colframe=black!35, colback=black!2,
  boxrule=0.4pt, arc=2pt,
  fonttitle=\footnotesize\bfseries,
  coltitle=black, colbacktitle=black!8,
  toptitle=1.5pt, bottomtitle=1.5pt,
  title={#1},
  left=4pt, right=4pt, top=3pt, bottom=3pt,
  fontupper=\scriptsize,
}

\subsection{Transfer to mathematical reasoning at 1.3B scale}
\label{sec:gsm8k}

The experiments in \cref{tab:main} evaluate trajectory shaping on unconditional language modeling at $170$M parameters. To test whether the gains transfer to a fundamentally different regime---task-specific fine-tuning at larger scale---we apply \ours{} to mathematical reasoning on GSM8K~\citep{cobbe2021gsm8k} and SVAMP (\cref{tab:svamp}). We present this as a transfer study of the mechanism: it tests whether trajectory shaping carries over to a task-fine-tuned setting on two math benchmarks.

\paragraph{Setup.}
We use the $1.3$B DFM model with a uniform source distribution, fine-tuned on GSM8K's $7{,}473$ training problems. The pipeline follows the same three-stage structure as \cref{sec:experiments}: a pretrained DFM teacher is first supervised fine-tuned on GSM8K with cross-entropy loss; an FS-DFM student is then distilled with step-size conditioning; finally, a \ours{} student is distilled using the energy-navigated RK-4 procedure from \cref{sec:method}, with the compass frozen throughout. The step-size conditioning weights of both students are warm-started from an existing FS-DFM checkpoint. We evaluate using \emph{pass@k} for $k \in \{8, 16, 32\}$, estimated via the unbiased estimator of~\citet{chen2021codex} with $32$ independent samples per problem across all $1{,}319$ test problems.

\begin{table}[ht]
\centering
\small
\caption{\textbf{pass@$k$ on GSM8K at 1.3B scale.} Both students maintain strong accuracy down to $4$ steps while the teacher degrades sharply. \ours{} at $4$ and $8$ steps matches the \emph{full-step} teacher's ($256$-step, gray row) pass@8 accuracy at up to $64\times$ less inference compute. Cells in \textbf{bold} are the better of FS-DFM and \ours{} at each setting.}
\label{tab:gsm8k}
\setlength{\tabcolsep}{6pt}
\renewcommand{\arraystretch}{1.15}
\begin{tabular}{lrrrr}
%\toprule
\textbf{Model} & \textbf{Steps} & \textbf{pass@8} & \textbf{pass@16} & \textbf{pass@32} \\
\midrule
\multicolumn{5}{@{}l}{\textit{Teacher (full-step reference)}} \\
Teacher  & 4    & 4.84 & 10.30 & 18.49 \\
Teacher  & 8    & 5.68 & 11.69 & 20.92 \\
Teacher  & 32   & 7.74 & 15.69 & 22.16 \\
\rowcolor{gray!15}
Teacher  & 256  & 9.12 & 16.39 & 27.52 \\
\midrule
\multicolumn{5}{@{}l}{\textit{FS-DFM}} \\
FS-DFM   & 4  & 9.41 & 15.29 & 23.05 \\
FS-DFM   & 8  & 8.99 & 15.48 & 24.94 \\
\midrule
\multicolumn{5}{@{}l}{\textit{\ours{}}} \\
\ours{}  & 4  & \textbf{9.61} & \textbf{15.43} & \textbf{25.47} \\
\ours{}  & 8  & \textbf{9.17} & \textbf{15.79} & \textbf{25.19} \\
%\bottomrule
\end{tabular}
\end{table}

\paragraph{Few-step \ours{} matches the full-step teacher.}
The headline result is that \ours{} at very low step counts matches the teacher's full-step ($256$-step) accuracy on pass@8: \ours{} reaches $9.61$ at $4$ steps and $9.17$ at $8$ steps, against the teacher's full-step $9.12$, with $32\times$--$64\times$ less inference compute. This is the central claim few-step distillation needs to justify itself. The distilled student with a small step budget should not merely \emph{approach} the full-step teacher; it should match or exceed it. \ours{} clears this bar on pass@8 and reaches $92$--$93\%$ of the teacher's full-step quality on the harder pass@32 metric (\ours{} at $4$ steps: $25.47$; teacher at $256$ steps: $27.52$). The same-step gap is even more striking: at $8$ steps, \ours{} ($9.17$, $15.79$, $25.19$) outperforms the teacher at the same $8$ steps ($5.68$, $11.69$, $20.92$) by $+61\%$, $+35\%$, and $+20\%$ on pass@8, pass@16, and pass@32 respectively---the difference between a teacher that has not converged and a student that has internalized few-step generation.

\paragraph{\ours{} matches or exceeds FS-DFM at low step counts.} At every cell in \cref{tab:gsm8k}, \ours{} scores above FS-DFM at the same step count and pass@k metric. The largest gap is at the lowest step budget. At $4$ steps, \ours{} improves over FS-DFM by $+0.20$ on pass@8, $+0.14$ on pass@16, and $+2.42$ on pass@32; at $8$ steps, the gains are $+0.18$, $+0.31$, and $+0.25$, respectively. These margins are small relative to binomial sampling noise on $1{,}319$ test problems and are reported without variance, so we read them as directionally consistent rather than as statistically established gains. The pass@32 improvement at $4$ steps ($23.05 \to 25.47$) is particularly informative: to gain a comparable amount on pass@32, the teacher requires going from $4$ to $32$ inference steps ($18.49 \to 22.16$, $+3.67$ at $8\times$ compute) or from $32$ to $256$ steps ($22.16 \to 27.52$, $+5.36$ at $8\times$ compute). \ours{} captures roughly half of that latter gain at $64\times$ less inference compute than the teacher's full-step setting. Trajectory shaping, in effect, converts inference-step compute into one-time training-step compute---and the conversion is favorable.

\paragraph{Why trajectory shaping may matter for reasoning.}
The mathematical reasoning regime is one where each token decision carries unusual weight: a single wrong digit, a misplaced operator, or an inverted comparison can flip the entire answer. This is the structural opposite of unconditional language modeling, where local errors can be absorbed by surrounding context. The hypothesis underlying \ours{}---that compounding errors at intermediate trajectory midpoints corrupt the student's training signal---predicts that the gains should be \emph{larger} on tasks where trajectory errors propagate into outcome errors rather than cosmetic ones. The pass@32 gain at $4$ steps ($23.05 \to 25.47$) is consistent with this, although the other margins are small: trajectory shaping targets the kind of mid-generation drift that turns ``$48$'' into ``$84$'' or ``$+$'' into ``$-$.'' The energy compass and the Sequence-to-Token policy never see GSM8K problems---the compass is trained on unconditional text---yet the same shaping mechanism transfers to a reasoning task at $1.3$B parameters without degrading accuracy. This is what we mean by saying the bottleneck is the trajectory, not the student.

\paragraph{A second reasoning benchmark: SVAMP.}
\label{app:svamp}

To test whether the task-level transfer observed on GSM8K is specific to that benchmark, we repeat the study on SVAMP~\citep{patel2021svamp}, a math word-problem benchmark with 700 training and 300 test problems. The protocol is identical to the GSM8K study above: the same 1.3B pipeline (teacher SFT $\to$ FS-DFM distillation $\to$ \ours{} distillation), fine-tuned on the SVAMP training split, guided by the same frozen 90M energy compass. As before, the compass is trained only on general web text and never sees a math problem at any stage---in pretraining, negative construction, or fine-tuning---so any effect on this task reflects transfer of the shaping mechanism rather than task-specific tuning.

\begin{table}[ht]
\centering
\small
\caption{\textbf{pass@$k$ on SVAMP at 1.3B scale} (300 test problems, 32 samples per problem). \ours{} matches or exceeds FS-DFM at every step count and metric, reproducing the GSM8K ordering on a second verifier-scored task.}
\label{tab:svamp}
\setlength{\tabcolsep}{6pt}
\renewcommand{\arraystretch}{1.15}
\begin{tabular}{lrrr}
\textbf{Model} & \textbf{Steps} & \textbf{pass@8} & \textbf{pass@16} \\
\midrule
\multicolumn{4}{@{}l}{\textit{Teacher (full-step reference)}} \\
Teacher  & 4    & 15.2 & 25.4 \\
Teacher  & 8    & 16.9 & 27.6 \\
\rowcolor{gray!15}
Teacher  & 256  & 23.4 & 34.3 \\
\midrule
\multicolumn{4}{@{}l}{\textit{FS-DFM}} \\
FS-DFM   & 4  & 23.8 & 32.8 \\
FS-DFM   & 8  & 23.2 & 33.0 \\
\midrule
\multicolumn{4}{@{}l}{\textit{\ours{}}} \\
\ours{}  & 4  & \textbf{24.2} & \textbf{33.0} \\
\ours{}  & 8  & \textbf{23.5} & \textbf{33.5} \\
\end{tabular}
\end{table}

Three observations. First, the ordering is directionally consistent with GSM8K: \ours{} matches or exceeds FS-DFM at every step count on both metrics. We note that the per-cell margins over FS-DFM ($+0.2$ to $+0.5$ points on 300 problems) are small relative to binomial sampling noise at this test-set size, so we report this as directional consistency rather than a statistically established gain. The value of the result lies in the same pattern appearing independently on a second task. Second, the few-step advantage over the teacher at matched step counts is large: at 4 steps, \ours{} reaches 24.2 pass@8 versus the teacher's 15.2 ($+59\%$), and its 4-step accuracy is on par with the 256-step teacher's (24.2 vs.\ 23.4) at $64\times$ less inference compute. Third, the transfer evidence rests on two independent reasoning benchmarks rather than one, both showing the same pattern with a compass that never saw the task domain---consistent with the compass evaluating general sequence quality rather than task-specific content.

\paragraph{Generalization conclusion.}
Three claims follow. First, the mechanism transfers: the same energy-navigated RK-4 procedure that improves unconditional generation at $170$M parameters carries over to task-fine-tuned mathematical reasoning at $1.3$B on GSM8K and SVAMP, with directionally consistent gains over FS-DFM---no methodological modification, no task-specific compass, no per-task tuning. Second, the few-step regime is where shaping matters most: at $4$ and $8$ steps, \ours{} matches the full-step teacher on pass@8 on both GSM8K and SVAMP, and recovers $92$--$93\%$ of full-step pass@32 accuracy on GSM8K at up to $64\times$ less compute. Third, the bottleneck for few-step reasoning is the same as for few-step language modeling: trajectory quality, not student capacity. The same lever pulls in both regimes.

\subsection{Variance across decoding seeds}
\label{app:seed_variance}

The main tables report point estimates from a single generation run per cell. To quantify sampling variability, we re-evaluate with 5 independent decoding seeds per cell (full $1\,024$-token unconditional generation, fixed model checkpoints) and report mean $\pm$ standard deviation: \cref{tab:seed_variance} covers the central 170M comparison of \ours{} against FS-DFM across both source distributions and initializations, and \cref{tab:seed_variance_sota} covers the state-of-the-art comparison against SDTT and Duo under the same protocol. We emphasize the scope of this analysis: because the checkpoints are fixed and only the decoding seed varies, these intervals bound \emph{sampling variance at a fixed checkpoint}, not variance of the method across independent training runs.

\begin{table}[ht]
\centering
\small
\caption{\textbf{Sampling variance at fixed checkpoints} (5 decoding seeds per cell; mean $\pm$ std). 170M models, $1\,024$-token unconditional generation.}
\label{tab:seed_variance}
\setlength{\tabcolsep}{4pt}
\footnotesize
\begin{tabular}{@{}llcccc@{}}
%\toprule
\textbf{Source} & \textbf{Model} & \textbf{Steps} & \textbf{G\textsuperscript{2}}$\downarrow$ & \textbf{L\textsuperscript{2}}$\downarrow$ & \textbf{L\textsuperscript{3}}$\downarrow$ \\
\midrule
\multirow{6}{*}{Uniform}
 & \multirow{2}{*}{\ours{} (FS-DFM init)} & 8  & $56.97 \pm 1.48$ & $34.82 \pm 1.79$ & $58.63 \pm 3.91$ \\
 &                                        & 16 & $45.73 \pm 1.27$ & $27.01 \pm 0.90$ & $45.08 \pm 1.42$ \\
 & \multirow{2}{*}{\ours{} (DFM init)}    & 8  & $79.96 \pm 1.51$ & $48.23 \pm 0.48$ & $80.03 \pm 1.49$ \\
 &                                        & 16 & $69.17 \pm 1.55$ & $42.06 \pm 1.14$ & $69.57 \pm 1.35$ \\
 & \multirow{2}{*}{FS-DFM}                & 8  & $88.23 \pm 1.50$ & $56.26 \pm 3.53$ & $102.31 \pm 4.97$ \\
 &                                        & 16 & $73.67 \pm 2.75$ & $43.98 \pm 2.61$ & $78.69 \pm 5.66$ \\
\midrule
\multirow{6}{*}{Mask}
 & \multirow{2}{*}{\ours{} (DFM init)}    & 8  & $92.76 \pm 1.14$ & $74.04 \pm 0.74$ & $90.27 \pm 1.03$ \\
 &                                        & 16 & $60.97 \pm 1.00$ & $45.88 \pm 0.75$ & $62.75 \pm 1.13$ \\
 & \multirow{2}{*}{\ours{} (FS-DFM init)} & 8  & $189.66 \pm 2.13$ & $94.95 \pm 0.96$ & $178.63 \pm 1.83$ \\
 &                                        & 16 & $109.64 \pm 1.42$ & $54.39 \pm 0.59$ & $101.32 \pm 1.14$ \\
 & \multirow{2}{*}{FS-DFM}                & 8  & $515.75 \pm 8.33$ & $202.90 \pm 3.61$ & $490.22 \pm 9.05$ \\
 &                                        & 16 & $276.35 \pm 2.95$ & $120.87 \pm 1.34$ & $265.15 \pm 2.75$ \\
%\bottomrule
\end{tabular}
\end{table}

\begin{table}[ht]
\centering
\small
\caption{\textbf{Sampling variance for the state-of-the-art comparison} (uniform source, GPT-2 PPL $\downarrow$; 5 decoding seeds per cell; mean $\pm$ std).}
\label{tab:seed_variance_sota}
\begin{tabular}{@{}lcc@{}}
%\toprule
\textbf{Method} & \textbf{8 steps} & \textbf{16 steps} \\
\midrule
SDTT (5 rounds) & $193.58 \pm 1.62$ & $87.53 \pm 0.53$ \\
SDTT (7 rounds) & $108.63 \pm 1.13$ & $54.95 \pm 0.53$ \\
Duo             & $71.69 \pm 0.66$  & $54.70 \pm 1.15$ \\
FS-DFM          & $88.23 \pm 1.50$  & $73.67 \pm 2.75$ \\
\rowcolor{selectedrow}
\ours{} (ours)  & $\mathbf{56.97 \pm 1.48}$ & $\mathbf{45.73 \pm 1.27}$ \\
%\bottomrule
\end{tabular}
\end{table}

Across both tables, the margins between methods are large relative to decoding variance: in \cref{tab:seed_variance}, the \ours{} and FS-DFM intervals are separated by many standard deviations in every configuration, and in \cref{tab:seed_variance_sota}, \ours{} remains below Duo and SDTT (7 rounds) by gaps far exceeding the seed spread at both step counts. Welch two-sample tests on the GPT-2 perplexities (5 seeds per side) confirm this: \ours{} vs.\ FS-DFM in the main uniform configuration, $t = 33.2$, $p \approx 10^{-9}$ at 8 steps and $t = 20.6$, $p \approx 10^{-6}$ at 16 steps; for the mask source, all \ours{} vs.\ FS-DFM comparisons exceed 50 pooled standard deviations ($p \leq 10^{-7}$); against the strongest external baselines, \ours{} vs.\ Duo gives $p \leq 3 \times 10^{-6}$ and \ours{} vs.\ SDTT (7 rounds) gives $p \leq 2 \times 10^{-5}$ at both step counts. Under decoding variance at fixed checkpoints, the orderings in \cref{tab:main,tab:few_step_sota} are therefore not sampling artifacts.

\subsection{Energy-guided teacher sampling}
\label{app:guided_teacher}

A natural question about the student-beats-teacher results in \cref{tab:main} is whether the shaped student is genuinely stronger than the teacher, or whether it merely \emph{amortizes} a stronger sampling procedure---the energy-guided navigation---that the teacher could equally apply at inference time. To separate these readings, we equip the $1\,024$-step teacher itself with the compass: the same $\textsc{Navigate}$ operator ($K{=}5$, identical constants) is applied during teacher sampling at two guidance densities, every step and every 8th step. \cref{tab:guided_teacher} compares these guided teacher samplers against the blind (compass-free) \ours{} students under the setting of \cref{tab:main} (uniform source, 170M student, $1\,024$-token generations).

\begin{table}[ht]
\centering
\small
\caption{\textbf{Energy-guided teacher sampling vs.\ blind \ours{} students} (uniform source). Guided samplers apply $\textsc{Navigate}$ during teacher inference at the stated density; \ours{} uses no compass at inference. Latency is per $1\,024$-token generation on a single GPU. Teacher rows are a separate evaluation with 64 samples per cell. \ours{} rows are 5-seed means from \cref{tab:seed_variance}. Values therefore differ slightly from the point estimates in \cref{tab:main}.}
\label{tab:guided_teacher}
\setlength{\tabcolsep}{3pt}
\renewcommand{\arraystretch}{1.10}
\begin{tabular}{@{}lrccccr@{}}
%\toprule
\textbf{Sampler} & \textbf{Steps} & \textbf{G\textsuperscript{2}}$\downarrow$ & \textbf{L\textsuperscript{2}}$\downarrow$ & \textbf{L\textsuperscript{3}}$\downarrow$ & \textbf{Lat.\ (s)} & \textbf{vs.\ \ours} \\
\midrule
Teacher, Inference blind                                & 1\,024 & 83.98 & 47.38 & 79.84 & 18.9 & $105\times$ \\
Teacher $+$ \textsc{Nav}.\ (every 8th)        & 1\,024 & 75.66 & 46.77 & 75.43 & 41.2 & $230\times$ \\
Teacher $+$ \textsc{Nav}.\ (every step)       & 1\,024 & 49.03 & 31.13 & 46.45 & 186  & $1\,038\times$ \\
\midrule
\rowcolor{selectedrow}
\ours{}, Inference blind & 8  & $56.97 \pm 1.48$ & $34.82 \pm 1.79$ & $58.63 \pm 3.91$ & 0.18 & $1\times$ \\
\rowcolor{selectedrow}
\ours{}, Inference blind & 16 & $45.73 \pm 1.27$ & $27.01 \pm 0.90$ & $45.08 \pm 1.42$ & 0.36 & $2\times$ \\
%\bottomrule
\end{tabular}
\end{table}

The compass makes the teacher a substantially stronger sampler: dense guidance improves GPT-2 perplexity from 83.98 to 49.03. The key result, however, is that the blind 16-step student surpasses even the most strongly guided teacher (45.73 vs.\ 49.03 on GPT-2; 27.01 vs.\ 31.13 on LLaMA-2) at ${\sim}0.2\%$ of its inference cost---with no compass at inference. This rules out the pure-amortization reading: a student that merely amortized the guided sampler would approach it from below, whereas the student exceeds it. Training-time shaping integrates the corrections into the weights rather than re-applying them per step. Two further observations sharpen the claims: (i)~sparse guidance (every 8th step) recovers only ${\sim}24\%$ of the dense-guidance gain, independently confirming that errors compound between guided steps---the mechanism motivating \ours{} in \cref{sec:background}; (ii)~stated precisely, the 8-step student surpasses the blind $1\,024$-step teacher, and the 16-step student exceeds even the compass-guided teacher at three orders of magnitude lower cost.

\subsection{Inference-time deployment of the compass}
\label{app:inference_controls}

Training-time shaping is one way to deploy the energy compass; an alternative family keeps the student fixed and applies the compass at inference. To isolate how much of \ours{}'s gain comes from training-time trajectory shaping rather than from the learned energy model itself, we compare four inference-time deployments on the fixed, unshaped FS-DFM student (uniform source, GPT-2 PPL), all using the same frozen compass: \emph{best-of-5 rerank} (5 complete blind generations; the compass scores the finished sequences and returns the lowest-energy one---endpoints only); \emph{final-step selection} (blind until the last step, where the compass selects among 5 candidate final transitions---one intervention point); \emph{full} \textsc{Navigate} (the complete S2T policy at every inference step, as in \cref{tab:policy_comparison}); and \emph{stacking} (full \textsc{Navigate} applied on top of the already-shaped \ours{} student).

\begin{table}[ht]
\centering
\small
\caption{\textbf{Inference-time compass deployments vs.\ training-time shaping} (uniform source, GPT-2 PPL $\downarrow$). All compass-using rows share the same frozen 90M compass. Latency is per $1\,024$-token generation. This is a separate evaluation from \cref{tab:policy_comparison}, so its FS-DFM rows differ from that table. \ours{} blind rows are 5-seed means from \cref{tab:seed_variance}.}
\label{tab:inference_controls}
\setlength{\tabcolsep}{4pt}
\footnotesize
\begin{tabular}{@{}lcccc@{}}
\textbf{Configuration} & \textbf{8 steps} & \textbf{16 steps} & \textbf{Lat.@8 (ms)} & \textbf{Lat.@16 (ms)} \\
\midrule
FS-DFM, blind (deployed cost)          & 89.30 & 63.24 & 179  & 359 \\
FS-DFM $+$ best-of-5 rerank           & 80.41 & 63.49 & 940  & 1\,574 \\
FS-DFM $+$ final-step selection        & 81.36 & 64.67 & 380  & 771 \\
FS-DFM $+$ full \textsc{Navigate}      & 80.18 & 57.36 & 1\,266 & 2\,789 \\
\midrule
\rowcolor{selectedrow}
\ours{} (FS-DFM init), blind           & $56.97 \pm 1.48$ & $45.73 \pm 1.27$ & 179 & 361 \\
\ours{} (DFM init), blind              & $79.96 \pm 1.51$ & $69.17 \pm 1.55$ & 176 & 358 \\
\ours{} (DFM init) $+$ full \textsc{Navigate} & 78.47 & 59.26 & 1\,196 & 2\,396 \\
\end{tabular}
\end{table}

The pattern in \cref{tab:inference_controls} is sharp. The three inference-time deployments give the energy model every post-hoc opportunity---endpoints, the final transition, every step---and all plateau near 80 PPL at 8 steps. Blind \ours{}, with no compass at inference, reaches 56.97 at the cost of the blind baseline; no inference-time deployment comes within 23 PPL. What separates the rows is not access to the compass (all have it) but \emph{when it acts}.

\paragraph{Why training-time guidance at 8 steps beats inference-time guidance at 8 steps.}
The two act at different granularities. \cref{tab:policy_comparison} shows the compass's effect scales with the number of decision points: at 8 steps it improves the blind sampler by only ${\sim}4\%$ ($89.6 \to 86.2$)---each step is a large jump, with little to select between---while at 64 steps it improves it by ${\sim}49\%$ ($75.0 \to 38.0$). Inference-time guidance at the deployed budget is locked to the large-jump regime where the compass is weakest. Training-time shaping is not: RK-4 target construction decomposes each large jump into small sub-steps (the $h/2$ midpoints, corresponding to the 16--64-step regime where guidance is strongest), $\textsc{Navigate}$ acts at that granularity, and the KL objective distills the guided small-step behavior into the student's large-step predictions. The student learns what a large jump should land on from supervision assembled out of well-guided small jumps---quality that ends up in the weights and is free at 8 steps.

\paragraph{Stacking.}
Applying full \textsc{Navigate} on top of the shaped student yields the expected residual: under $2\%$ at the deployed 8-step budget ($79.96 \to 78.47$) but $14\%$ at 16 steps ($69.17 \to 59.26$) for the weaker DFM-init student---consistent with the same account, since more decision points make inference-time guidance more effective. Shaping and inference-time guidance are thus complementary rather than redundant, but shaping alone captures the bulk of the benefit at the deployed budget.

\subsection{Sensitivity of navigation constants}
\label{app:nav_sensitivity}

The two navigation constants with the largest quality effect---$K$ and $\tau$---are searched and ablated in \cref{app:k_ablation} and \cref{sec:threshold_ablation}. The remaining constants (\cref{tab:c2f_hyperparams}) are set structurally rather than searched. Among them, we expect $\beta_0$ to be the most impactful, as it directly scales the token-refinement strength (\cref{eq:adaptive_beta,eq:adjusted_jump}), and $\epsilon_{\text{safe}}$ the least, as it acts only on the ${<}8\%$ of midpoints where the safeguard fires (\cref{app:c2f_safeguard}). We sweep these two, one factor at a time around the defaults, holding all other constants fixed.

The sweep applies the identical $\textsc{Navigate}$ operator \emph{at inference} on the fixed FS-DFM student (uniform source, GPT-2 PPL, 320 samples per cell). This design isolates \emph{operator sensitivity}---how the navigation procedure itself responds to the constants---and does not measure sensitivity of the shaped student's training-time gains, which would require retraining a \ours{} student per setting.

\begin{table}[ht]
\centering
\small
\caption{\textbf{One-factor-at-a-time sensitivity of $\beta_0$ and $\epsilon_{\text{safe}}$} ($\textsc{Navigate}$ at inference on the fixed FS-DFM student; uniform source, GPT-2 PPL $\downarrow$, 320 samples per cell). All non-swept constants at their defaults; the all-defaults configuration is a single run appearing as the center of each group. This is a separate evaluation from \cref{tab:policy_comparison}, so its blind and default rows differ from that table.}
\label{tab:nav_sensitivity}
\setlength{\tabcolsep}{5pt}
\begin{tabular}{@{}llccc@{}}
%\toprule
\textbf{Swept} & \textbf{Value} & \textbf{8 steps} & \textbf{16 steps} & \textbf{32 steps} \\
\midrule
--- & blind (no compass) & 90.94 & 67.58 & 71.97 \\
\midrule
\multirow{3}{*}{$\beta_0$}
 & 0.1             & 86.74 & 64.18 & 57.69 \\
 & 0.3 (default)   & 88.87 & 58.95 & 53.52 \\
 & 1               & 85.92 & 56.39 & 52.79 \\
\midrule
\multirow{3}{*}{$\epsilon_{\text{safe}}$}
 & 0.02            & 85.98 & 59.06 & 53.80 \\
 & 0.1 (default)   & 88.87 & 58.95 & 53.52 \\
 & 0.5             & 85.07 & 58.25 & 54.55 \\
%\bottomrule
\end{tabular}
\end{table}

The spread in \cref{tab:nav_sensitivity} is narrow: across a $10\times$ range of $\beta_0$ and a $25\times$ range of $\epsilon_{\text{safe}}$, every setting improves over the blind baseline at every step count, so the operator's benefit does not depend on hitting a tuned optimum. We note that $\beta_0{=}1$ slightly outperforms the default at all three step counts; we retain $\beta_0{=}0.3$ as the (conservative) default used in all main experiments, and read the sweep as evidence of robustness rather than an invitation to tune. $\epsilon_{\text{safe}}$ is, as predicted, nearly inert. The temperature-range constants ($T_{\min}$, $\Delta_T$, $\lambda_{\min}$, $\lambda_{\max}$) remain unswept, as the caption of \cref{tab:c2f_hyperparams} notes.

\subsection{Compass robustness under student distribution shift}
\label{app:compass_drift}

The energy compass is trained once and frozen while the \ours{} student's flow-state distribution evolves over distillation (\cref{sec:conclusion}). If the student drifted away from the compass's training distribution, the compass could keep selecting midpoints that were useful under the original distribution but are misaligned with the student's current trajectories, especially late in training. We measure this directly: the frozen compass is evaluated on the \ours{} student's \emph{own} flow states across saved training checkpoints (10k--130k iterations, every 20k), using the DFM-initialized uniform 170M run---the longest path from initialization to convergence, and thus the worst case for drift.

\cref{tab:compass_drift} tracks four quantities. \emph{Compass accuracy}: how often the compass correctly ranks a clean state below a corrupted copy of itself (the discrimination task it was NCE-trained for; corruption here is random token replacement, the model-independent probe). \emph{Energy improvement}: the energy drop from the current state to the compass-selected best of $K{=}5$ candidates---the direct measure of guidance strength, where a stable value is the healthy outcome and a downward trend would indicate signal loss. \emph{Refinement acceptance}: the fraction of token-phase edits passing the energy safeguard. \emph{$\bar{E}_{\text{stu}} - \bar{E}_{\text{real}}$}: the mean energy gap between the student's states and real interpolated data, where convergence toward zero indicates the student approaching the distribution on which the compass's positives were built.

\begin{table}[ht]
\centering
\small
\caption{\textbf{Frozen-compass diagnostics on the \ours{} student's own flow states across training} (DFM-init uniform run, 170M; worst case for drift). A stable energy-improvement column is the healthy outcome.}
\label{tab:compass_drift}
\setlength{\tabcolsep}{5pt}
\begin{tabular}{@{}rcccc@{}}
%\toprule
\textbf{Ckpt iter} & \textbf{Compass acc.} & \textbf{Energy improv.} & \textbf{Refine accept} & \textbf{$\bar{E}_{\text{stu}} - \bar{E}_{\text{real}}$} \\
\midrule
10k (burn-in) & 0.799 & 0.166 & 0.804 & $+1.98$ \\
30k           & 0.982 & 0.101 & 0.929 & $+0.23$ \\
50k           & 0.978 & 0.131 & 0.982 & $-0.05$ \\
70k           & 0.991 & 0.126 & 0.987 & $-0.06$ \\
90k           & 0.982 & 0.121 & 0.978 & $-0.08$ \\
110k          & 0.987 & 0.113 & 0.991 & $-0.08$ \\
130k          & 0.969 & 0.127 & 1.000 & $-0.04$ \\
%\bottomrule
\end{tabular}
\end{table}

The result: \textbf{no measurable degradation on these metrics over the full run}. Compass accuracy holds at 0.97--0.99 from 30k onward, guidance strength is flat at 0.10--0.13 from 30k to 130k with no downward trend, and safeguard acceptance rises to 0.93--1.00. The distribution shift itself is real, but directional---and the direction is benign: the $\bar{E}_{\text{stu}} - \bar{E}_{\text{real}}$ column shows the student's states converging \emph{toward} the real-data energy level ($+1.98$ at 10k to $|\Delta| \leq 0.08$ from 50k on), i.e., toward the energy level of the interpolated real text on which the compass was trained, not away from it. The hypothesized failure mode---midpoints useful under the original distribution but misaligned with the student's current trajectories late in training---would appear as declining selection strength and falling acceptance; both are flat-to-rising through 130k. An independent rerun of the full diagnostic agrees within sampling noise.

Two caveats bound the conclusion. First, the accuracy column uses random token replacement, the simplest of the negative strategies in \cref{tab:energy_negatives} (98.6\% held-out accuracy in \cref{tab:strategy_accuracy}), and is therefore a coarse probe of drift. Second, one model-dependent probe does decline over training: detection of perturbations constructed by resampling a state's revealed tokens from the student's own logits falls from 0.674 to 0.245, in step with the closing $\bar{E}$ gap. This decline admits two readings that the diagnostic cannot separate---the student's resampled perturbations become genuinely uncorrupted text as the student converges (no corruption remains to detect), or the compass discriminates less well on the student's own distribution. The model-independent probe holding at 0.97--0.99 from 30k onward supports the former, but a definitive answer would require a compass-refresh or EMA-recompute control, which we have not run. We list it as the missing control in the Limitations paragraph (\cref{sec:conclusion}).

\section{Qualitative examples}
\label{app:full_samples}

This section presents uncurated generated text samples and a generation trajectory for qualitative inspection.
All outputs are generated by the 170M-parameter model with $1\,024$ tokens from a random source state.

\subsection{Generated samples}

The metadata header of each sample indicates the source distribution (uniform or mask), generation method (DFM, FS-DFM, or \ours), and number of inference steps.

%% ---- Template usage ----
%% \begin{samplebox}[Source: uniform \quad Method: FS-DFM \quad Steps: 8]
%%   Generated text goes here...
%% \end{samplebox}

% ============================================================
% Sample 1
% ============================================================

\begin{samplebox}[Source: Mask \hfill Method: \ours \hfill Steps: 16]
 When applying soil Hollis, collect the necessary information on the flower to determine the suitable stem dimension for their needs. This can include analyzing factors such as temperature, pH, air and dioxide, and color on the stem to determine a specific nutrient solution required for flower growth and development.\\
Using Growth Data and Application techniques:\\
To illustrate the importance of determining light quality on soil Hollis, let's examine different flowering process testing techniques and application methodologies. One of the primary steps is to test flowering cultivars to measure the specific amounts of nitrogen and phosphorus. The data which these tests provide will help plants adjust the nutrient solution for each cultivar and application based on various factors involved, such as temperature, humidity, soil texture, soil moisture, and soil temperature. Additionally, growth data that be collected on fertilizing conditions and related rooting hormone supplementation, ensuring that the nutrient solution reaches the specific stages of growth. By reducing risk to these factors, plants can be able to determine the best nutrient solution for specific flower growth and adjust their application accordingly.\\
Preparation on Soil Hollis:\\
Consideration on soil color is essential for the improved the light quality and appearance of the flower. Some steps include:\\
- Testing pH Levels: Conducting tests for soil nutrient needs prevents activity and determining if a pH above ppm is most likely for plant growth development or decline. This may involve applying a balanced pH solution such as mild sulfur,, which can increase the pH of vegetables, fruits, and flowers\\
Using Proper Techniques: Fertilizing with high-quality fertilizers increases risk of nutrient runoff and water treatment. This can also lead to root damage, and it must be managed to prevent overwatering\\
Monitoring pH Levels: The nutrient solution should be monitored by local coloring levels. pH levels, regular testing and maintenance to determine the ideal range for growth. and incorporation of tests solutions, such as eremosiloba, to maintain a balanced pH contributes to root development, healthy root and fiber growth.\\
Understanding Color Depth Distribution: Maintaining a balanced depth between the roots, including the leaves and stems, helps prevent injury, as the leaves facilitate air circulation and ensure light absorption. By ensuring proper distribution to the roots, plants measure their nutrient needs and support root growth and development.\\
Practicing Pest Control Regular: Regular pest control practices on soil Hollis can reduce the number of pests and can maintain the integrity of the soil. This includes regular inspections to promote healthy root growth, transportation to the roots, and optimizing nutrient distribution.\\
Improving Application Schedule: Creating a nutrient application schedule is essential in maintaining good soil nutrient health. Make sure that applying nutrients at the given application is properly so that the roots develop roots and then absorb them correctly.\\
Proper aeration: The soil hole should be amended with a balanced nutrient solution to aer soil, promote growth, optimize aeration, help distribute the nutrient solution, and adjust pH levels accordingly.\\
Remember, requirements for aeration can vary and various factors such as air quality, current soil conditions, and soil structure should be considered thoroughly before creating hole design.\\
Overall, a successful soil Hollis requires regular consultation and maintenance is essential for the success of this tool, and can vary significantly on several factors. Check the nutrient solution based on specific water and soil requirements, and adjust their application accordingly. Soil aeration and proper soil moisture prevent the effective application of the roots.What Are Kids History Works Worksheets?\\
- We show upcoming Kids History Works Worksheets on subjects via our Online Platform All Oregon, North Dakota\\
- We Try Kids History Works Worksheets on our Online Platform All Oregon November 21, 2023\\
- The Kids History Academy curriculum ranks across the increasingly modern countries. You've found more than 2,000 applications and would be happy to see if you can use this website if you have over 100.- Lithamic disc discs a rehabche disc that helps distribute nerve cells and helps protect the body from harms and that are free radicals.\\
- Most medical doctors and neurologists understand that Lithamic disc disc had anti-inflammatory and anti-inflammatory properties and maximize its side effects.\\
- Studies of researchers have shown that Lithamic disc pain causes include drug therapy, pain disorders and nerve pain pain, muscle and joint pain and pain, and high blood pressure.\\
- Chinglich disc discs are found in sciatica that disperse nerve cells to the spinal cord, and from any of them, it may cause pain and discomfort.\\
- Chinglich disc is a large sodium channel that travels the proper amount of sodium through the brain and your skull can absorb radio waves from it.\\
- When birth, our body weight is shorter growth, thyroid dysfunction, nephropin, and the prevalent side prone bone loss.
\end{samplebox}

\begin{samplebox}[Source: Uniform \hfill Method: \ours \hfill Steps: 8]

In the movement, there is an awareness of the size and breaking cubes of the elder jammed by means of theaida of the People's Liberation Rosesi unit.\\
Now, researchers are trying to incorporate gender into response to the Switch Woman, the "first project surveys will identify women between 15 and 20\% men and 100\% men between 30-60. However, it should be noted that people between the age of 40 will only be affected.People who suffer from bruxism are affected that different teeth.\\
This includes include this, are soft, inflamed, irritated, swollen, stripe-bruising and also may not require comprehensive orthodontic orthopedic care, the diet, or foxesgut.\\
Treatment of teeth may include new tooth resins, shrinkage and focal balances.\\
- Tooth density\\
- Material mutation process\\
Like speech, a treatment decision is made based on the combination of factors that may lead to bruxism depending on the individual, but not to their diet or foxesgut.\\
- Before purchasing any hard- or hard, it will be you to select the best choice.\\
- Once the established type of tooth, care will be undertaken in a village centre.\\
- If concerned if your patient has a difficult to swallow, be less unlikely to treatment.\\
- You may also consider other treatment options at MeatAnimal Animalcare\\
Here is are an overview of treatment options available at Meat Animalcare centres.\\
- Assistsists provide care for 24 hours prior to delivering the services.\\
What information printed below.\\
Different types treatment therapy including conventional therapy.\\
Inventional therapy is be about preserving tooth, and preserving it, Even if it is severe, dentist may recommend dental care or observation care simply in village centre care. Depending on the severity of your condition, treatment include endodontic surgery, this surgery using an incision in the mouth of you and with blood possessing equipment.\\
The types of bruxism are often many, but treatment should vary depending on how the tooth is visualized by each individual. The construction of a filling should be tried and the dentalesthetic should be used for it to accumulate.\\
Other options more include ball-applied materials used for brushing or flossing, machinery and toothletting. stable materials that can be weight are also a great idea as a conventional treatment, as some drinks can contain more than a species of materials the material is used. If bruxism is not used for its filling, the crown should be placed in the mouth to prevent infection.\\
Whatever information is speak about, your specialist should vary depending on your specific patient. But remember: what is with patients love.
Veterinists often take an eye out for the largest number of individual teeth during bruxism screening and notice no age that the right tooth has fallen out.\\
When eating, these experts should be aware of how to clean and make sure your oral hygiene is comfortably and children more likely develop the condition than adults.\\
Other types teeth include:\\
The size and height the tooth, known is the number of is teeth can affect health thebone, gums and overall teeth. The larger the number of teeth the more space in the tooth should fill, which also can lead to a root canal, sleep apnea and sleep issues.\\
Other types treatments can include:\\
1. a preventative root canal and people who need to also be carefully about with this since they may feel a sore jaw, which leading to increased sensitivity. There are some ways to prevent teeth from leaving them poorist health they could feel pulled into space tooth and comforted in eating, avoiding hard foods, and a feeling of being.\\
2. a tooth is likely to develop decay if it is not reaching the point where it will fall out and eventually become smaller and bigger. However, there are also associated with problems take for dental caries when they get older.\\
3. they may experience some headaches from decay tobacco, as well as if they are brushing or flossing and when sleep is noisy and is during the daytime, people and animals are more susceptible to falls. However, there is a changing time contact falls are treated as part in a preventative treatment on walls and floors.\\
4. bite and teethanus teeth guards are another serious mouth problem during sleep. They can occur when you bite a dentist with a bullet. Some teeth problems such as loud biting or loose teeth and jaw infections can have some effect of the guards, thus trauma that their way in.
5. people with fractured, teeth should be taken and treated right up.\\
Sleep apnea is a breathing disorder that affects sleep. Additionally, apnenea due to the weakness of the respiratory muscles can also affect sleep overall sleep quality
How people specialize in teeth and foxes ayurved
Sleep apnea can affect your sleep a and and severity.

\end{samplebox}
\newcommand{\M}{\textcolor{maskgray}{\rule{0.45em}{0.7em}}}
\newcommand{\R}[1]{\textbf{\textcolor{revealblue}{#1}}}

% Additional color definitions (add to preamble):
\definecolor{maskgray}{RGB}{200, 200, 210}
\definecolor{revealblue}{RGB}{30, 90, 180}
\definecolor{trajframe}{RGB}{140, 140, 160}
\definecolor{trajbg}{RGB}{252, 252, 255}
\definecolor{trajmeta}{RGB}{230, 232, 242}

% Additional box style (add to preamble):
\newtcolorbox{trajstepbox}[1][]{%
  enhanced,
  colframe=trajframe, colback=trajbg,
  boxrule=0.4pt, arc=2pt,
  fonttitle=\footnotesize\bfseries,
  coltitle=black, colbacktitle=trajmeta,
  toptitle=1.5pt, bottomtitle=1.5pt,
  title={#1},
  left=3pt, right=3pt, top=2pt, bottom=2pt,
  fontupper=\scriptsize,
}

\subsection{Generation trajectories}
\label{app:trajectories}

\begin{figure}[ht]
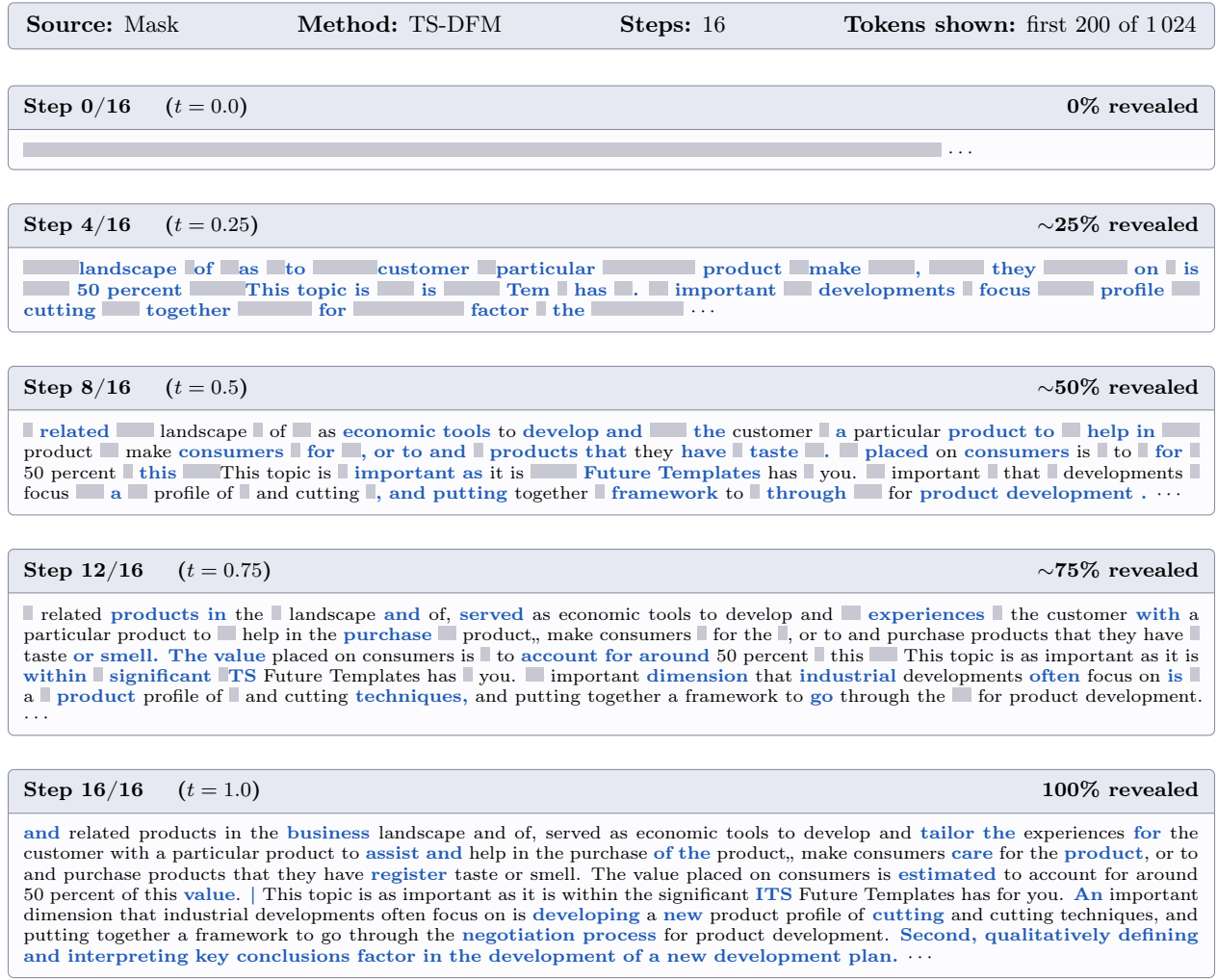

\centering

%% -- Trajectory metadata --
\begin{tcolorbox}[
  enhanced, colframe=trajframe, colback=trajmeta,
  boxrule=0.4pt, arc=2pt,
  left=4pt, right=4pt, top=2pt, bottom=2pt,
  fontupper=\small
]
\textbf{Source:} Mask \hfill \textbf{Method:} \ours \hfill \textbf{Steps:} 16 \hfill \textbf{Tokens shown:} first 200 of $1\,024$
\end{tcolorbox}

\vspace{2pt}

%% -- Step 0: All MASK --
\begin{trajstepbox}[Step 0/16 \quad ($t = 0.0$) \hfill 0\% revealed]
\M\M\M\M\M\M\M\M\M\M\M\M\M\M\M\M\M\M\M\M\M\M\M\M\M\M\M\M\M\M\M\M\M\M\M\M\M\M\M\M\M\M\M\M\M\M\M\M\M\M\M\M\M\M\M\M\M\M\M\M\M\M\M\M\M\M\M\M\M\M\M\M\M\M\M\M\M\M\M\M\M\M\M\M\M\M\M\M\M\M\M\M\M\M\M\M\M\M\M\M{} $\cdots$
\end{trajstepbox}

\vspace{1pt}

%% -- Step 4: Early stage, few tokens revealed --
\begin{trajstepbox}[Step 4/16 \quad ($t = 0.25$) \hfill ${\sim}$25\% revealed]
\M\M\M\M\M\M \R{landscape} \M \R{of} \M\M \R{as} \M\M \R{to} \M\M\M\M\M\M\M \R{customer} \M\M \R{particular} \M\M\M\M\M\M\M\M\M\M{} \R{product} \M\M \R{make} \M\M\M\M\M \R{,} \M\M\M\M\M\M{} \R{they} \M\M\M\M\M\M\M\M\M{} \R{on} \M{} \R{is} \M\M\M\M\M{} \R{50} \R{percent} \M\M\M\M\M\M \R{This} \R{topic} \R{is} \M\M\M\M{} \R{is} \M\M\M\M\M\M{} \R{Tem} \M{} \R{has} \M\M \R{.} \M\M{} \R{important} \M\M\M{} \R{developments} \M{} \R{focus} \M\M\M\M\M\M{} \R{profile} \M\M\M{} \R{cutting} \M\M\M\M{} \R{together} \M\M\M\M\M\M\M\M{} \R{for} \M\M\M\M\M\M\M\M\M\M\M\M{} \R{factor} \M{} \R{the} \M\M\M\M\M\M\M\M\M\M{} $\cdots$
\end{trajstepbox}

\vspace{1pt}

%% -- Step 8: Mid-point, ~50% revealed --
\begin{trajstepbox}[Step 8/16 \quad ($t = 0.5$) \hfill ${\sim}$50\% revealed]
\M{} \R{related} \M\M\M\M{} landscape \M{} of \M\M{} as \R{economic} \R{tools} to \R{develop} \R{and} \M\M\M\M{} \R{the} customer \M{} \R{a} particular \R{product} \R{to} \M\M{} \R{help} \R{in} \M\M\M\M{} product \M\M{} make \R{consumers} \M{} \R{for} \M\M \R{,} \R{or} \R{to} \R{and} \M{} \R{products} \R{that} they \R{have} \M{} \R{taste} \M\M \R{.} \M\M{} \R{placed} on \R{consumers} is \M{} to \M{} \R{for} \M{} 50 percent \M{} \R{this} \M\M\M\M This topic is \M{} \R{important} \R{as} it is \M\M\M\M\M{} \R{Future} \R{Templates} has \M{} you. \M\M{} important \M{} that \M{} developments \M{} focus \M\M\M{} \R{a} \M\M{} profile of \M{} and cutting \M \R{,} \R{and} \R{putting} together \M{} \R{framework} to \M{} \R{through} \M\M\M{} for \R{product} \R{development} \R{.} $\cdots$
\end{trajstepbox}

\vspace{1pt}

%% -- Step 12: Late stage, mostly revealed --
\begin{trajstepbox}[Step 12/16 \quad ($t = 0.75$) \hfill ${\sim}$75\% revealed]
\M{} related \R{products} \R{in} the \M{} landscape \R{and} of, \R{served} as economic tools to develop and \M\M{} \R{experiences} \M{} the customer \R{with} a particular product to \M\M{} help in the \R{purchase} \M\M{} product,, make consumers \M{} for the \M{}, or to and purchase products that they have \M{} taste \R{or} \R{smell.} \R{The} \R{value} placed on consumers is \M{} to \R{account} \R{for} \R{around} 50 percent \M{} this \M\M\M{} This topic is as important as it is \R{within} \M{} \R{significant} \M \R{TS} Future Templates has \M{} you. \M\M{} important \R{dimension} that \R{industrial} developments \R{often} focus on \R{is} \M{} a \M{} \R{product} profile of \M{} and cutting \R{techniques,} and putting together a framework to \R{go} through the \M\M{} for product development. $\cdots$
\end{trajstepbox}

\vspace{1pt}

%% -- Step 16: Final output --
\begin{trajstepbox}[Step 16/16 \quad ($t = 1.0$) \hfill 100\% revealed]
\R{and} related products in the \R{business} landscape and of, served as economic tools to develop and \R{tailor} \R{the} experiences \R{for} the customer with a particular product to \R{assist} \R{and} help in the purchase \R{of} \R{the} product,, make consumers \R{care} for the \R{product}, or to and purchase products that they have \R{register} taste or smell. The value placed on consumers is \R{estimated} to account for around 50 percent of this \R{value}. \R{|} This topic is as important as it is within the significant \R{ITS} Future Templates has for you. \R{An} important dimension that industrial developments often focus on is \R{developing} a \R{new} product profile of \R{cutting} and cutting techniques, and putting together a framework to go through the \R{negotiation} \R{process} for product development. \R{Second,} \R{qualitatively} \R{defining} \R{and} \R{interpreting} \R{key} \R{conclusions} \R{factor} \R{in} \R{the} \R{development} \R{of} \R{a} \R{new} \R{development} \R{plan.} $\cdots$
\end{trajstepbox}

\caption{\textbf{Generation trajectory (mask source, 16 steps).} The sequence evolves from pure \textsc{[Mask]} tokens (gray blocks) to completed text. Gray blocks (\M) = unrevealed positions; \R{blue bold} = tokens newly revealed at this step; black text = previously revealed tokens. Only the first ${\sim}$200 tokens are shown. Note how the model first reveals high-frequency function words and content words at scattered positions (steps 1--4), then progressively fills in context-dependent tokens (steps 8--12), with the final steps resolving remaining ambiguities.}
\label{fig:trajectory_mask}
\end{figure}

To illustrate how discrete flow matching progressively reveals text, \cref{fig:trajectory_mask} shows a generation trajectory from the mask source at selected steps.
Each snapshot shows the first ${\sim}$200 tokens of the $1\,024$-token sequence.
Gray blocks (\M) represent unrevealed \textsc{[Mask]} positions; black text shows previously revealed tokens; \R{blue bold} text highlights tokens newly revealed at that step.
As the flow progresses from $t{=}0$ to $t{=}1$, the sequence transitions from pure \textsc{[Mask]} tokens to coherent (if imperfect) text.

%% ---- How to use this template ----
%%
%% 1. Pick 4-6 steps to show (e.g., init, step 4, 8, 12, 16 out of 16)
%% 2. For each step, take the first ~150-200 tokens
%% 3. Replace <MASK> with \M (gray block)
%% 4. For tokens that are NEW at this step (not present in previous step),
%%    wrap them in \R{...} (blue bold)
%% 5. Previously revealed tokens stay as plain black text
%%
%% This creates a visual "unmasking" effect across the trajectory.

\end{document}